\documentclass[10pt,twocolumn,letterpaper]{article}

\usepackage[pagenumbers]{iccv} 

\usepackage{booktabs}
\usepackage{graphicx} 
\usepackage{multirow} %
\usepackage{bbding}
\usepackage{amssymb} 
\usepackage{pifont} 
\usepackage{colortbl}
\usepackage{tabularray}

\newcommand{\tabref}[1]{Table~\ref{#1}}

\def\ourmodel{USCNet}
\def\ourdataset{USC12K}
\definecolor{iccvblue}{rgb}{0.21,0.49,0.74}
\usepackage[pagebackref,breaklinks,colorlinks,allcolors=iccvblue]{hyperref}

\def\paperID{2555} 
\def\confName{ICCV}
\def\confYear{2025}

\title{Rethinking Detecting Salient and Camouflaged Objects in Unconstrained Scenes}

\author{
Zhangjun Zhou$^{1 }$\thanks{Both authors contributed equally to this research.}  \quad
  Yiping Li$^{1 *}$ \quad
  Chunlin Zhong$^{1 *}$ \quad
  Jianuo Huang$^1$ \quad \\
  Jialun Pei$^2$ \quad 
  Hua Li$^3$ \quad  
  He Tang$^{1 }$\thanks{Corresponding author: He Tang.} \\
  [2mm]
  $^1$School of Software Engineering, Huazhong University of Science and Technology \\
  $^2$School of Computer Science and Engineering, The Chinese University of Hong Kong\\
  $^3$	School of Computer Science and Technology, Hainan University
}

\begin{document}
\maketitle
\begin{abstract}
While the human visual system employs distinct mechanisms to perceive salient and camouflaged objects, existing models struggle to disentangle these tasks. Specifically, salient object detection (SOD) models frequently misclassify camouflaged objects as salient, while camouflaged object detection (COD) models conversely misinterpret salient objects as camouflaged. We hypothesize that this can be attributed to two factors: (i) the specific annotation paradigm of current SOD and COD datasets, and (ii) the lack of explicit attribute relationship modeling in current models. Prevalent SOD/COD datasets enforce a mutual exclusivity constraint, assuming scenes contain either salient or camouflaged objects, which poorly aligns with the real world. Furthermore, current SOD/COD methods are primarily designed for these highly constrained datasets and lack explicit modeling of the relationship between salient and camouflaged objects. In this paper, to promote the development of unconstrained salient and camouflaged object detection, we construct a large-scale dataset, \textbf{USC12K}, which features comprehensive labels and four different scenes that cover all possible logical existence scenarios of both salient and camouflaged objects. To explicitly model the relationship between salient and camouflaged objects, we propose a model called \textbf{USCNet}, which introduces two distinct prompt query mechanisms for modeling inter-sample and intra-sample attribute relationships. Additionally, 
We designed \textbf{CSCS} to evaluate the model's ability to distinguish salient and camouflaged objects. Our method achieves SOTA performance across all scenes.  Code and dataset: \href{https://github.com/ssecv/USCNet}{GitHub}.



\end{abstract}



\section{Introduction}
\label{sec:intro}

The attention mechanism is one of the key cognitive functions of humans~\cite{posner1990attention}. In real-world scenarios, people are often drawn to salient objects while overlooking camouflaged ones. From the perspective of the human visual recognition system, salient and camouflaged objects represent opposing concepts. The goal of Salient Object Detection (SOD) is to detect objects in an image that the human visual system considers most salient or attention-grabbing, while Camouflaged Object Detection (COD) aims to detect objects that are difficult to perceive or blend seamlessly with their surroundings~\cite{li2021uncertainty}. In computer vision, these two tasks typically represent the model outputs using binary masks.
Both of them exhibit significant potential across various fields, such as anomaly detection in medical image analysis~\cite{tang2023source}, obstacle recognition in autonomous driving, camouflage detection in military reconnaissance~\cite{lin2019metaheuristic}, and wildlife tracking in environmental monitoring~\cite{stevens2009animal}.

Existing SOD and COD methods have achieved remarkable progress in detecting salient and camouflaged objects. Some unified methods~\cite{liu2023explicit,luo2024vscode,Spider,li2021uncertainty} have improved the model's generalization ability in both SOD and COD tasks by jointly training SOD and COD datasets. However, we observe a counterintuitive phenomenon: although salient and camouflaged objects are conceptually opposing, SOD models frequently misclassify camouflaged objects as salient, while COD models conversely misinterpret salient objects as camouflaged. Such misdetections are undesirable and contradict the intended objectives of these two tasks.

\begin{table}[t!]
    \centering
    \caption{Misinterpretation is prevalent when SOD/COD models applied across tasks. \textbf{Left:} Inference results of SOD models on COD datasets. VSCode uses the SOD prompt. \textbf{Right:} Inference results of COD models on SOD datasets. VSCode uses the COD prompt. The metric is $F_\beta^\omega$. EV refers to the Expected Value.}
    \label{tab:experimental_phenomenon}
    \begin{minipage}{0.5\linewidth}
    \centering
    \scriptsize
    \setlength\tabcolsep{100pt}
    \renewcommand{\arraystretch}{1.06}
    \renewcommand{\tabcolsep}{0.8mm}
    \begin{tabular}{l|c|c|c}
        \toprule
         & \multicolumn{2}{c|}{\textbf{COD Datasets}} \\
        \cline{2-3} 
        \multirow{-2}{*}{\textbf{SOD Models}} & COD10K & NC4K & \multirow{-2}{*}{\textbf{EV}}\\
        \hline \hline
        ICON~\cite{zhuge2022salient} & 0.6384 & 0.7522 & 0\\
        F3Net~\cite{wei2020f3net} & 0.4327 & 0.6229 & 0\\
        VSCode~\cite{luo2024vscode} & 0.7145 & 0.8054 & 0\\
        \bottomrule
    \end{tabular}
    \end{minipage}%
    \begin{minipage}{0.5\linewidth}
    \centering
        \scriptsize
    \setlength\tabcolsep{100pt}
    \renewcommand{\arraystretch}{1.06}
    \renewcommand{\tabcolsep}{0.8mm}
    \begin{tabular}{l|c|c|c}
        \toprule
         & \multicolumn{2}{c|}{\textbf{SOD Datasets}} \\
        \cline{2-3} 
        \multirow{-2}{*}{\textbf{COD Models}} & DUTS & HKU-IS & \multirow{-2}{*}{\textbf{EV}}\\
        \hline \hline
        SINet-V2~\cite{fan2021concealed} & 0.7412 & 0.7691 & 0\\
        PFNet~\cite{mei2021camouflaged} &0.7361  &0.7657  & 0\\
        VSCode~\cite{luo2024vscode} & 0.8614 & 0.8733 & 0\\
        \bottomrule
    \end{tabular}
    \end{minipage}%
\end{table}

To further validate this phenomenon, we select five models that used their original pre-trained weights to perform inference on four datasets. As shown in Table \ref{tab:experimental_phenomenon}, the SOD model, such as ICON~\cite{zhuge2022salient}, achieves an unexpectedly high $F_\beta^\omega$ of 0.6384 when inferring on the COD dataset COD10K~\cite{fan2020camouflaged}, while the COD model, such as SINet-V2~\cite{fan2021concealed}, reaches 0.7412 on the SOD dataset DUTS~\cite{wang2017learning}. The unified model VSCode~\cite{luo2024vscode}, using the COD prompt, achieves 0.8733 on the SOD dataset HKU-IS. The false positive scores of these methods remain significantly higher than the expected value of 0.

\textbf{We assume the first reason lies in the specific annotation paradigm of existing SOD and COD datasets.} While existing SOD and COD datasets have significantly advanced their respective fields through specialized design~\cite{fan2023advances,gao2024multi}, they impose strict constraints, assuming that scenes contain either salient or camouflaged objects and providing only a single type of attribute label. For instance, in COD datasets, certain scenarios include salient objects, yet these objects are categorized as background simply because they are non-camouflaged. As a result, COD models become insensitive to salient objects, failing to accurately delineate the boundaries of the saliency feature space. A similar issue arises in SOD models. This specialized annotation paradigm presents even greater challenges for unified models that aim to address both tasks, as objects with identical visual characteristics may receive conflicting labels across different dataset types. Specifically, salient objects are labeled as background in COD datasets while remaining labeled as salient in SOD datasets, and the same inconsistency applies to camouflaged objects. Such labeling discrepancies may theoretically lead to information loss during the multi-task learning process, impairing the model’s ability to generalize effectively.

In this paper, our goal is to advance the development of unconstrained salient and camouflaged object detection, enabling the model to detect objects in all possible logical scenarios of both salient and camouflaged objects. To achieve this, we construct a new dataset named \textbf{USC12K}. The dataset consists of 12,000 images with complete annotations, covering four scenes that encompass all logically possible cases of the existence of salient and camouflaged objects. In addition to the correction and optimization of existing datasets, we also collect and manually annotate 2,617 images containing both salient and camouflaged objects, and 1,436 images containing neither, from the internet, to ensure a balanced distribution across the four scenes. We evaluate 21 methods relevant to SOD and COD tasks on USC12K to establish a comprehensive benchmark for unconstrained salient and camouflaged object detection, aiming to drive further research in SOD and COD.

\begin{figure}[t]
\centering
\includegraphics[width=0.95\linewidth]{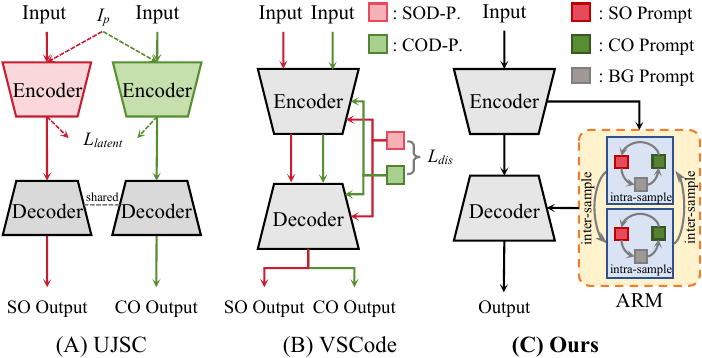}
\caption{Architecture comparisons of unified models. SOD-P and COD-P represent prompts for the SOD and COD tasks. SO prompt, CO prompt, and BG prompt represent prompts for the salient, camouflaged, and background attributes.
}
\label{fig:unifymodel-comp-forUSCOD}
\end{figure}

\textbf{We assume the second reason lies in the absence of adaptive mechanisms for attribute relationship modeling under unconstrained conditions.} The existing SOD and COD models have achieved significant success in single-task performance. However, they are primarily designed for restricted, single-scene datasets, and the detection strategy focused on a single attribute of objects is not conducive to adapting to the detection of objects exhibiting two opposite visual modes. Although unified models achieve the detection of both object attributes by sharing network components, their process remains suboptimal for unconstrained salient and camouflaged object detection. Specifically, they typically employ a separate approach to learn the detection of salient and camouflaged objects, which is designed for previously constrained datasets. Although some models attempt to establish connections between the two attributes through techniques such as contrastive learning, as shown in Figure~\ref{fig:unifymodel-comp-forUSCOD}, UJSC~\cite{li2021uncertainty} minimizes the Latent Space Loss $L_{latent}$ to increase the distance between the distributions of the salient encoder and camouflaged encoder feature spaces on an additional  dataset \( I_p \). Similarly, VSCode~\cite{luo2024vscode} minimizes the Discrimination Loss $L_{dis}$ to reduce the similarity between the salient and camouflaged prompts. However, they still lack explicit modeling of the relationship between salient and camouflaged objects within the network, and their loss functions are designed to model inter-sample relationships between saliency and camouflage outside the network, without capturing attribute relationships within individual samples. As a result, these models are unable to effectively learn their deep commonalities and differences in unconstrained scenarios.

To explicitly model the relationship between salient and camouflaged objects in unconstrained scenarios, we propose a network with a unified optimization pipeline, named \textbf{U}nconstrained \textbf{S}alient and \textbf{C}amouflaged \textbf{Net} \textbf{(USCNet)}, which has the following two advantages: 1) USCNet employs an attribute prompt framework based on SAM, incorporating a SAM encoder with an adapter to leverage SAM's scene generalization capabilities and enhance adaptability in complex scenarios. 2) An Attribute Relation Modeling (ARM) module is introduced to model the relationship of saliency and camouflage combined with two complementary prompt queries: Inter-SPQ learns general features and captures global information by modeling attribute relationships between samples, while Intra-SPQ models attribute relationships within a sample by focusing on its contextual information to learn the specific relationship between salient and camouflaged objects within the same sample. 

Moreover, existing metrics fail to quantify the model's confusion between salient and camouflaged objects, as they mainly evaluate the degree of foreground-background separation, such as weighted F-measure~\cite{margolin2014evaluate}. To fill this gap, we design a metric called Camouflage-Saliency Confusion Score (CSCS) to evaluate the model's ability to distinguish salient from camouflaged objects.

In summary, our contributions are listed as follows:
\begin{itemize}
\item We construct a large-scale unconstrained salient and camouflaged object detection dataset, \textbf{USC12K}, with comprehensive annotations. To our knowledge, this is the first dataset that is unconstrained regarding the existence of salient and camouflaged objects.
\item We propose a SAM-based model, \textbf{USCNet}, with an ARM module to learn attribute relationships and two query types for modeling inter- and intra-sample relationships.

\item A new metric, \textbf{CSCS}, is introduced to assess the model's confusion between salient and camouflaged objects.
\item We evaluated 21 related methods on the USC12K, establishing a comprehensive benchmark. Our model achieved SOTAs performance across all metrics in all scenarios.
\end{itemize}

\section{Related Work}
\label{sec:formatting}
\subsection{Salient and Camouflaged Object Detection}

\textbf{Classic SOD and COD.} In recent years, salient object detection models have focused on better detecting salient objects in images using various approaches. The main approaches can be divided into attention-based methods~\cite{liu2018picanet,piao2019depth,zhang2018progressive,pei2022transformer}, multi-level feature-based methods~\cite{fang2022densely,hou2017deeply,pang2020multi,wang2017stagewise}, and recurrent-based methods~\cite{deng2018r3net,liu2016dhsnet,wang2018salient}.
Saliency detection~\cite{zhang2021auto,liu2021visual,zhuge2022salient,pei2024calibnet} primarily focuses on achieving saliency predictions while preserving the structure. Compared to SOD methods, current COD methods~\cite{fan2021concealed,mei2021camouflaged,pang2022zoom,he2023camouflaged,pei2022osformer} focus primarily on edge-aware perception and texture perception. They are mainly divided into the following two types: multi-level feature-based methods~\cite{zhang2022preynet,yang2021uncertainty,ren2021deep,zhai2022exploring} and edge joint learning~\cite{zhai2021mutual,sun2022boundary,he2023camouflaged}. Both the SOD and COD models are designed for two tasks independently and lack modeling of the relationship between the two.

\noindent\textbf{Unified.} Recently, some works~\cite{liu2023explicit,luo2024vscode,li2021uncertainty,Spider} have already begun to unify the two tasks and attempt to establish connections between the two tasks. VSCode~\cite{luo2024vscode} learns discriminative salient prompts and camouflage prompts by minimizing the cosine similarity between SOD and COD task prompts. UJSC~\cite{li2021uncertainty} introduces the additional PASCAL VOC 2007 dataset to enhance the separation between the salient feature extractor and the camouflage feature extractor. Spider~\cite{Spider} distinguishes different attributes of the target by mining foreground/background-related semantic cues in the global context. EVP~\cite{liu2023explicit} learns task-specific visual prompts to distinguish salient and camouflaged objects.
While existing unified models still struggle to distinguish between salient and camouflaged objects due to the lack of directly modeling the relationship between saliency and camouflage, we address this limitation by introducing an ARM module to explicitly model their relationship via inter-sample and intra-sample interactions.

\subsection{Applications of SAM}

The Segment Anything Model (SAM)~\cite{kirillov2023segment} represents a significant advancement in scene segmentation using large vision models. Current works leveraging SAM ~\cite{chen2023sam,xiong2023efficientsam,ma2024segment} showcase its adaptability to downstream tasks, notably in areas where traditional segmentation models struggle, such as EfficientSAM~\cite{xiong2023efficientsam} and MedSAM~\cite{ma2024segment}. More recently, the release of SAM2~\cite{ravi2024sam} enhances the original SAM's ability to handle video content while demonstrating improved segmentation accuracy and inference efficiency in image segmentation across various downstream applications~\cite{yan2024biomedical,lian2024evaluation,pei2024evaluation,pei2025synergistic}. Some works that use SAM for SOD and COD are closely related to our research. MDSAM~\cite{gao2024multi} is a multi-scale and detail-enhanced SOD model based on SAM, aimed at improving the performance and generalization capability of the SOD task. SAM-Adapter~\cite{chen2023sam} and SAM2-Adapter~\cite{chen2024sam2} offer a parameter-efficient fine-tuning way to enhance the performance of SAM and SAM2 in downstream tasks by adding task-specific knowledge. Building on the strong generalization capabilities of SAM, we strive to explore its capacity for detecting salient and camouflaged objects in unconstrained scenarios.
\section{The Proposed \ourdataset~Dataset}

The current datasets for camouflaged object detection, such as COD10K~\cite{fan2020camouflaged}, CAMO~\cite{le2019anabranch}, NC4K~\cite{lv2021simultaneously}, primarily feature scenes with exclusively camouflaged objects. Similarly, datasets for salient object detection, such as DUTS~\cite{wang2017learning} and HKU-IS~\cite{li2015visual}, predominantly focus on scenes with solely salient objects.
Samples containing both salient and camouflaged objects are extremely scarce. Moreover, even when a small number of samples with both salient and camouflaged objects are available, only one attribute of the objects is annotated, which hinders the realization of unconstrained salient and camouflaged object detection.
Therefore, we introduce the \textbf{USC12K}, a dataset that includes more comprehensive and complex scenarios for unconstrained salient and camouflaged object detection. It includes scenes with both salient and camouflaged objects, scenes with only one type, and scenes without either. We will describe the details of USC12K in terms of three key aspects, as follows.


\subsection{Data Collection}

\begin{figure}[t]
\centering
\includegraphics[width=0.96\linewidth]{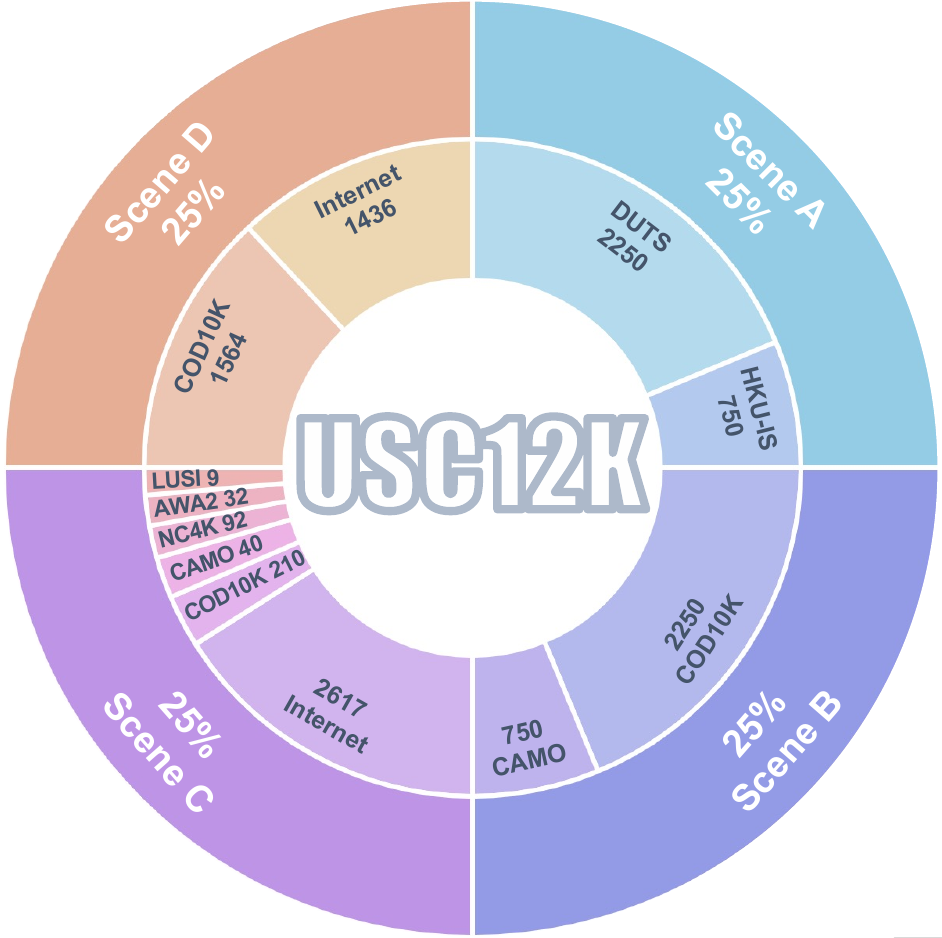}
\caption{The data sources and distribution of the four scenes.
}
\label{fig:data source}
\end{figure}

Under the premise of ensuring sample balance, we collect 12,000 images from 8 different sources and divide them into four scenes after manual filtering: (A) Scenes with only salient objects: 3,000 images containing only salient objects selected from SOD datasets DUTS and HKU-IS; (B) Scenes with only camouflaged objects: 3,000 images containing only camouflaged objects selected from COD datasets COD10K and CAMO; (C) Scenes with both salient and camouflaged objects: 342 images from the COD datasets COD10K, CAMO, and NC4K, along with 41 images from the datasets LSUI~\cite{peng2023u} and AWA2~\cite{xian2018zero}, and an additional 2,617 images collected from the internet, making a total of 3,000 images; (D) Scenes without salient and camouflaged objects, considered as background: 1,564 images from COD10K, and 1,436 images from the internet, making a total of 3,000 images. Each scene has been meticulously reviewed by human annotators to ensure correct scene classification. Finally, we get 12,000 images, with the training set containing 8,400 images and the test set containing 3,600 images. The data source is shown in Figure~\ref{fig:data source}.


\begin{table}[t!]
    \centering
    \caption{
    Data analysis of existing datasets.
    }\label{tab:dataset_table1}
    
    \scriptsize
    \setlength\tabcolsep{100pt}
    \renewcommand{\arraystretch}{0.9}
    \renewcommand{\tabcolsep}{0.8mm}

    \begin{tabular}{cl|cccccc}
        \toprule
            &Dataset& \#Ann. IMG & Class & {Scene A} &  {Scene B} & Scene C & Scene D \\
    \hline \hline 
        \multirow{13}{*}{}  
& SOD~\cite{movahedi2010design}  & 300 & - & 300& \textcolor{lightgray}{\XSolidBrush}  & \textcolor{lightgray}{\XSolidBrush} & \textcolor{lightgray}{\XSolidBrush}   \\
& PASCAL-S~\cite{li2014secrets} & 850 & - & 850& \textcolor{lightgray}{\XSolidBrush}  & \textcolor{lightgray}{\XSolidBrush} & \textcolor{lightgray}{\XSolidBrush}   \\
& ECSSD~\cite{yan2013hierarchical}  & 1000 & - & 1000& \textcolor{lightgray}{\XSolidBrush}  & \textcolor{lightgray}{\XSolidBrush} & \textcolor{lightgray}{\XSolidBrush}     \\
& HKU-IS~\cite{li2015visual}  & 4447 & - & 4447& \textcolor{lightgray}{\XSolidBrush}  & \textcolor{lightgray}{\XSolidBrush} & \textcolor{lightgray}{\XSolidBrush}   \\
& MSRA-B~\cite{DBLP:journals/pami/LiuYSWZTS11}  & 5000 & - & 5000& \textcolor{lightgray}{\XSolidBrush}  & \textcolor{lightgray}{\XSolidBrush} & \textcolor{lightgray}{\XSolidBrush}   \\
& DUT-OMRON~\cite{yang2013saliency}  & 5168 & - & 5168& \textcolor{lightgray}{\XSolidBrush}  & \textcolor{lightgray}{\XSolidBrush} & \textcolor{lightgray}{\XSolidBrush}   \\
& MSRA10K~\cite{DBLP:journals/pami/ChengMHTH15}  & 10000 & - & 10000& \textcolor{lightgray}{\XSolidBrush}  & \textcolor{lightgray}{\XSolidBrush} & \textcolor{lightgray}{\XSolidBrush}   \\
& DUTS~\cite{wang2017learning}  & 15572 & - & 15572& \textcolor{lightgray}{\XSolidBrush}  & \textcolor{lightgray}{\XSolidBrush} & \textcolor{lightgray}{\XSolidBrush}   \\
& SOC~\cite{fan2018SOC}  & 3000 & 80 & 3000& \textcolor{lightgray}{\XSolidBrush}  & \textcolor{lightgray}{\XSolidBrush} & \textcolor{lightgray}{\XSolidBrush}   \\
    \hline 
        \multirow{5}{*} {} 
& CAMO~\cite{le2019anabranch}  & 1250 & 8   & \textcolor{lightgray}{\XSolidBrush}& 1250 & \textcolor{lightgray}{\XSolidBrush}& \textcolor{lightgray}{\XSolidBrush}   \\
& CHAMELEON~\cite{chameleon} & 76 & -   & \textcolor{lightgray}{\XSolidBrush} & 76& \textcolor{lightgray}{\XSolidBrush}& \textcolor{lightgray}{\XSolidBrush}   \\
& NC4K~\cite{lv2021simultaneously}  & 4121 & -   & \textcolor{lightgray}{\XSolidBrush}& 4121 & \textcolor{lightgray}{\XSolidBrush}& \textcolor{lightgray}{\XSolidBrush}   \\
& COD10K~\cite{fan2020camouflaged}  & 7000 & 78  &\textcolor{lightgray}{\XSolidBrush}  & 5066& \textcolor{lightgray}{\XSolidBrush}& 1934 \\
    \hline 
    \cellcolor{iccvblue!20} &\cellcolor{iccvblue!20}{\textbf{USC12K(Ours)}}  &\cellcolor{iccvblue!20}12000 &\cellcolor{iccvblue!20}179 &\cellcolor{iccvblue!20}3000 &\cellcolor{iccvblue!20}3000 &\cellcolor{iccvblue!20}3000 &\cellcolor{iccvblue!20}3000   \\
        \bottomrule  
    \end{tabular}
\end{table}

\begin{figure*}[!t]
    \centering
    \footnotesize

    \renewcommand{\arraystretch}{1.4} 
    \begin{tabular}{@{}
    c@{\hskip 5pt}
    c@{\hskip 1pt}
    c@{\hskip 1pt}
    c@{\hskip 1pt}
    c@{\hskip 1pt}
    c@{\hskip 1pt}
    c@{\hskip 1pt}
    c@{\hskip 1pt}
    c@{\hskip 1pt}c@{\hskip 1pt}c@{}}
        
        & \multicolumn{2}{c}{\cellcolor{blue!10} \makebox[2.95cm][c]{Underwater, Aquatic}} 
        &
        \multicolumn{2}{c}{\cellcolor{yellow!10} \makebox[2.95cm][c]{Wild, Animal}} 
        & 
        \multicolumn{2}{c}{\cellcolor{green!10} \makebox[2.95cm][c]{Forest, Plant}} 
        
         & 
        \multicolumn{2}{c}{\cellcolor{blue!13} \makebox[2.95cm][c]{Sky, Bird}} 
        
        & 
        \multicolumn{2}{c}{\cellcolor{gray!10} \makebox[2.95cm][c]{City, Human}} 
        \\
    \end{tabular}
    
    \renewcommand{\arraystretch}{0.3}
    \begin{tabular}{@{}c@{\hskip 5pt}c@{\hskip 1pt}
    c@{\hskip 1pt}
    c@{\hskip 1pt}
    c@{\hskip 1pt}
    c@{\hskip 1pt}
    c@{\hskip 1pt}
    c@{\hskip 1pt}
    c@{\hskip 1pt}
    c@{\hskip 1pt}
    c@{}}

        \raisebox{0.1cm}{\makebox[0pt][c]{\rotatebox{90}{\footnotesize \textit{Scene A}}}} &
        \includegraphics[width=1.69cm, height=1.06cm]{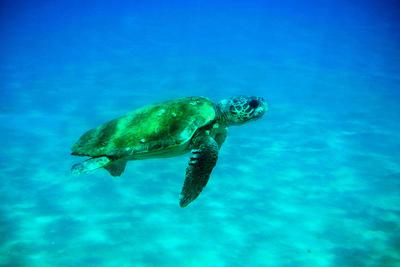} &
        \includegraphics[width=1.69cm, height=1.06cm]{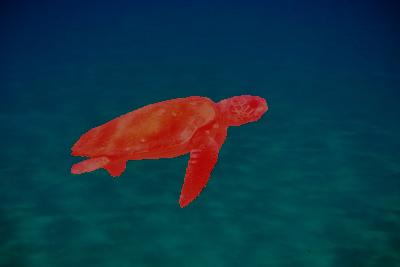} &
        \includegraphics[width=1.69cm, height=1.06cm]{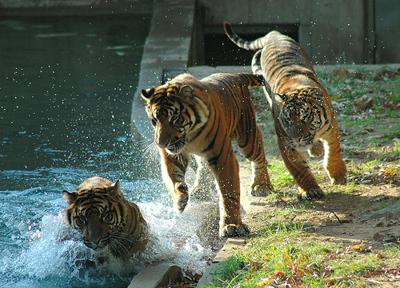} &
        \includegraphics[width=1.69cm, height=1.06cm]{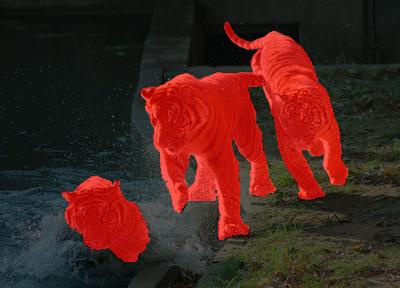} &
        \includegraphics[width=1.69cm, height=1.06cm]{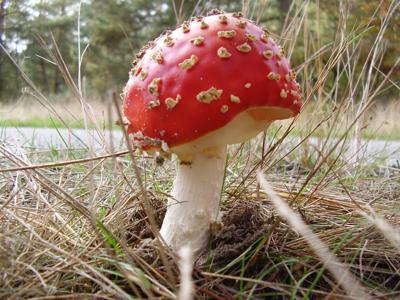} &
        \includegraphics[width=1.69cm, height=1.06cm]{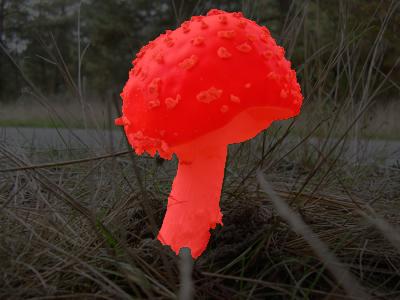} &
        \includegraphics[width=1.69cm, height=1.06cm]{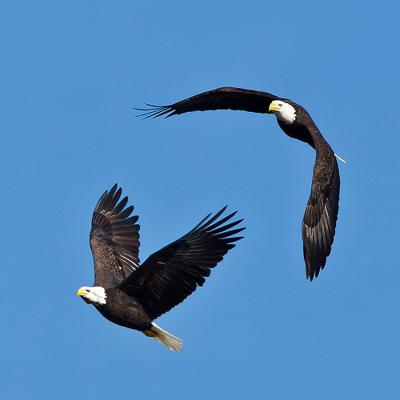} &
        \includegraphics[width=1.69cm, height=1.06cm]{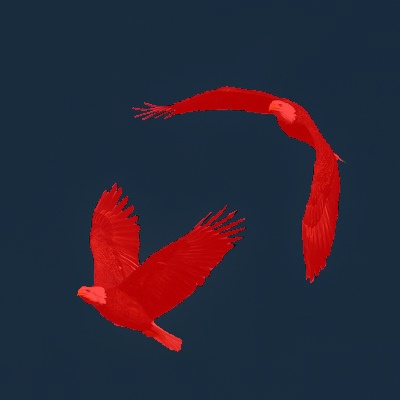} &
        \includegraphics[width=1.69cm, height=1.06cm]{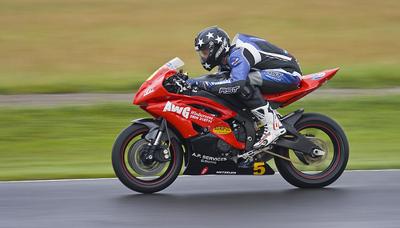} &
        \includegraphics[width=1.69cm, height=1.06cm]{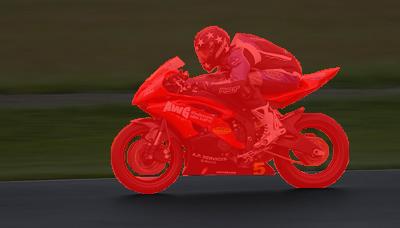} \\

        \raisebox{0.1cm}{\makebox[0pt][c]{\rotatebox{90}{\footnotesize \textit{Scene B}}}} &
        \includegraphics[width=1.69cm, height=1.06cm]{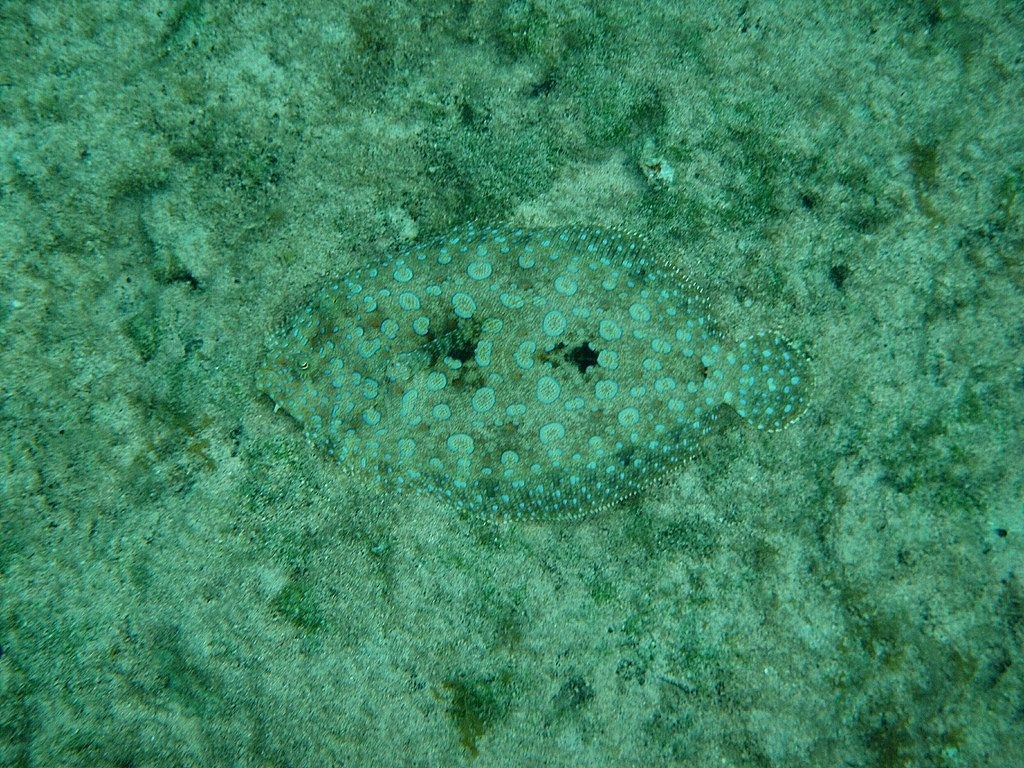} &
        \includegraphics[width=1.69cm, height=1.06cm]{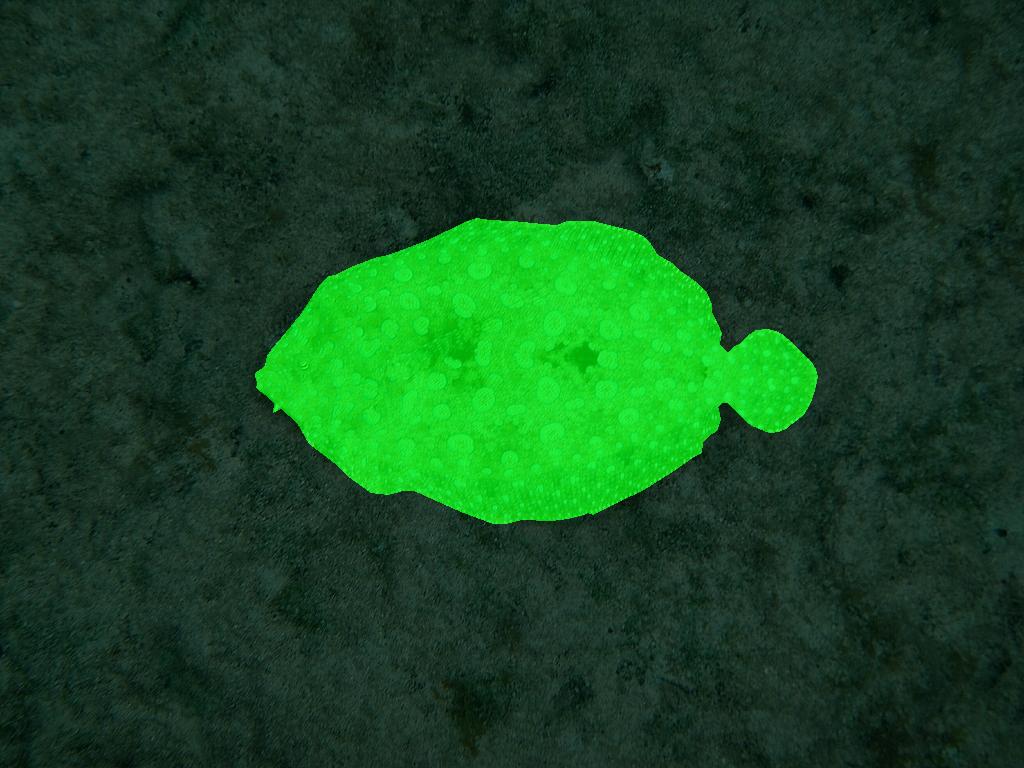} &
        \includegraphics[width=1.69cm, height=1.06cm]{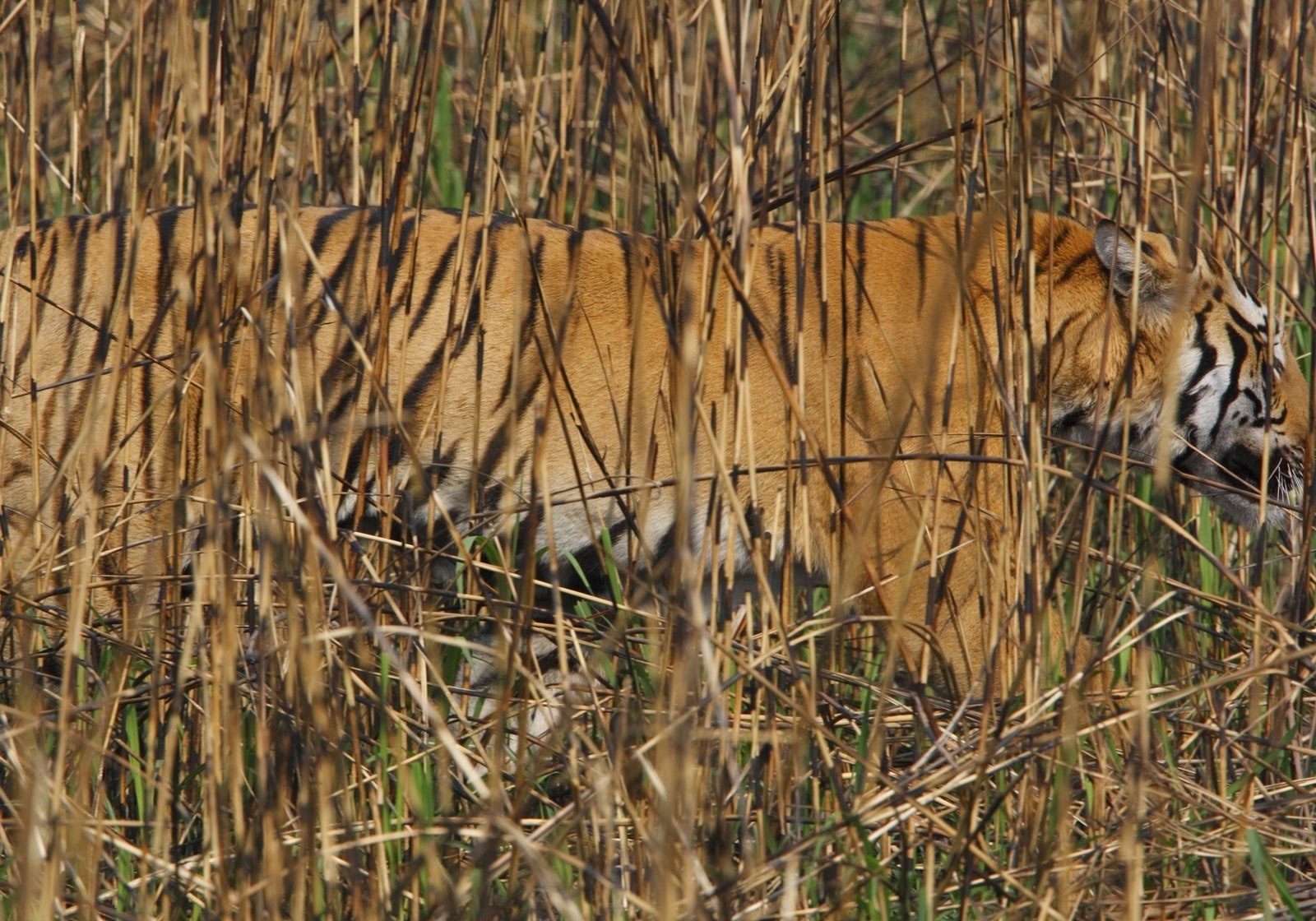} &
        \includegraphics[width=1.69cm, height=1.06cm]{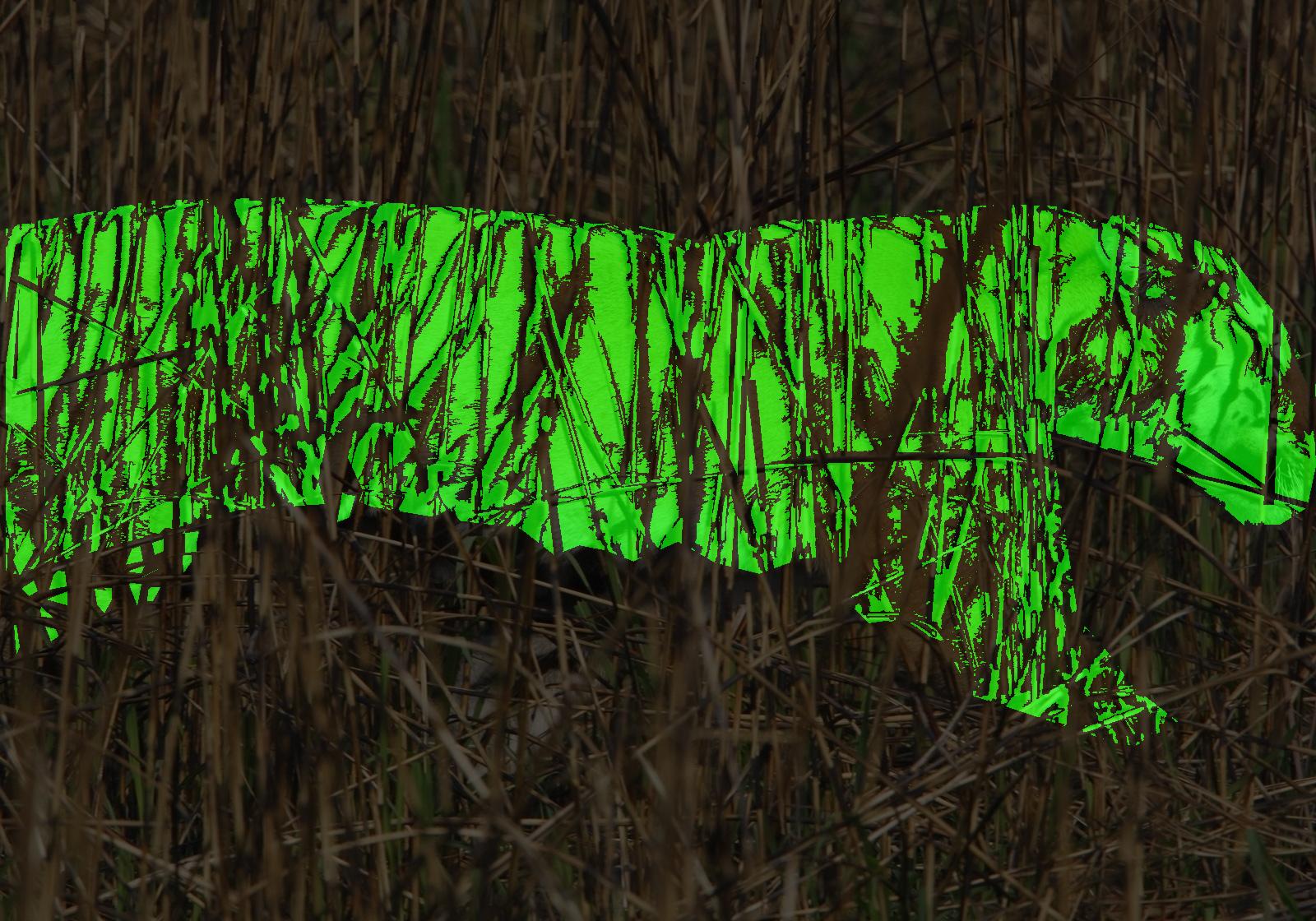} &
        \includegraphics[width=1.69cm, height=1.06cm]{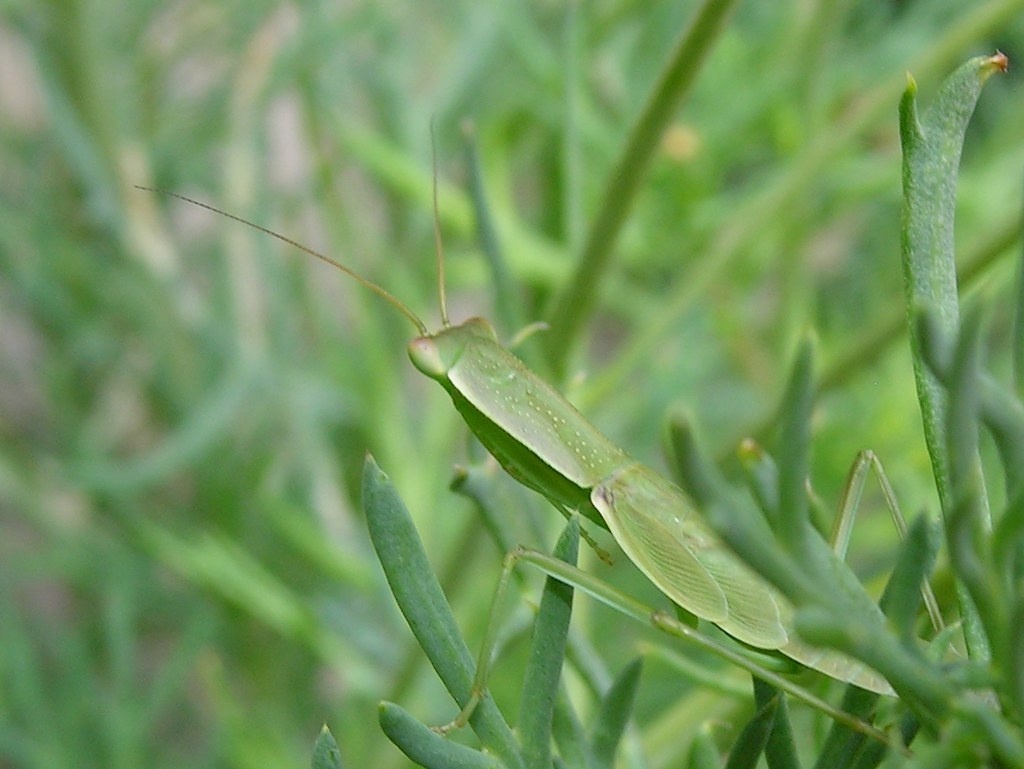} &
        \includegraphics[width=1.69cm, height=1.06cm]{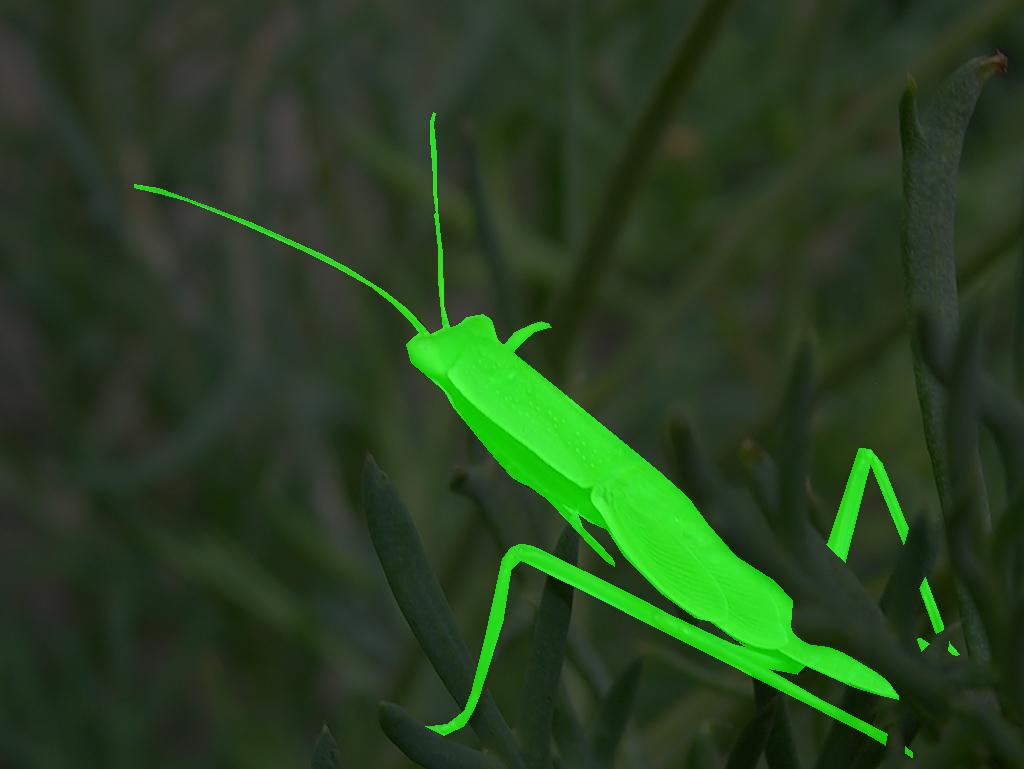} &
        \includegraphics[width=1.69cm, height=1.06cm]{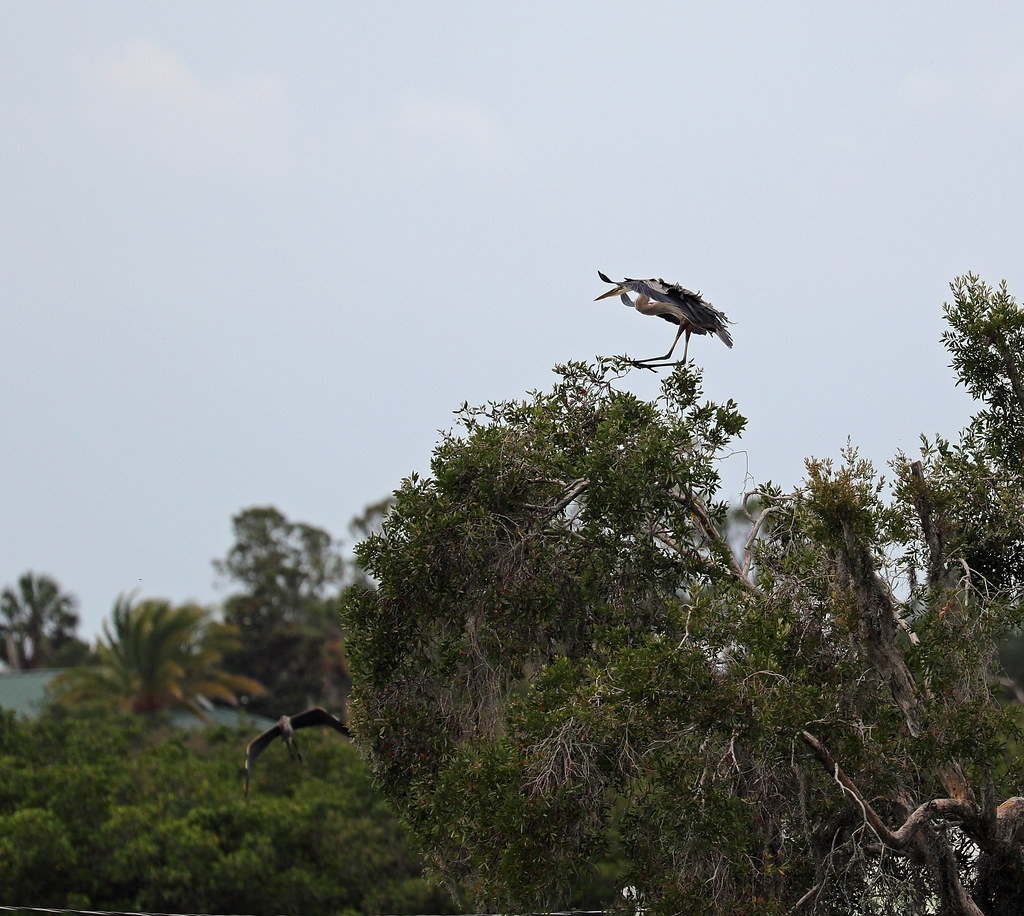} &
        \includegraphics[width=1.69cm, height=1.06cm]{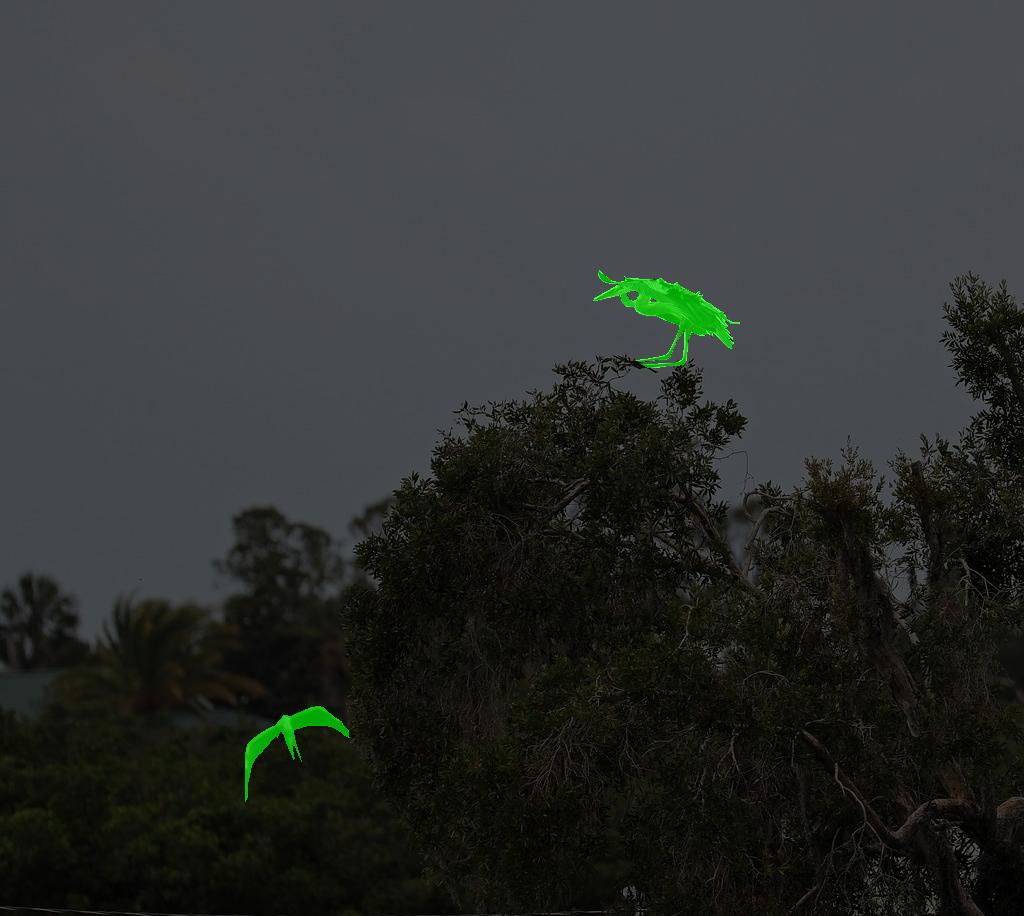} &
        \includegraphics[width=1.69cm, height=1.06cm]{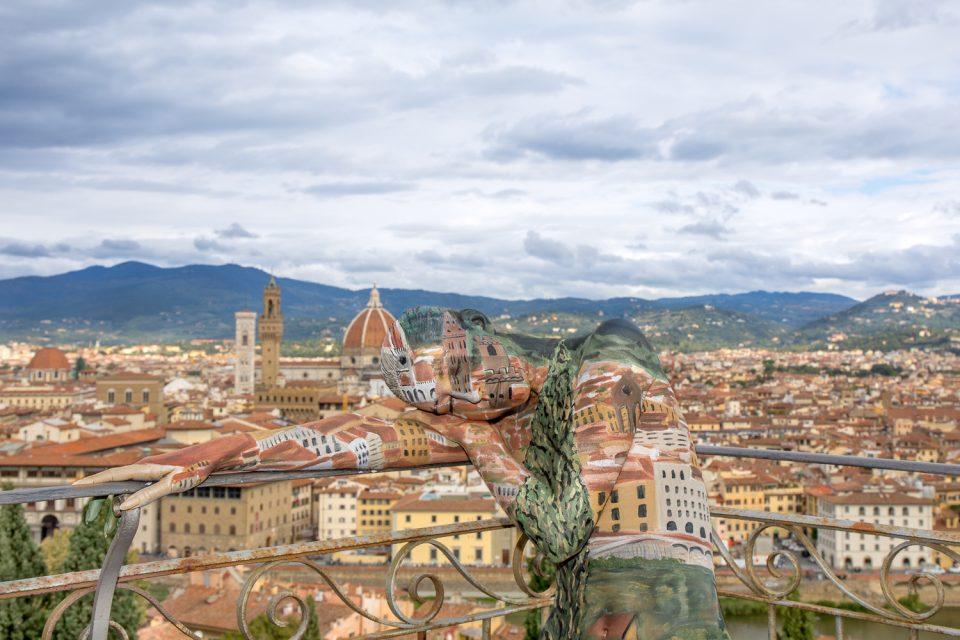} &
        \includegraphics[width=1.69cm, height=1.06cm]{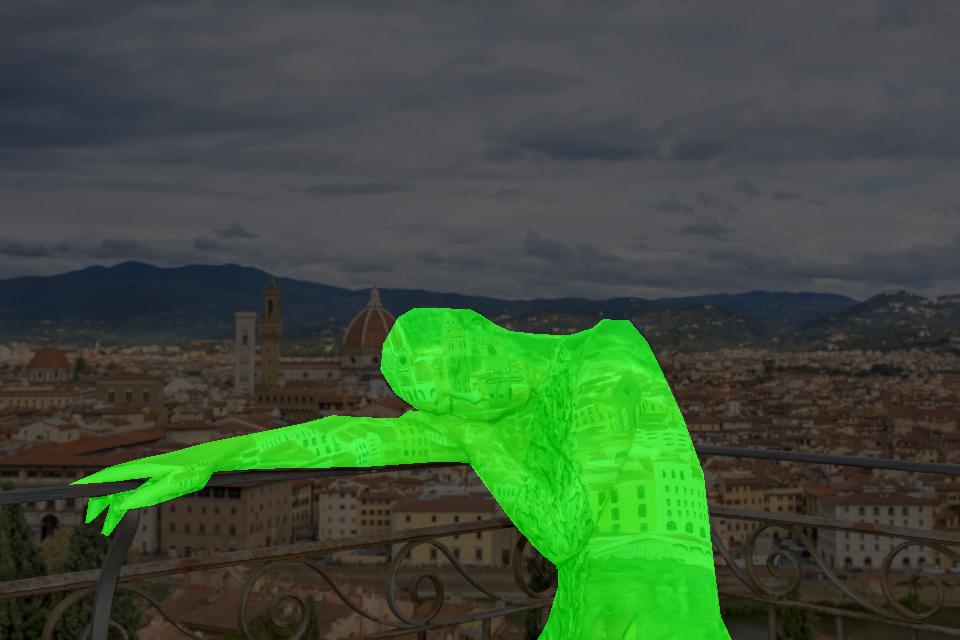} \\

        \raisebox{0.1cm}{\makebox[0pt][c]{\rotatebox{90}{\footnotesize \textit{Scene C}}}} &
        \includegraphics[width=1.69cm, height=1.06cm]{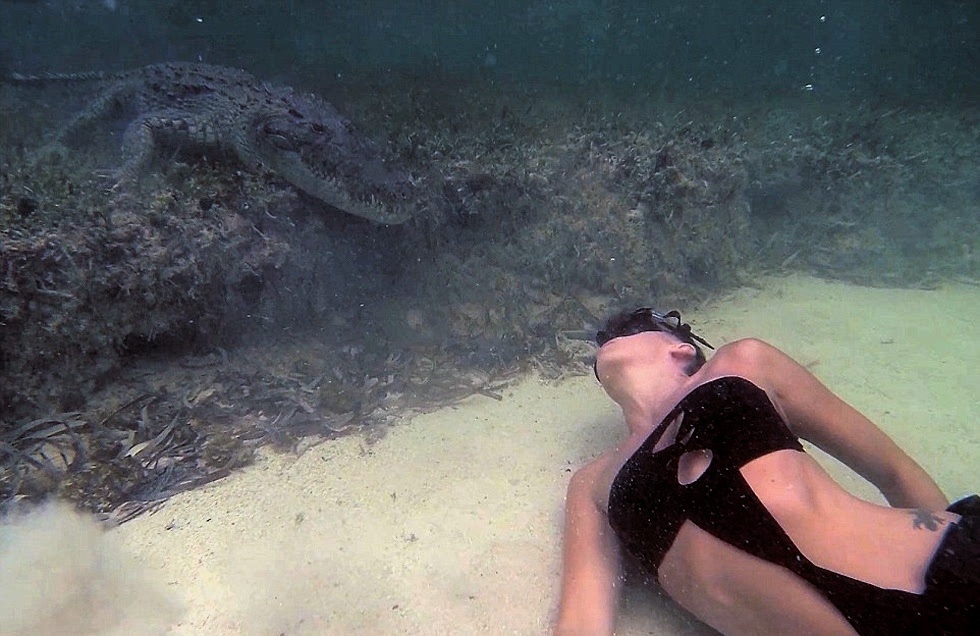} &
        \includegraphics[width=1.69cm, height=1.06cm]{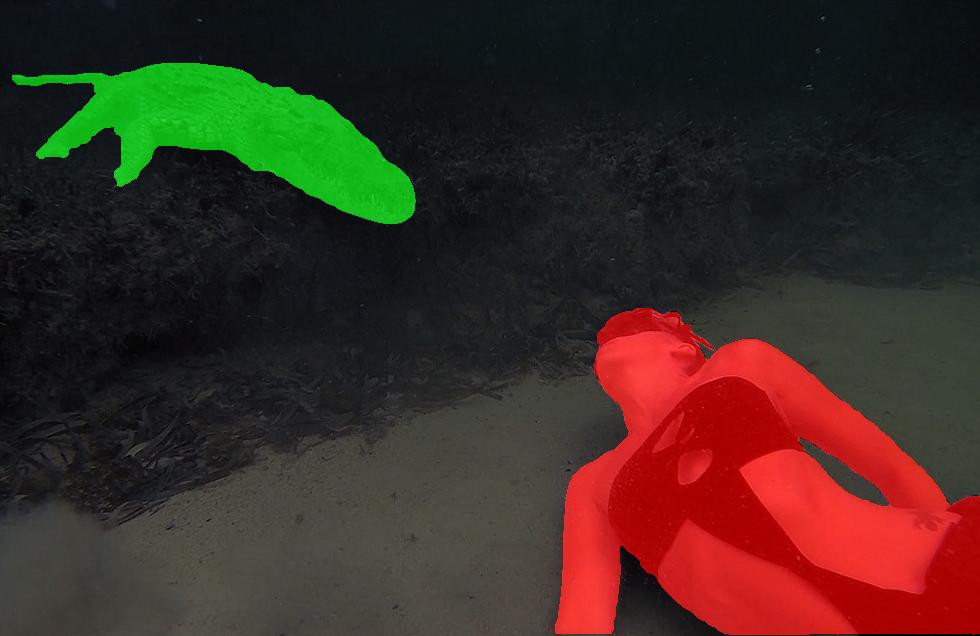} &
        \includegraphics[width=1.69cm, height=1.06cm]{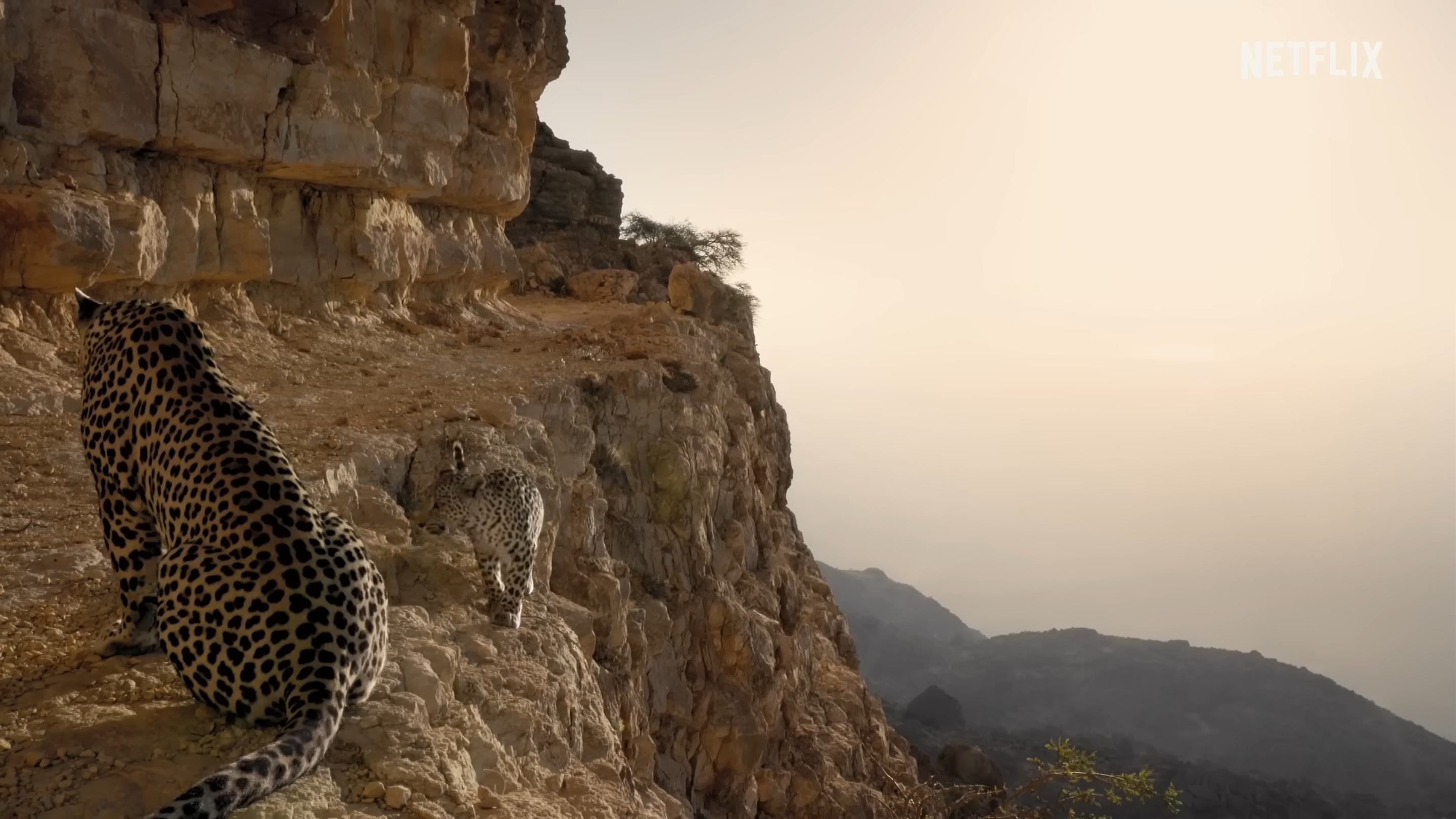} &
        \includegraphics[width=1.69cm, height=1.06cm]{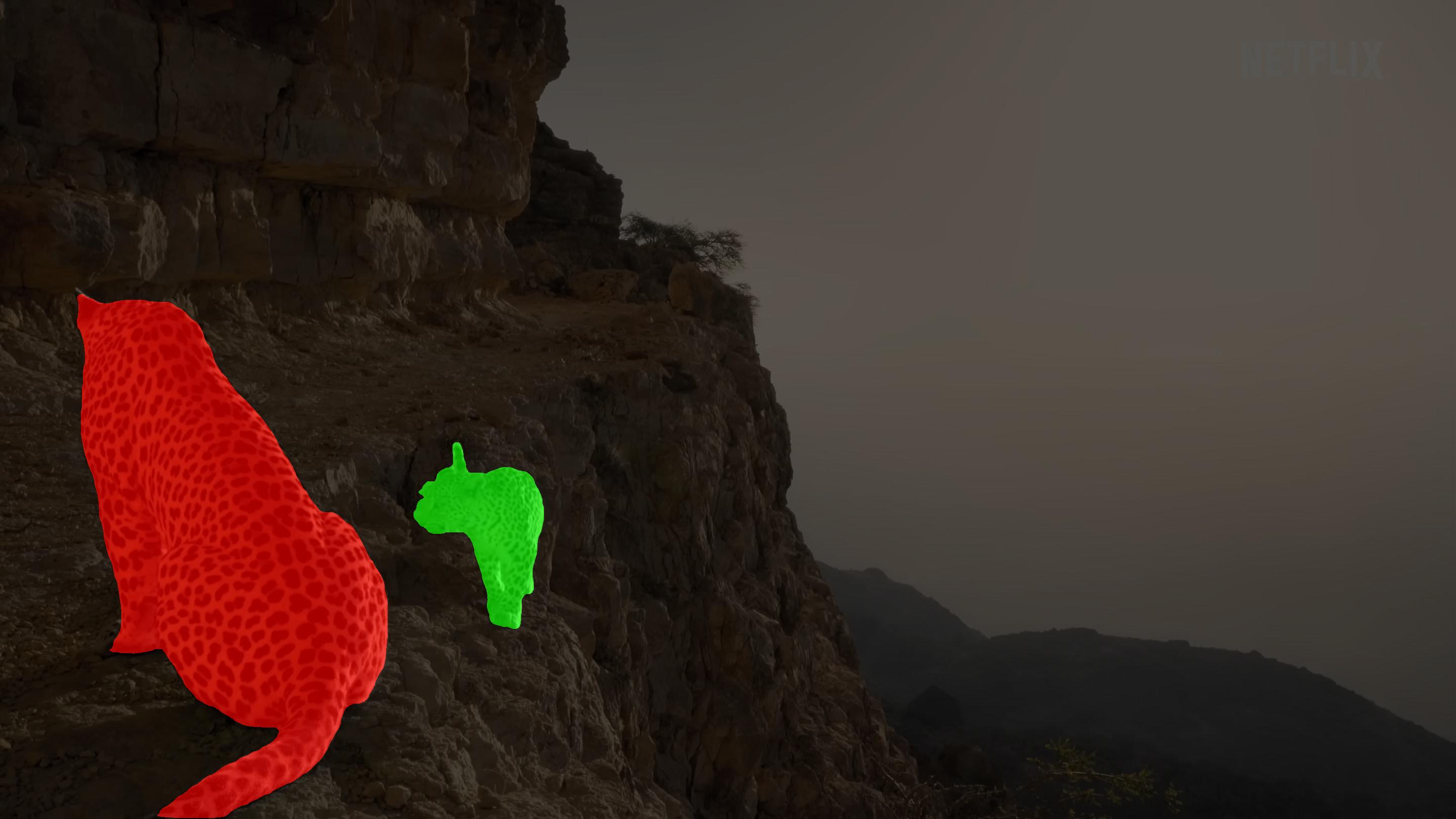} &
        \includegraphics[width=1.69cm, height=1.06cm]{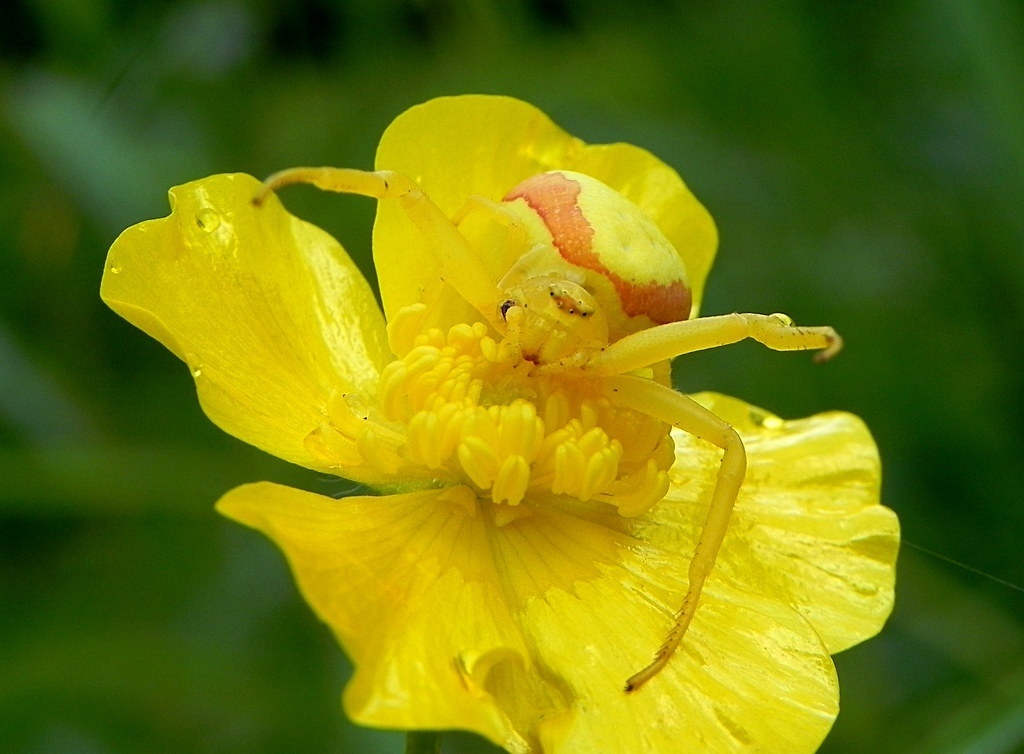} &
        \includegraphics[width=1.69cm, height=1.06cm]{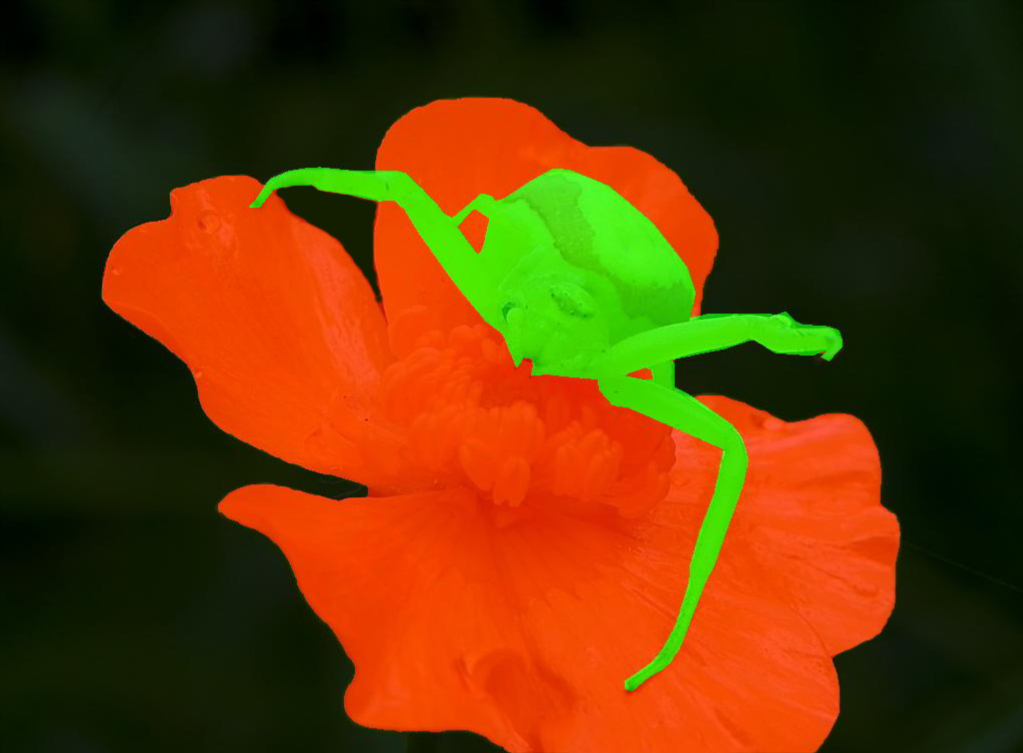} &
        \includegraphics[width=1.69cm, height=1.06cm]{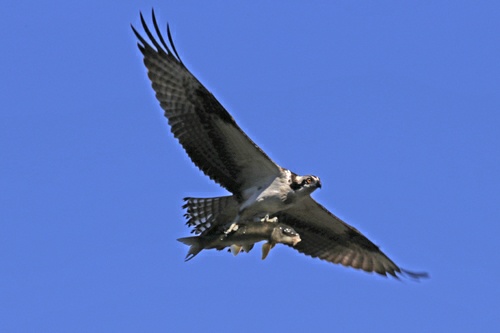} &
        \includegraphics[width=1.69cm, height=1.06cm]{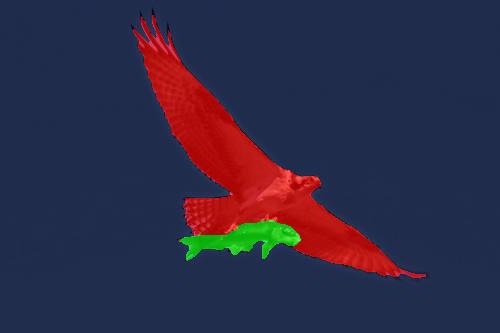} &
        \includegraphics[width=1.69cm, height=1.06cm]{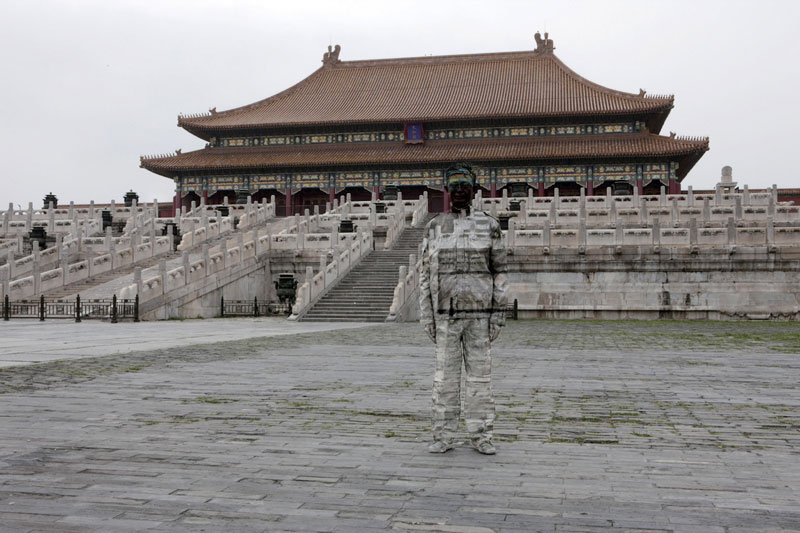} &
        \includegraphics[width=1.69cm, height=1.06cm]{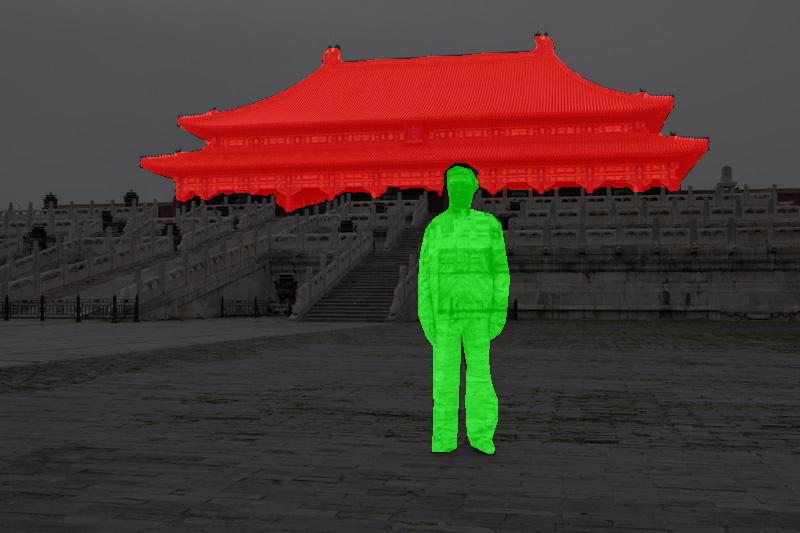} \\

        \raisebox{0.1cm}{\makebox[0pt][c]{\rotatebox{90}{\footnotesize \textit{Scene D}}}} &
        \includegraphics[width=1.69cm, height=1.06cm]{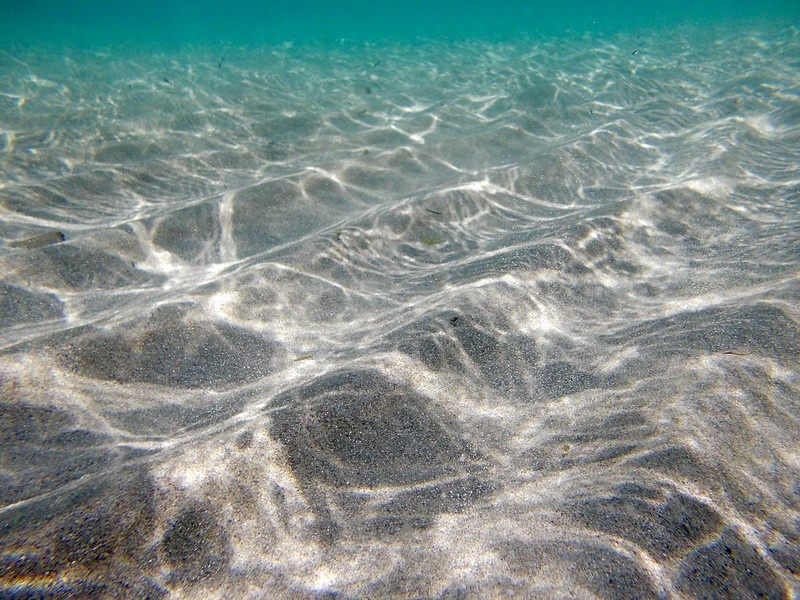} &
        \includegraphics[width=1.69cm, height=1.06cm]{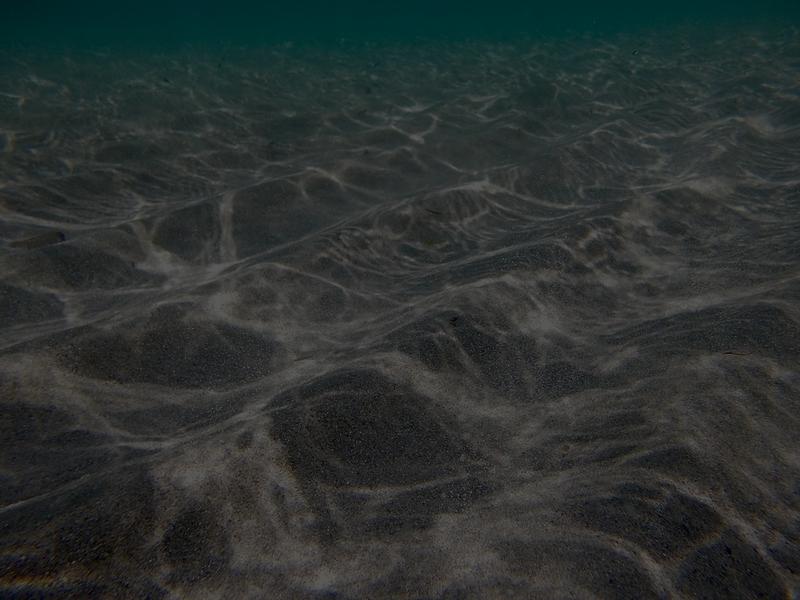} &
        \includegraphics[width=1.69cm, height=1.06cm]{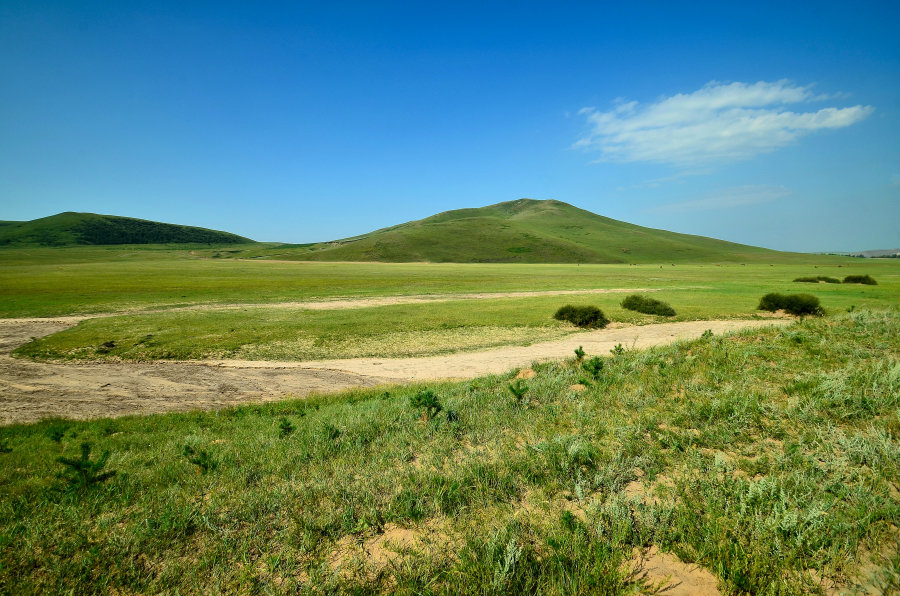} &
        \includegraphics[width=1.69cm, height=1.06cm]{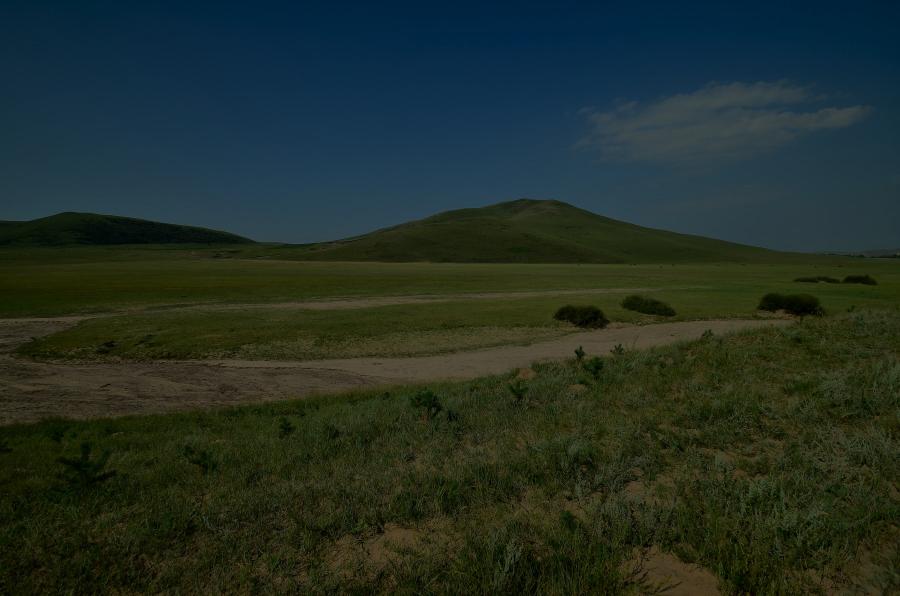} &
        \includegraphics[width=1.69cm, height=1.06cm]{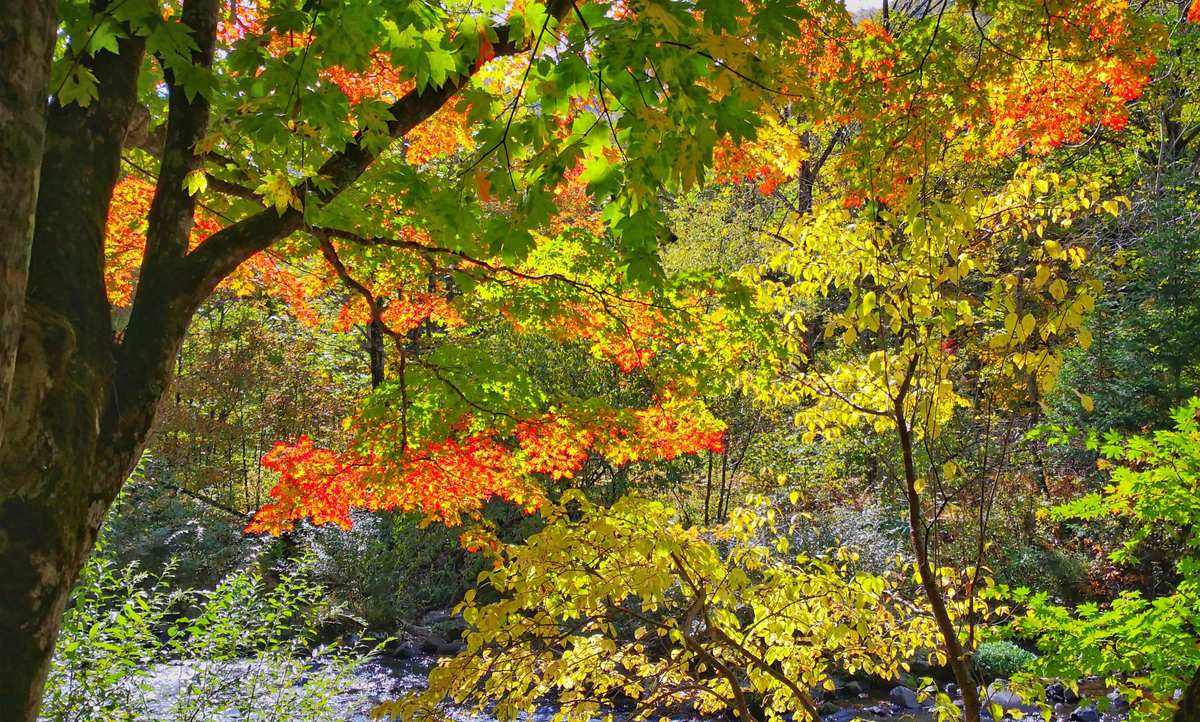} &
        \includegraphics[width=1.69cm, height=1.06cm]{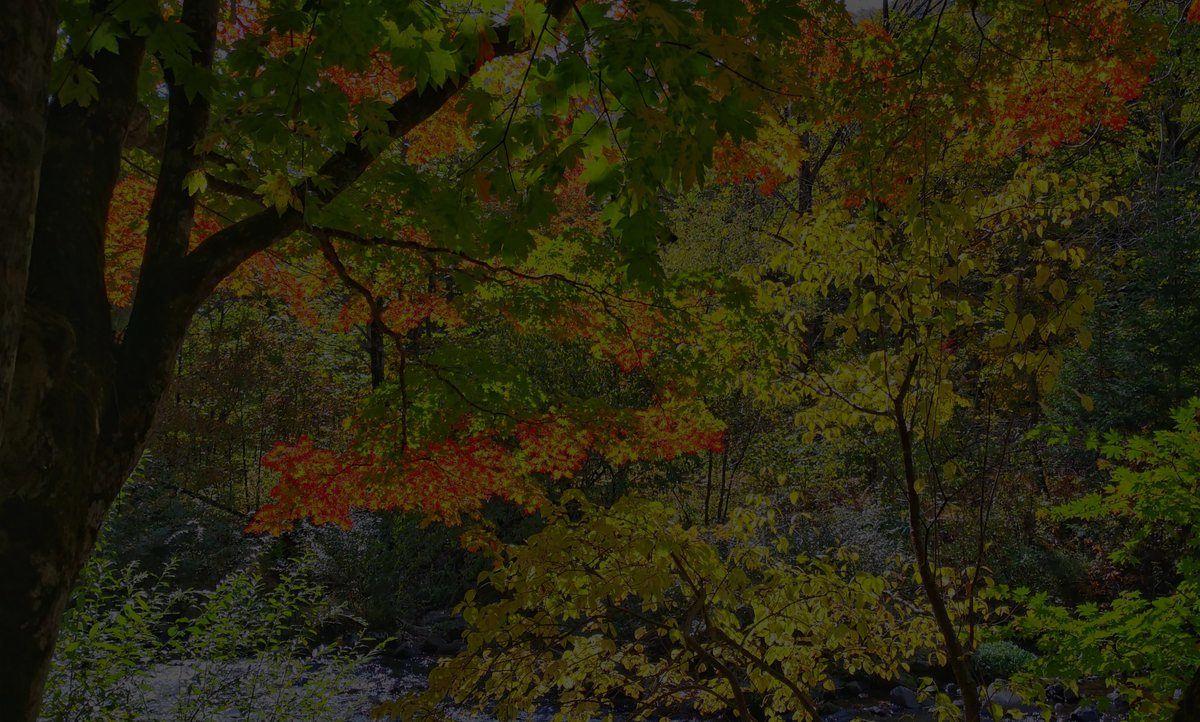} &
        \includegraphics[width=1.69cm, height=1.06cm]{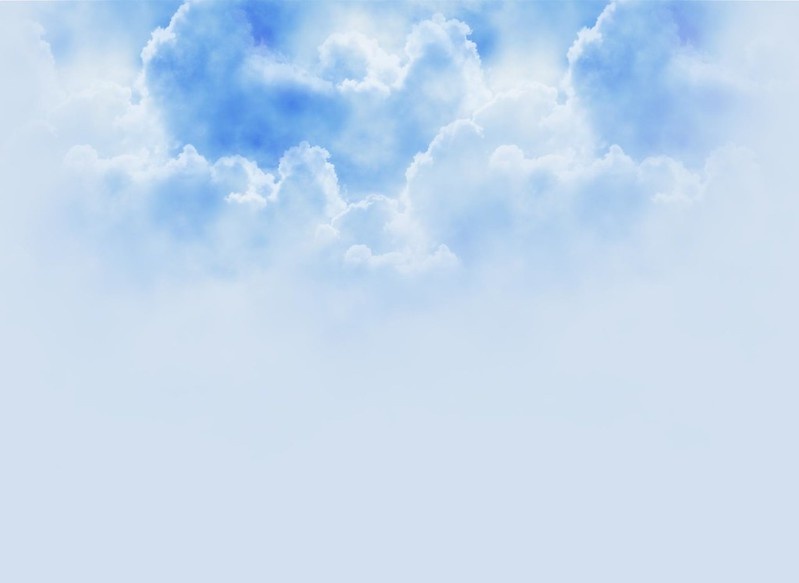} &
        \includegraphics[width=1.69cm, height=1.06cm]{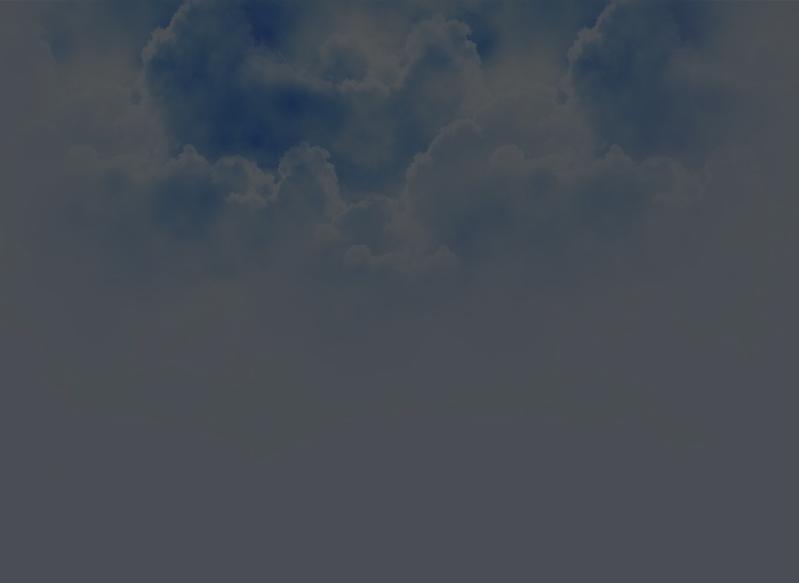} &
        \includegraphics[width=1.69cm, height=1.06cm]{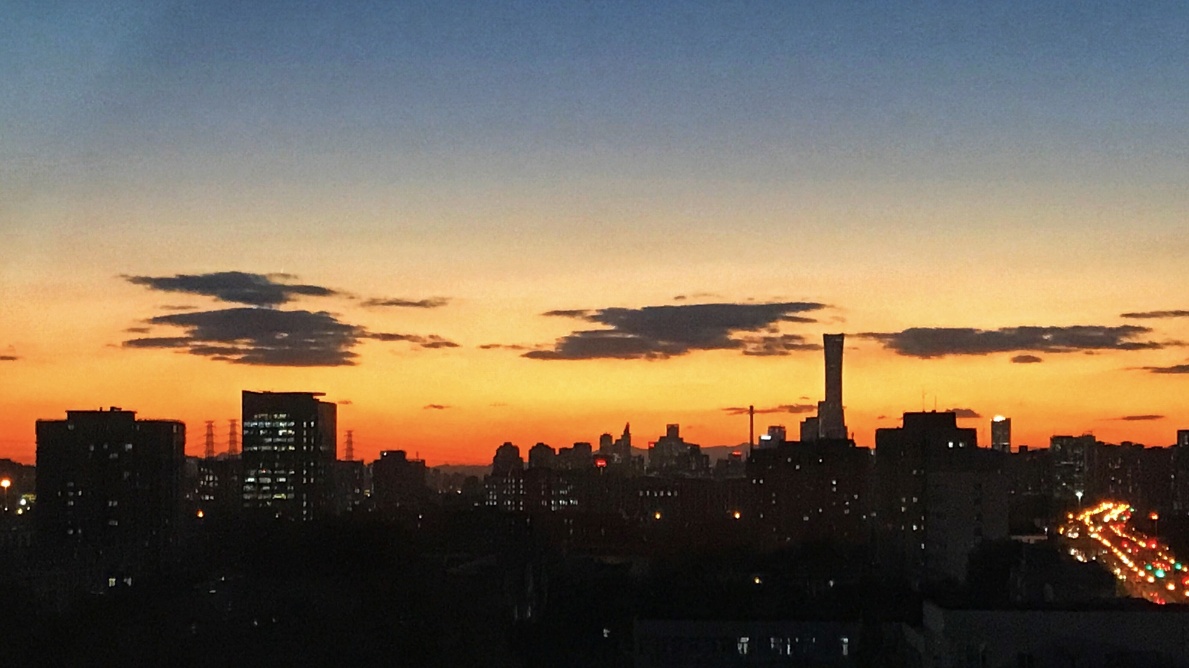} &
        \includegraphics[width=1.69cm, height=1.06cm]{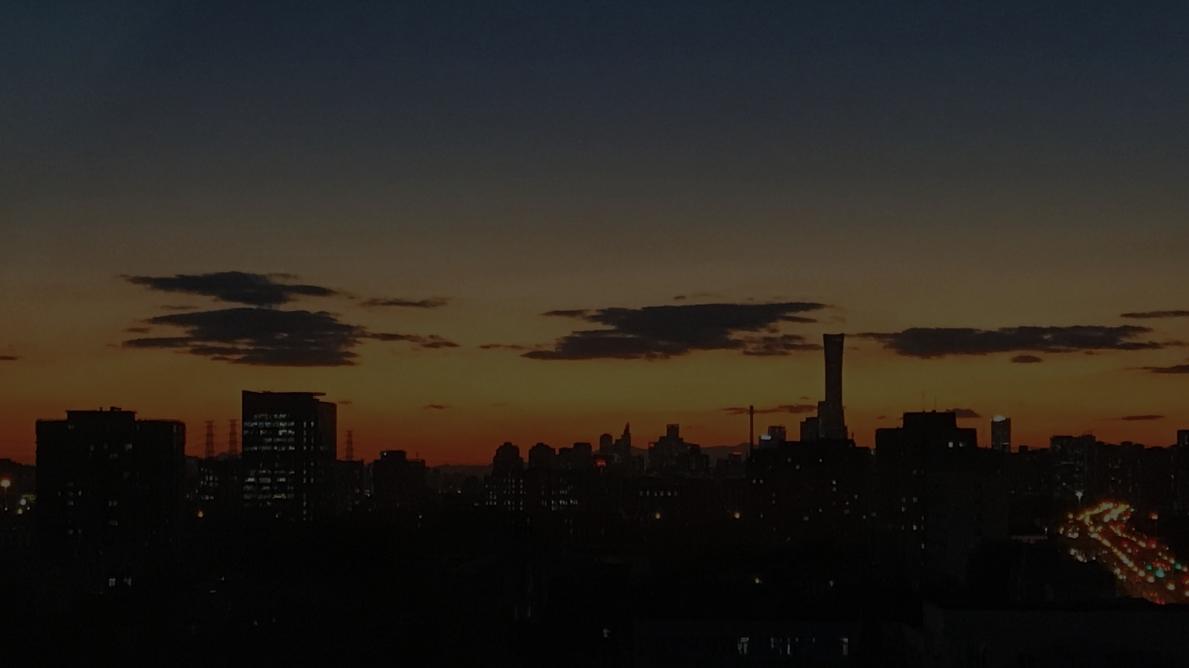} \\

        & \makebox[1.69cm][c]{Image} & \makebox[1.69cm][c]{GT} & \makebox[1.69cm][c]{Image} & \makebox[1.69cm][c]{GT}& \makebox[1.69cm][c]{Image} & \makebox[1.69cm][c]{GT} & 
        \makebox[1.69cm][c]{Image} & \makebox[1.69cm][c]{GT} & \makebox[1.69cm][c]{Image} & \makebox[1.69cm][c]{GT} \\
        
    \end{tabular}
    \caption{Example images from the USC12K dataset: Scene A: Only camouflaged object. Scene B: Only salient object. Scene C: Both salient and camouflaged objects simultaneously. Scene D: Background, none of them. More examples can be found in \textcolor[rgb]{0.8588, 0.2666, 0.2156}{Appendix.\S 5}.}
    
    \label{fig:cs12k_example}
\end{figure*}

\subsection{Data Annotation}
For the data with finalized scene classification, we retain the original labels for Scene A and B. For Scene C, we use SAM~\cite{kirillov2023segment} for coarse annotation, followed by manual refinement, to supplement mask labels, including single-attribute images and web-collected data.
During category annotation, we set all pixels outside the mask of the entire dataset to 0. We obtain an initial coarse classification using CLIP~\cite{radford2021learning}, followed by manual verification and refinement. Except for images collected from COD10K~\cite{fan2020camouflaged}, which already include camouflage object category labels, all other objects require classification. Some examples of different scenes from our USC12K dataset are shown in Figure~\ref{fig:cs12k_example}. Then we assign category labels to each image, covering 9 super-classes and 179 sub-classes. Figure \textcolor{iccvblue}{3} in {\textcolor[rgb]{0.8588, 0.2666, 0.2156}{Appendix. \S 5}} illustrates the class breakdown of our USC12K dataset.

\subsection{Data Analysis}\label{sec:DA}
For deeper insights into SOD and COD datasets, we compare our \ourdataset~against 13 other related datasets including: (1) nine SOD datasets: SOD~\cite{movahedi2010design}, PASCAL-S~\cite{li2014secrets}, ECSSD~\cite{yan2013hierarchical},
HKU-IS~\cite{li2015visual}, MSRA-B~\cite{DBLP:journals/pami/LiuYSWZTS11}, DUT-OMRON~\cite{yang2013saliency}, MSRA10K~\cite{DBLP:journals/pami/ChengMHTH15}, DUTS~\cite{wang2017learning}, and SOC~\cite{fan2018SOC}; (2) four COD datasets: CAMO~\cite{le2019anabranch}, CHAMELEON~\cite{chameleon}, COD10K~\cite{fan2020camouflaged}, and NC4K~\cite{lv2021simultaneously}; 
Table \ref{tab:dataset_table1} shows the detailed information of these datasets. It can be seen that except for COD10K, all SOD datasets only contain salient objects, and nearly all COD datasets only contain camouflaged objects. The scenarios covered by these datasets are relatively limited. Notably, although the COD dataset COD10K contains some images with salient objects and images without any objects, these images lack labels and are not included in the training process. In contrast, the USC12K dataset we propose imposes no restrictions on scenes and includes labels for three attributes: saliency, camouflage, and background, with a well-balanced distribution.

\section{The Proposed \ourmodel~Baseline}

\textbf{Overview.} 
As illustrated in Figure~\ref{fig:our-APG}, the main components of the proposed~\ourmodel~include: (1) A SAM image encoder to extract object feature representation with adapter layers. (2) An Attribute Relation Modeling (ARM) module that generates three attribute prompts (saliency, camouflaged, background), enables interactions within the module, and introduces two distinct prompt query mechanisms to model attribute relationships at both inter-sample and intra-sample levels. (3) A frozen mask decoder of SAM that is applied to predict the final saliency, camouflage, and background masks based on different attribute prompts.


\subsection{SAM Encoder with Adapter}
\begin{figure*}[tb]
  \centering
  \includegraphics[width=0.98\textwidth]{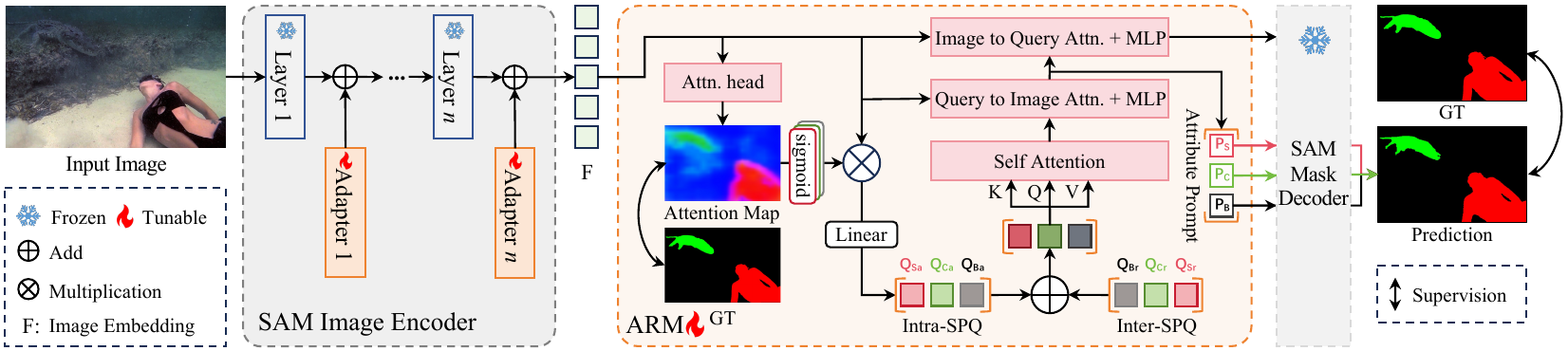}
  \caption{Architecture of our~\ourmodel.~\ourmodel~includes: SAM image encoder with adapter, ARM module, and frozen SAM mask decoder.}
  \label{fig:our-APG}
\end{figure*}

SAM~\cite{kirillov2023segment} consists of an image encoder, a prompt encoder, and a mask decoder. In~\ourmodel, we utilize the prompt architecture of SAM for identifying three attributes: saliency, camouflage, and background. The attribute prompts are generated by a designed ARM module, eliminating the need for the manual prompt. Each attribute prompt is mapped to a distinct binary mask. Following SAM-Adapter~\cite{chen2023sam}, \ourmodel~integrates adapters into each layer of the SAM encoder with a parameter-efficient fine-tuning approach, as depicted in Figure~\ref{fig:our-APG}. Through this approach, \ourmodel~blending SOD- and COD-specific knowledge with the general knowledge acquired by the larger model, better adapts to unconstrained scenes.

\subsection{Attribute Relation Modeling}

To elucidate the intricate relationship between saliency and camouflage in unconstrained scenarios, we introduce the Attribute Relation Modeling (ARM) module.

Our insight is that modeling the relationship of salient and camouflaged objects in unconstrained scenes necessitates the consideration of interactions across two dimensions: (i) Inter-sample relationship modeling: Salient and camouflaged objects often exhibit distinct features across samples. Modeling their inter-sample relationships helps identify universal discriminative features, such as size, location, color, and texture, crucial for distinguishing between them. (ii) Intra-sample relationship modeling: In complex scenarios where salient and camouflaged objects coexist, inter-sample modeling alone may be inadequate. For example, when salient and camouflaged objects share similar colors or categories, intra-sample contextual information becomes essential for accurate differentiation. Building upon this foundation, we introduce ARM, a module that synergistically integrates both Inter-Sample Prompt Query (Inter-SPQ) and Intra-Sample Prompt Query (Intra-SPQ) to discriminatively query attribute-specific prompts across different samples and within individual samples, respectively. The Inter-SPQ is designed to extract generic features between samples, capturing attributes that are universally common across all samples,while the Intra-SPQ concentrates on extracting features within individual samples, emphasizing the distinctive contextual information inherent to each specific sample.

The Inter-SPQ consists of a set of learnable query embeddings $Q_{S_r}\in\mathbb{R}^{N\times C}$, $Q_{C_r}\in\mathbb{R}^{N\times C}$, and $Q_{B_r}\in\mathbb{R}^{N\times C}$, where $N$ represents the number of queries, and $C$ represents the dimensionality of the queries. The Inter-SPQ remains unchanged during inference, regardless of the specific sample. The generation of Intra-SPQ proceeds as follows: Initially, the feature $F$ is extracted from the encoder and passed through an attention head composed of two 3 $\times$ 3 convolutional layers to generate an attention map. The attention map is supervised by the ground truth to encourage it to focus on the salient and camouflaged regions within the image. Subsequently, the sigmoid function is applied to each channel of the attention map. This processed attention map is element-wise multiplied with the original feature $F$ to isolate attribute-specific features. These features are subsequently passed through a linear layer to perform downsampling, generating the Intra-SPQ, which ensures dimensionality consistency with the Inter-SPQ. The Intra-SPQ changes during inference based on variations in the sample. The Intra-SPQ generation process can be described by the following formula:
\begin{equation}\small
[Q_{S_a}, Q_{C_a}, Q_{B_a}] = Linear(\sigma(\Phi_{AH}(F)) \otimes F),
\end{equation}
where \(Q_{S_a}\), \(Q_{C_a}\), and \(Q_{B_a}\) represent the Intra-SPQ for saliency, camouflage, and background, respectively. \(\sigma\) denotes the sigmoid function, and \(\Phi_{AH}\) represents the attention head. The \(\otimes\) denotes element-wise multiplication. 

The Intra-SPQ captures feature information from specific images, whereas the Inter-SPQ discerns the fundamental differences among attributes. By integrating these two components through summation, our ARM is endowed with the capability to handle unconstrained scenarios, regardless of the presence of salient or camouflaged objects. Subsequently, we employ self-attention ($SA$) to establish relationships between queries and query-to-image (\(Q2I\)) attention to interact with the image embedding $F$, ultimately generating attribute prompts: \( P = \left\{P_{S} \in \mathbb{R}^{N \times C},\; P_{C} \in \mathbb{R}^{N \times C},\; P_{B} \in \mathbb{R}^{N \times C}\right\} \). The process can be formulated as follows:
\begin{equation}\small
P= MLP(Q2I(SA(\text{Intra-SPQ} + \text{Inter-SPQ}), F)),
\end{equation}
where \(Q2I\) denotes the cross-attention from queries to the image embedding \(F\),
enabling the model to focus on relevant parts of the input based on the queries. Furthermore, we use image-to-query (\(I2Q\)) attention to focus on features related to attributes. Based on the three attribute-specific prompts fed into the pre-trained mask decoder in SAM, three masks—\(M_S\), \(M_C\), and \(M_B\)—are obtained, representing the predicted saliency, camouflage, and background, respectively.
The process can be described as:
\begin{equation}\small
[M_S,M_C, M_B] = MaskDe([P_{S}, P_{C}, P_{B}], F),
\end{equation}
where $MaskDe$ denotes the frozen SAM mask decoder. Finally, a softmax function is applied to the three masks to generate the final prediction.

\subsection{Loss Function}
We use the ground truth to supervise the final prediction and attention map. The total loss
function of~\ourmodel~can be defined as:
\begin{equation}\small
\label{Loss_total}
   \mathcal{L}_{Total} = \lambda_{p}\mathcal{L}_{pred.} + \lambda_{a}\mathcal{L}_{att.},
\end{equation}
where $\mathcal{L}_{pred.}$ represents the loss of the prediction map and $\mathcal{L}_{att.}$ indicates the loss of the attention map in the ARM. Both of them are calculated using the Focal loss~\cite{lin2017focal}:
\begin{equation}\small
\label{Loss_focal}
   \mathcal{L}_{focal} = -\frac{1}{N} \sum_{i=1}^{N} \alpha_{t_i} (1 - p_{t_i})^\gamma \log(p_{t_i}),
\end{equation}
where $p_{t_i}$ is the predicted probability for the target class $t_i$, $\alpha_{t_i}$ is the weighting factor for class $t_i$ (empirically set to 1:4:6 for background, salient, and camouflaged pixels based on their pixel count ratio), $\gamma$ is the default value of 2 to account for the varying difficulty of different objects, and $N$ is the number of samples.
Besides, $\lambda_{p}$, and $\lambda_{a}$ are empirically set to 1 and 0.5 respectively to balance the total loss.

\section{\ourdataset~Benchmark}
\begin{table*}[!t]
\begin{center}
\caption{Quantitative comparisons with 21 related methods for SOD and COD. $\text{IoU}_S$$\uparrow$: IoU score for salient objects. $\text{IoU}_C$$\uparrow$: IoU score for camouflaged objects. The best two scores are highlighted in \textcolor[RGB]{219, 68, 55}{\textbf{red}} and \textcolor[RGB]{15, 157, 88}{\textbf{green}}, respectively. All metrics presented in the table are expressed as percentages (\%). We use mIoU$\uparrow$, mAcc$\uparrow$, and CSCS$\downarrow$ to evaluate the models in overall scenes.}\label{tab:my_label_sota}
    \scriptsize
    \setlength\tabcolsep{100pt}
    \renewcommand{\arraystretch}{1.07}
    \renewcommand{\tabcolsep}{2.6mm}
\begin{tabular}{c|l|l|c|c|c|cc|cc|ccc}
\toprule
&&&\ \textbf{Update} &\textbf{Scene A}&  \textbf{Scene B}&\multicolumn{2}{c|}{\textbf{Scene C}}&\multicolumn{5}{c}{\textbf{Overall Scenes}}\\
\cline{5-13} 

\multirow{-2}{*}{\textbf{Task}} &
\multirow{-2}{*}{\textbf{Model}} & 
\multirow{-2}{*}{\textbf{Venue}} &
\textbf{Para.(M)} &
\multicolumn{1}{c|}{$\text{IoU}_S$} & 
\multicolumn{1}{c|}{$\text{IoU}_C$} & 
\multicolumn{1}{c|}{$\text{IoU}_S$} & 
\multicolumn{1}{c|}{$\text{IoU}_C$} & 
\multicolumn{1}{c|}{$\text{IoU}_S$} & 
\multicolumn{1}{c|}{$\text{IoU}_C$} & 
\multirow{1}{*}{mIoU} &
\multirow{1}{*}{mAcc} &
\multirow{1}{*}{CSCS}
  \\

\hline\hline

& GateNet~\cite{zhao2020suppress} & ECCV'20& 128&68.32 & 54.26 & 66.85 & 35.03& 65.08 & 44.17& 68.27& 78.07  & 11.30   \\
 &F3Net~\cite{wei2020f3net}& AAAI'20& 26&70.05 & 52.62 & 67.20 & 36.38& 66.12 & 44.81& 68.80& 77.86  & 9.36 \\
& MSFNet~\cite{zhang2021auto}& MM'21& 28&70.14 & 54.78 & 69.92 & 36.64& 66.69& 45.89& 69.40& 79.77  & 9.90    \\
& VST~\cite{liu2021visual} & ICCV'21  & 43&68.14 & 49.82 & 61.61 & 22.56& 63.18& 38.45& 65.55  & 74.77  & 11.30   \\
\multirow{-2}{*}{\textbf{SOD}}
&EDN~\cite{wu2022edn} &TIP'22 & 43&71.59 & 57.94 & 69.37 & 37.70& 68.00& 48.27& 70.70  & 80.60  & 9.23 \\

& PGNet~\cite{xie2022pyramid} & CVPR'22 & 73& 74.69&	57.31& 71.94& 37.21&  70.72& 	48.78& 	71.82& 	80.76& 7.71  \\

& ICON~\cite{zhuge2022salient} & TPAMI'22 & 32&68.09 & 50.57 & 67.48 & 30.65& 65.86& 45.53& 68.99  & 79.53  & 10.24  \\

& MDSAM~\cite{gao2024multi} & MM'24 & 14&72.96 & 56.16 & 67.21 & 36.06& 69.67& 	49.05& 	71.67& 	82.92&  10.21  \\

\cline{1-13}
& SINet-V2~\cite{fan2021concealed}& TPAMI'21 & 27&72.96 & 56.16 & 67.21 & 36.06& 69.50& 47.47 & 70.20  & 79.58  &  8.83\\
& PFNet~\cite{mei2021camouflaged}& CVPR'21  & 47&69.07 & 52.83 & 67.20 & 32.81& 65.73& 43.76& 68.30& 78.00& 10.04 \\
& ZoomNet~\cite{pang2022zoom}& CVPR'22  & 33&74.11 & 51.12 & 66.79 & 29.69& 66.43& 43.28& 68.35  & 77.72  & 8.88\\
\multirow{5}{*}{\textbf{COD}}

& FEDER~\cite{he2023camouflaged}& CVPR'23  & 44&74.35 & 58.04 & 67.66 & 32.26& 68.65 & 46.46 & 70.32& 81.27& 10.01\\   
& PRNet~\cite{hu2024efficient}& TCSVT'24  & 13&76.10 & 61.54 & 60.10& 32.16&  68.68 & 50.88 & 71.87 & 82.89  & 8.40\\

& ICEG~\cite{he2024strategic}& ICLR'24 & 100& 73.67 & 68.38 & 68.43 & 44.33& 69.22& 58.71& 74.68  & 83.53  & 8.16\\

& CamoDiffusion~\cite{chen2023camodiffusion}&
AAAI'24  &
72&
75.01&
59.39 &
53.49&
 \textcolor[RGB]{15, 157, 88}{\textbf{45.03}}&
63.49 &
52.80 & 
70.70 &
77.73  &
7.73\\
& CamoFormer~\cite{yin2024camoformer}&
TPAMI'24  &
71&
75.88 &
66.19 & 
\textcolor[RGB]{15, 157, 88}{\textbf{73.33}}& 
44.14&
\textcolor[RGB]{15, 157, 88}{\textbf{71.86}} &
56.09 &
74.81 &
84.17  &
\textcolor[RGB]{15, 157, 88}{\textbf{7.57}}\\
& PGT~\cite{wang2024camouflaged}&
 CVIU'24  &
 68&
 72.75 &
  61.51 &
 70.01&
 41.21&
 71.46 &
\textcolor[RGB]{15, 157, 88}{\textbf{56.83}} &
 \textcolor[RGB]{15, 157, 88}{\textbf{75.03}} &
83.35  &
  9.09\\
& SAM-Adapter~\cite{chen2023sam}&
 ICCVW'23  &
 \textcolor[RGB]{15, 157, 88}{\textbf{4.11}}&
 \textcolor[RGB]{15, 157, 88}{\textbf{78.90}} &
 67.69 & 
 68.19&
27.73&
 70.66 &
 52.69 &
 73.38 &
 83.35  &
  10.28\\
&SAM2-Adapter~\cite{chen2024sam2} & arXiv'24  & 4.36&78.75&\textcolor[RGB]{15, 157, 88}{\textbf{70.28}}&69.01&38.20&71.42& 56.71& 74.98 & \textcolor[RGB]{15, 157, 88}{\textbf{84.74}}  & 9.12  \\
\cline{1-13}
\multirow{3}{*}{\centering\textbf{Unified}}
&EVP~\cite{liu2023explicit}& CVPR'23  & 4.95&75.85 & 59.81 & 71.41 & 37.64& 70.30& 50.36& 72.16  & 79.96  & 8.67\\

&VSCode~\cite{luo2024vscode}& CVPR'24  &60&76.04 &60.31 &72.58 &39.46 &71.08   &55.14  &74.17&84.01  &8.17   \\

\multirow{1}{*}{\centering }

& \cellcolor{iccvblue!20}\textbf{\ourmodel~(Ours)} & \cellcolor{iccvblue!20}\textbf{-}&
\cellcolor{iccvblue!20}\textcolor[RGB]{219, 68, 55}{\textbf{4.04}} &
\cellcolor{iccvblue!20}\textcolor[RGB]{219, 68, 55}{\textbf{79.70}} & 
\cellcolor{iccvblue!20}\textcolor[RGB]{219, 68, 55}{\textbf{74.99}} & 
\cellcolor{iccvblue!20}\textcolor[RGB]{219, 68, 55}{\textbf{74.80}} & 
\cellcolor{iccvblue!20}\textcolor[RGB]{219, 68, 55}{\textbf{45.73}} & 
\cellcolor{iccvblue!20}\textcolor[RGB]{219, 68, 55}{\textbf{75.57}} & \cellcolor{iccvblue!20}\textcolor[RGB]{219, 68, 55}{\textbf{61.34}} & \cellcolor{iccvblue!20}\textcolor[RGB]{219, 68, 55}{\textbf{78.03}} & \cellcolor{iccvblue!20}\textcolor[RGB]{219, 68, 55}{\textbf{87.92}} & \cellcolor{iccvblue!20}\textcolor[RGB]{219, 68, 55}{\textbf{7.49}} \\
\bottomrule
\end{tabular}
\end{center}
\end{table*}

We evaluate 21 relevant methods on USC12K to establish a comprehensive benchmark. All models are trained and tested on the training set of USC12K (8,400 images) and the test set of USC12K (3,600 images).

\begin{table}[t!]
    \centering
    \caption{After training with USC12K, misinterpretation is largely eliminated. \textbf{Left:} Inference results of SOD models on COD datasets. VSCode uses the SOD prompt. \textbf{Right:} Inference results of COD models on SOD datasets. VSCode uses the COD prompt. The metric is $F_\beta^\omega$. EV refers to the Expected Value.}
    \label{tab:after_train}
    \begin{minipage}{0.5\linewidth}
    \centering
    \scriptsize
    \setlength\tabcolsep{100pt}
    \renewcommand{\arraystretch}{1.19}
    \renewcommand{\tabcolsep}{0.9mm}
    \begin{tabular}{l|c|c|c}
        \toprule
         & \multicolumn{2}{c|}{\textbf{COD Datasets}} \\
        \cline{2-3} 
        \multirow{-2}{*}{\textbf{SOD Models}} & COD10K & NC4K & \multirow{-2}{*}{\textbf{EV}}\\
        \hline \hline
        ICON~\cite{zhuge2022salient} & 0.0146 & 0.0834 & 0\\
        F3Net~\cite{wei2020f3net} & 0.0129 & 0.0787 & 0\\
        VSCode~\cite{luo2024vscode} & 0.0097 & 0.0626 & 0\\
        \bottomrule
    \end{tabular}
    \end{minipage}%
    \begin{minipage}{0.5\linewidth}
    \centering
        \scriptsize
    \setlength\tabcolsep{100pt}
    \renewcommand{\arraystretch}{1.19}
    \renewcommand{\tabcolsep}{0.8mm}
    \begin{tabular}{l|c|c|c}
        \toprule
         & \multicolumn{2}{c|}{\textbf{SOD Datasets}} \\
        \cline{2-3} 
        \multirow{-2}{*}{\textbf{COD Models}} & DUTS & HKU-IS & \multirow{-2}{*}{\textbf{EV}}\\
        \hline \hline
        SINet-V2~\cite{fan2021concealed} & 0.0708 & 0.0443 & 0\\
        PFNet~\cite{mei2021camouflaged} &0.1152  &0.0874  & 0\\
        VSCode~\cite{luo2024vscode} & 0.0537 & 0.0391 & 0\\
        \bottomrule
    \end{tabular}
    \end{minipage}%
\end{table}

\noindent\textbf{Metrics.} USC12K benchmark involves three distinct attributes: saliency, camouflage, and background. We employ three established metrics for semantic segmentation:  mAcc$\uparrow$, IoU$\uparrow$, and mIoU$\uparrow$~\cite{long2015fully,xie2021segformer}. Inspired by~\cite{li2024size}, we also employ AUC$\uparrow$, SI-AUC$\uparrow$, $F_m^{\beta}$$\uparrow$, SI-$F_m^{\beta}$$\uparrow$, $F_{\max}^{\beta}$$\uparrow$, SI-$F_{\max}^{\beta}$$\uparrow$, $E_m$$\uparrow$ to evaluate the model's capability in detecting objects of varying sizes. Additionally, to evaluate the model's ability to distinguish salient from camouflaged objects, we propose a novel metric, the \textbf{C}amouflage-\textbf{S}aliency \textbf{C}onfusion \textbf{S}core \textbf{(CSCS$\downarrow$)}, which is formulated as follows:
\begin{equation}\small
\label{CSCS_equa}
\text{CSCS} = \frac{1}{2} (\frac{\mathcal{P}_{CS}}{\mathcal{P}_{BS} + \mathcal{P}_{SS} + \mathcal{P}_{CS}} + \frac{\mathcal{P}_{SC}}{\mathcal{P}_{BC} + \mathcal{P}_{SC} + \mathcal{P}_{CC}}), 
\end{equation}
where $ \mathbb{P} = \left\{ \mathcal{P}_{\lambda \theta} \,|\, \lambda \in \Theta, \theta \in \Theta \right\},\ \Theta = \left\{B, C, S\right\} $, the B, C and S denote background, camouflage, and saliency, $\mathcal{P}_{CS}$ represents regions where camouflage is predicted as salient, while $\mathcal{P}_{SC}$ represents regions where saliency is predicted as camouflage; both are regions of confusion. A lower CSCS indicates a stronger robustness to distinguish between salient and camouflaged objects. More details of CSCS can be seen in {\textcolor[rgb]{0.8588, 0.2666, 0.2156}{Appendix. \S 1}}.

\noindent\textbf{Competitors.} We compare with 21 related models, including 
(\uppercase\expandafter{\romannumeral1}) SOD models: GateNet~\cite{zhao2020suppress}, F3Net~\cite{wei2020f3net}, MSFNet~\cite{zhang2021auto}, VST~\cite{liu2021visual}, EDN~\cite{wu2022edn}, ICON~\cite{zhuge2022salient}, MDSAM~\cite{gao2024multi}; (\uppercase\expandafter{\romannumeral2}) COD models: SINet-V2~\cite{fan2021concealed}, PFNet~\cite{mei2021camouflaged}, ZoomNet~\cite{pang2022zoom}, FEDER~\cite{he2023camouflaged}, ICEG~\cite{he2024strategic}, PRNet~\cite{hu2024efficient}, CamoDiffusion~\cite{chen2023camodiffusion},
CamoFormer~\cite{yin2024camoformer},
PGT~\cite{wang2024camouflaged},
SAM-Adapter~\cite{chen2023sam} and SAM2-Adapter~\cite{chen2024sam2}; (\uppercase\expandafter{\romannumeral3}) Unified methods: VSCode~\cite{luo2024vscode} and EVP~\cite{liu2023explicit}.

\begin{table}[!t]
\begin{center}
\caption{Generalization performance of six methods on SOD dataset: DUTS, HKU-IS and COD dataset: NC4K and COD10K. More comparisons can be found in {\textcolor[rgb]{0.8588, 0.2666, 0.2156}{Appendix. \S 3}}.}
\label{tab:supp_combined}
\scriptsize

\setlength\tabcolsep{0.7mm}
\renewcommand{\arraystretch}{1.10}

\begin{tabular}{l|ccc|ccc|cc|cc}
\toprule
\multirow{2}{*}{\textbf{Model}} & 
\multicolumn{3}{c|}{\textbf{DUTS}} & 
\multicolumn{3}{c|}{\textbf{HKU-IS}} &
\multicolumn{2}{c|}{\textbf{NC4K}} &
\multicolumn{2}{c}{\textbf{COD10K}} \\
& $F_\beta^\text{max}\uparrow$ & $E_{\phi}^\text{m}\uparrow$ & & 
$F_\beta^\text{max}\uparrow$ & $E_{\phi}^\text{m}\uparrow$ & & 
$F_\beta^\text{max}\uparrow$ & $E_{\phi}^\text{m}\uparrow$ & 
$F_\beta^\text{max}\uparrow$ & $E_{\phi}^\text{m}\uparrow$ \\
\hline \hline
ICON~\cite{zhuge2022salient}    & .679 & .785 & & .814 & .874 & & .540 & .715 & .631 & .752 \\
F3Net~\cite{wei2020f3net}       & .703 & .794 & & .832 & .881 & & .576 & .744 & .661 & .773 \\
\hline
SINet-V2~\cite{fan2021concealed}  & .732 & .821 & & .838 & .884 & & .609 & .763 & .662 & .769 \\
PFNet~\cite{mei2021camouflaged}  & .691 & .790 & & .818 & .876 & & .556 & .730 & .660 & .769 \\
\hline
VSCode~\cite{luo2024vscode}     & .724 & .812 & & .834 & .885 & & .626 & .787 & .684 & .783 \\
\cellcolor{iccvblue!20}\textbf{\ourmodel} & \cellcolor{iccvblue!20}\textbf{.784} & \cellcolor{iccvblue!20}\textbf{.852} & \cellcolor{iccvblue!20}&
\cellcolor{iccvblue!20}\textbf{.844} & \cellcolor{iccvblue!20}\textbf{.886} & \cellcolor{iccvblue!20}&
\cellcolor{iccvblue!20}\textbf{.794} & \cellcolor{iccvblue!20}\textbf{.877} & 
\cellcolor{iccvblue!20}\textbf{.743} & \cellcolor{iccvblue!20}\textbf{.869} \\
\bottomrule
\end{tabular}
\end{center}
\end{table}

\noindent\textbf{Technical Details.} All models are retrained on the USC12K training set at a resolution of 352 $\times$ 352. Horizontal flipping and random cropping are applied for data augmentation. Experiments are conducted on one NVIDIA L40 GPU. The number of trained parameters for all models is detailed in Table \ref{tab:my_label_sota}. We use the hiera-large version of SAM2 following the SAM2-Adapter~\cite{chen2024sam2}. AdamW optimizer with warm-up and linear decay is used. The initial learning rate is set to 0.0001. The batch size is set to 24, and the maximum number of epochs is set to 90. More technical details of all methods can be found in {\textcolor[rgb]{0.8588, 0.2666, 0.2156}{Appendix. \S 4}}.

\begin{figure*}[!t]
    \centering
    \footnotesize
    \renewcommand{\arraystretch}{0.2}
    \begin{tabular}{@{}c@{\hskip 5pt}c@{\hskip 0.9pt}c@{\hskip 0.9pt}c@{\hskip 0.9pt}c@{\hskip 0.9pt}c@{\hskip 0.9pt}c@{\hskip 0.9pt}c@{\hskip 0.9pt}c@{\hskip 0.9pt}c@{}}
        \raisebox{0.05cm}{\makebox[0pt][c]{\rotatebox{90}{\small \textit{Scene A}}}} &
         \includegraphics[width=1.8cm, height=1.2cm]{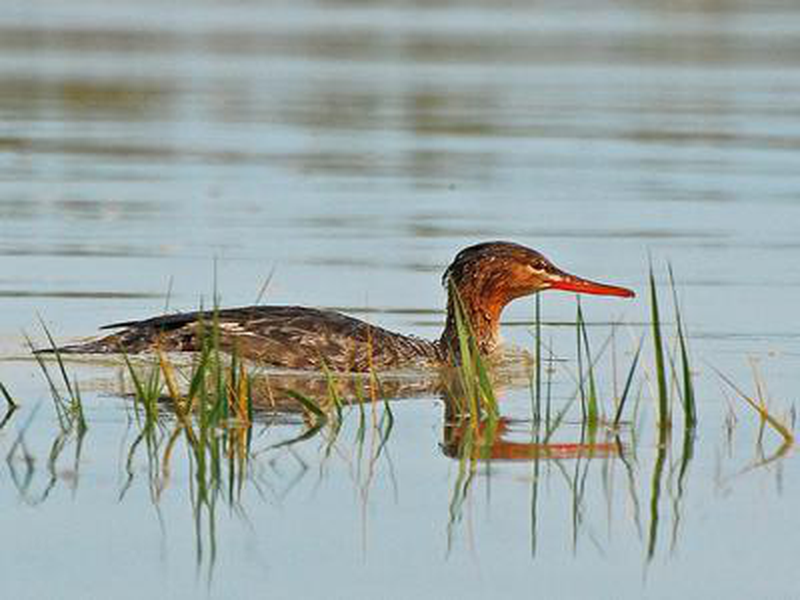} &
         \includegraphics[width=1.8cm, height=1.2cm]{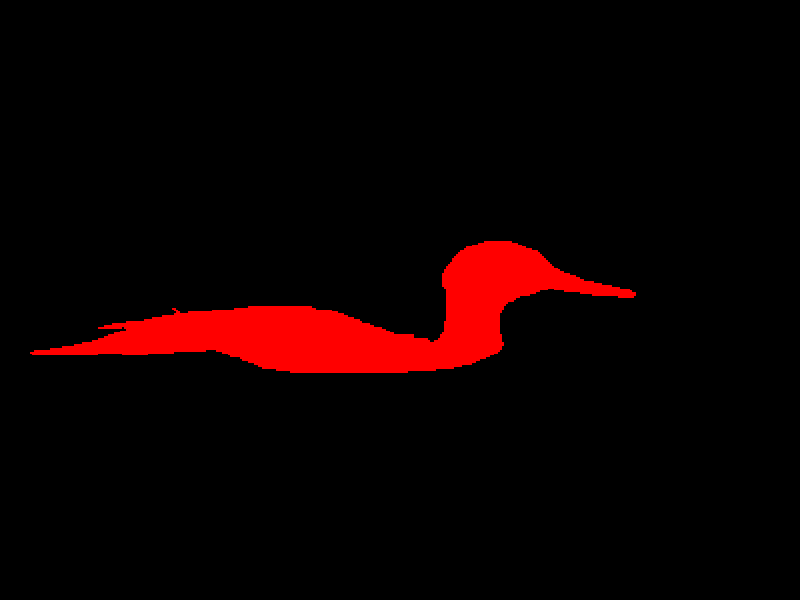} &
         \includegraphics[width=1.8cm, height=1.2cm]{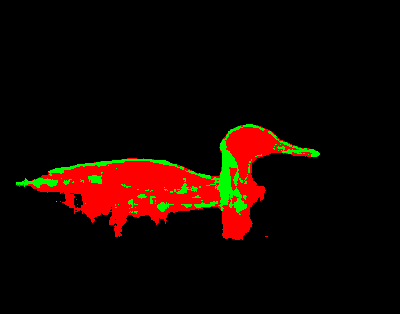} &
         \includegraphics[width=1.8cm, height=1.2cm]{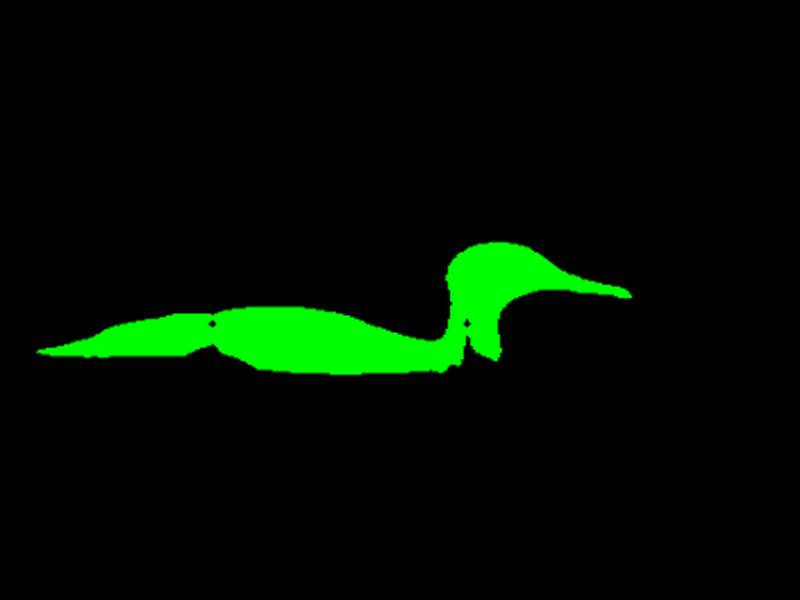} &
         \includegraphics[width=1.8cm, height=1.2cm]{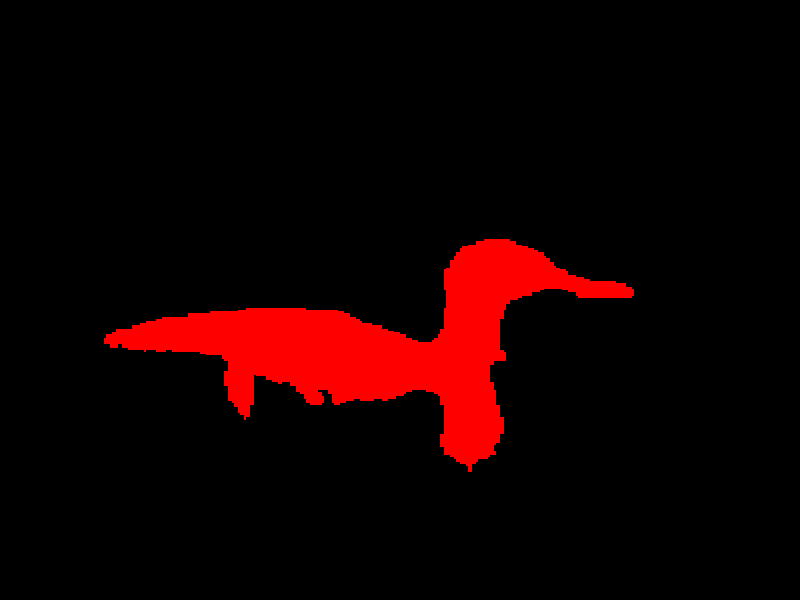} &
         \includegraphics[width=1.8cm, height=1.2cm]{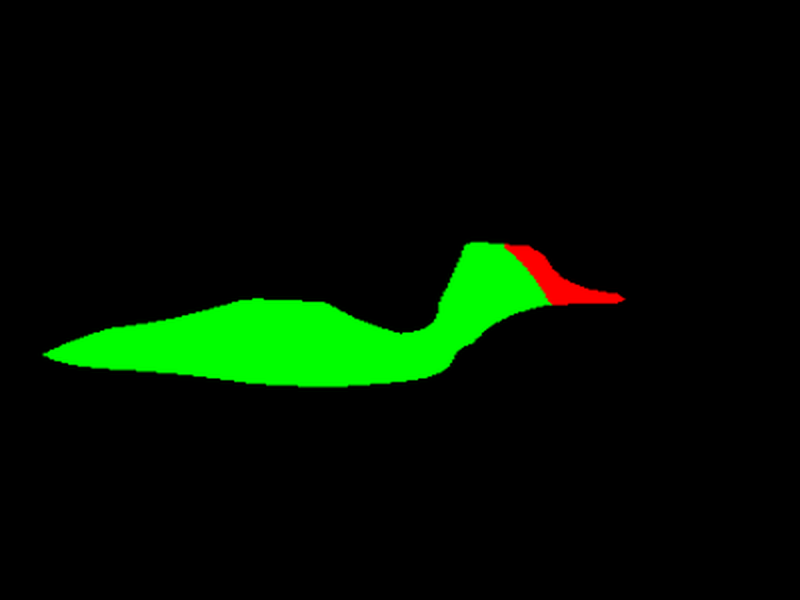} &
         \includegraphics[width=1.8cm, height=1.2cm]{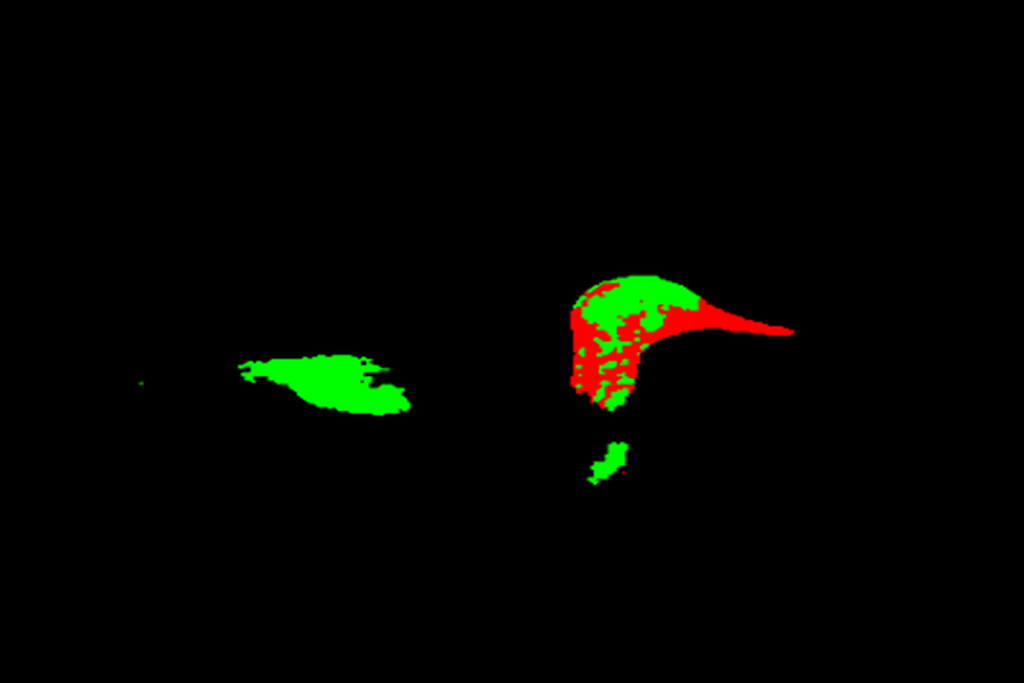}&
         \includegraphics[width=1.8cm, height=1.2cm]{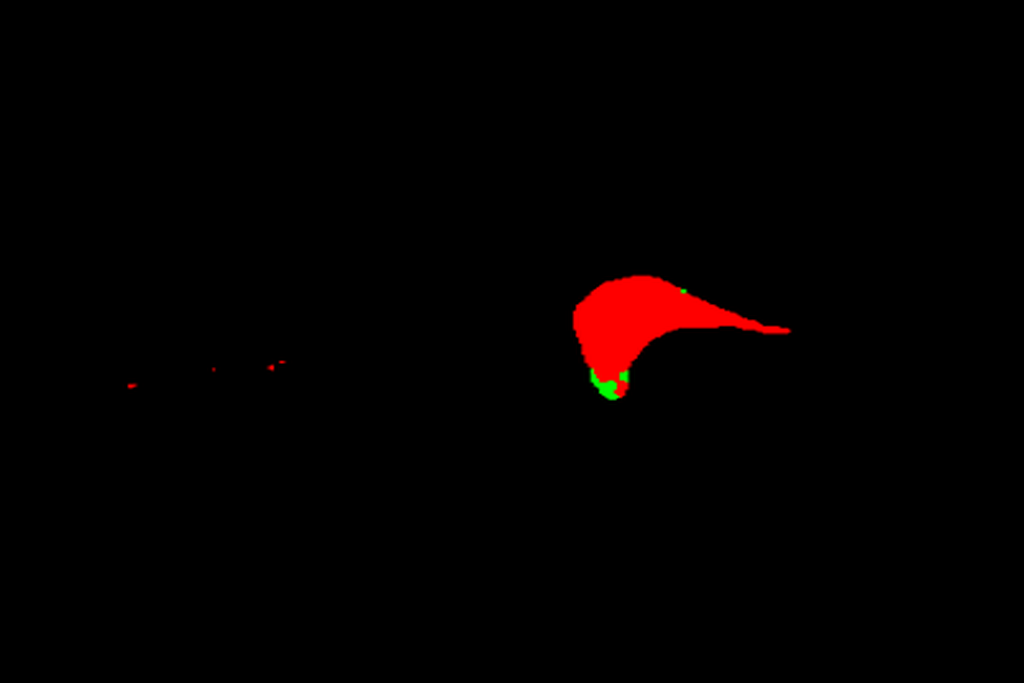}&
         \includegraphics[width=1.8cm, height=1.2cm]{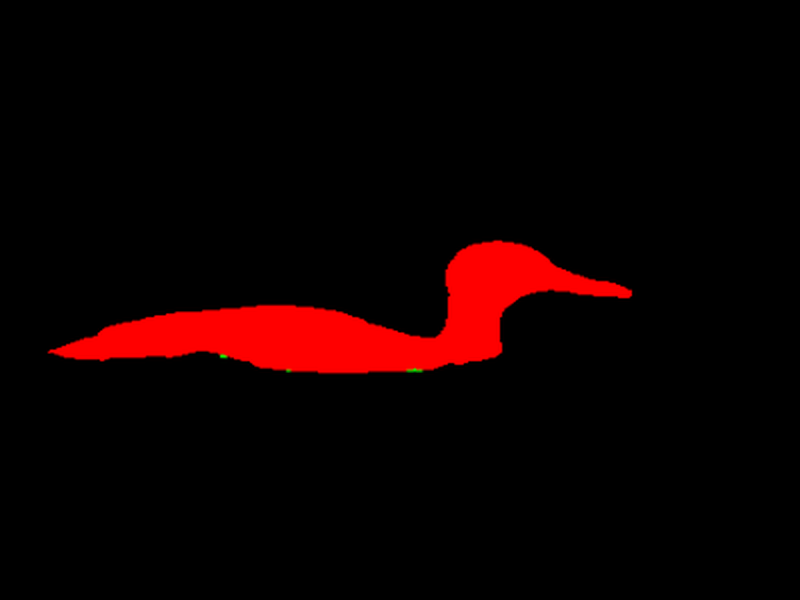} 
        \\

        \raisebox{0.10cm}{\makebox[0pt][c]{\rotatebox{90}{\small \textit{Scene B}}}} &
         \includegraphics[width=1.8cm, height=1.2cm]{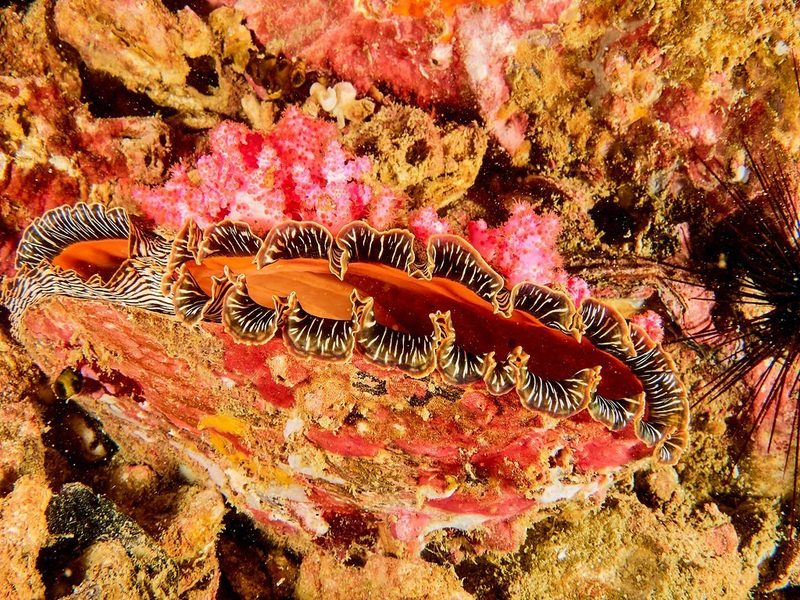} &
         \includegraphics[width=1.8cm, height=1.2cm]{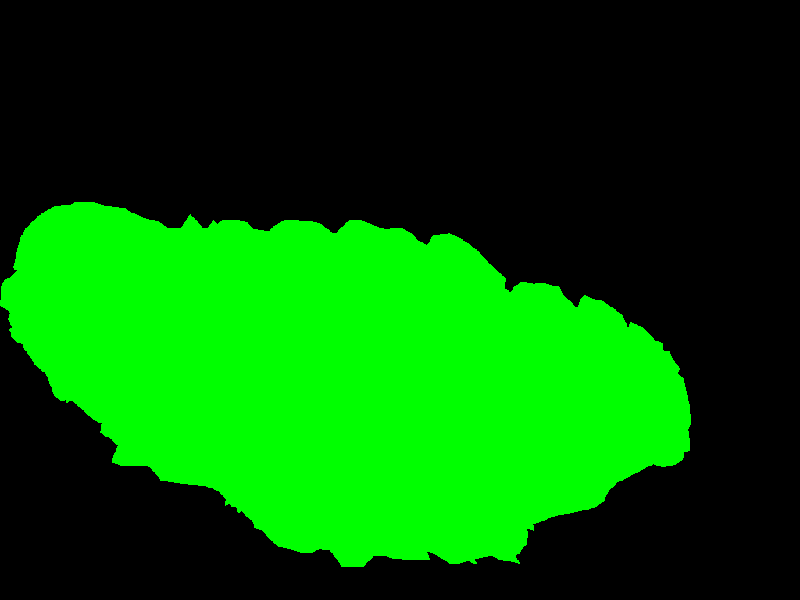} &
         \includegraphics[width=1.8cm, height=1.2cm]{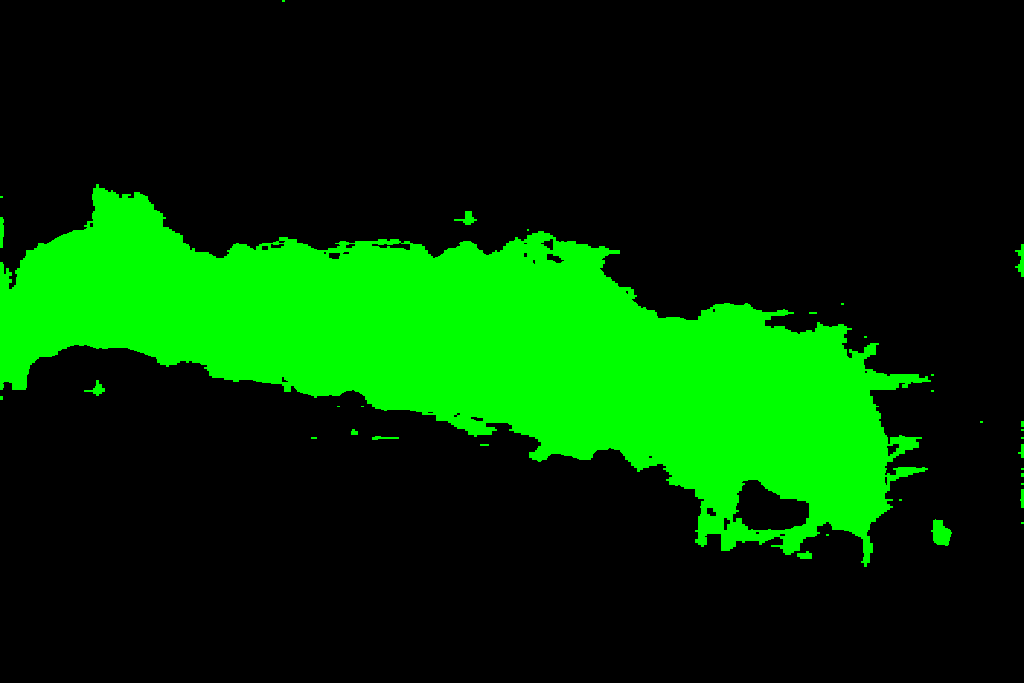} &

         \includegraphics[width=1.8cm, height=1.2cm]{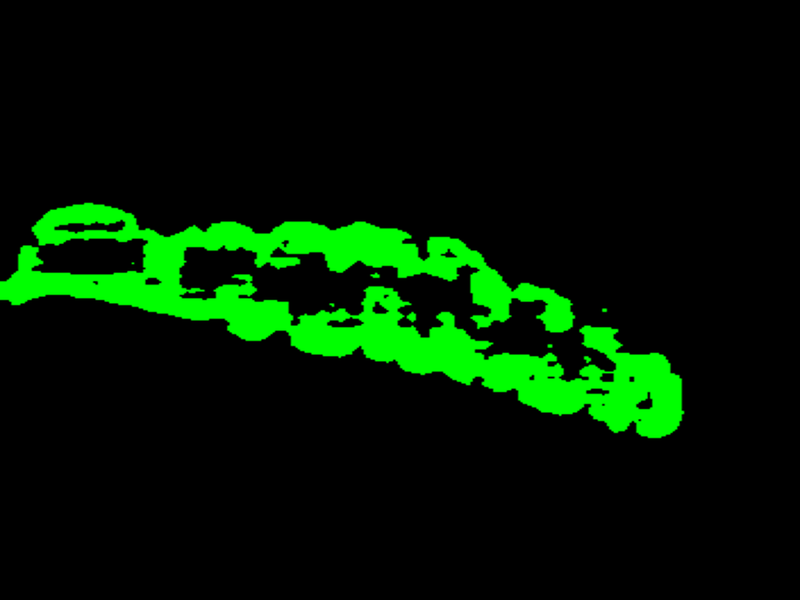} &
         \includegraphics[width=1.8cm, height=1.2cm]{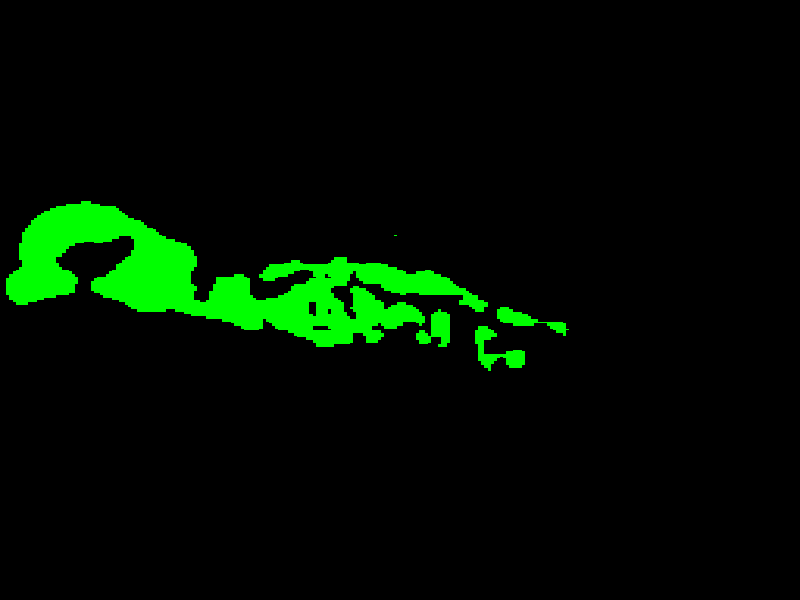} &
         \includegraphics[width=1.8cm, height=1.2cm]{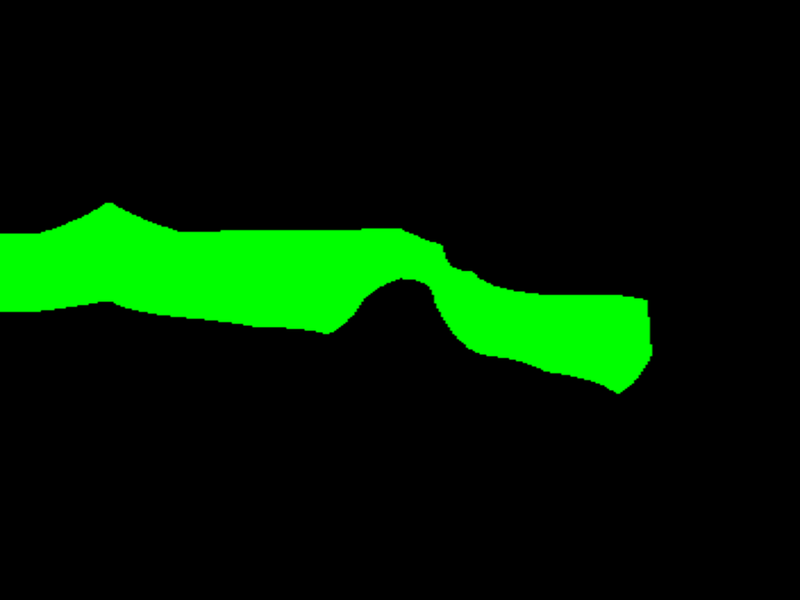} &
         \includegraphics[width=1.8cm, height=1.2cm]{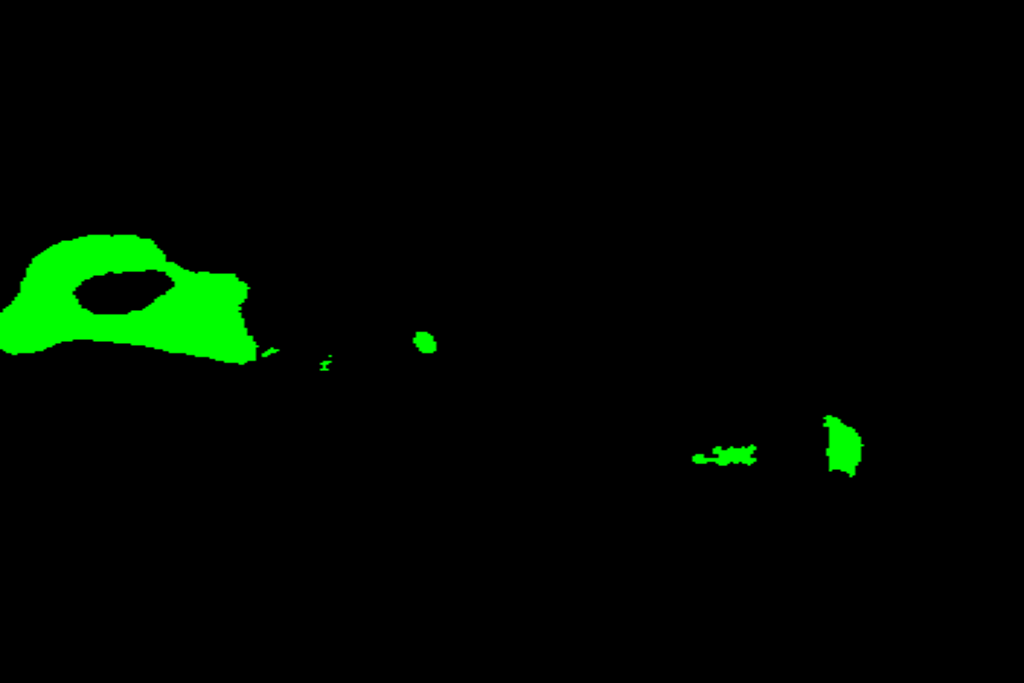}&
         \includegraphics[width=1.8cm, height=1.2cm]{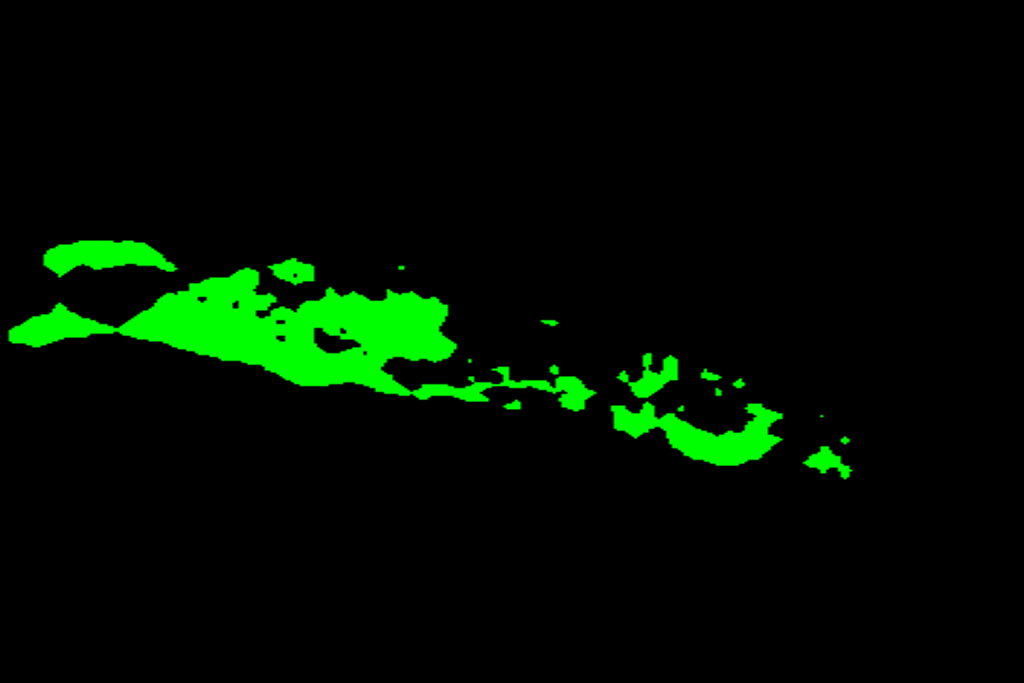} &
         \includegraphics[width=1.8cm, height=1.2cm]{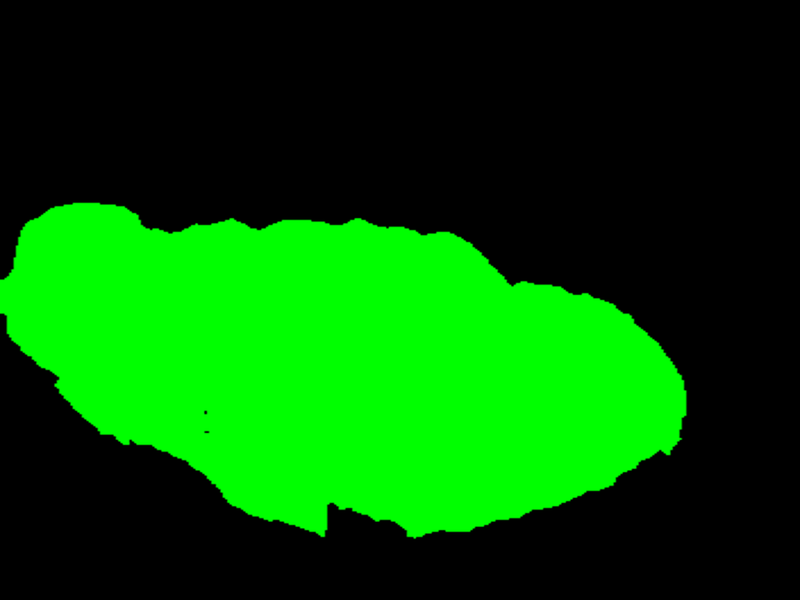} 
        \\

        \raisebox{0.10cm}{\makebox[0pt][c]{\rotatebox{90}{\small \textit{Scene C}}}} &
         \includegraphics[width=1.8cm, height=1.2cm]{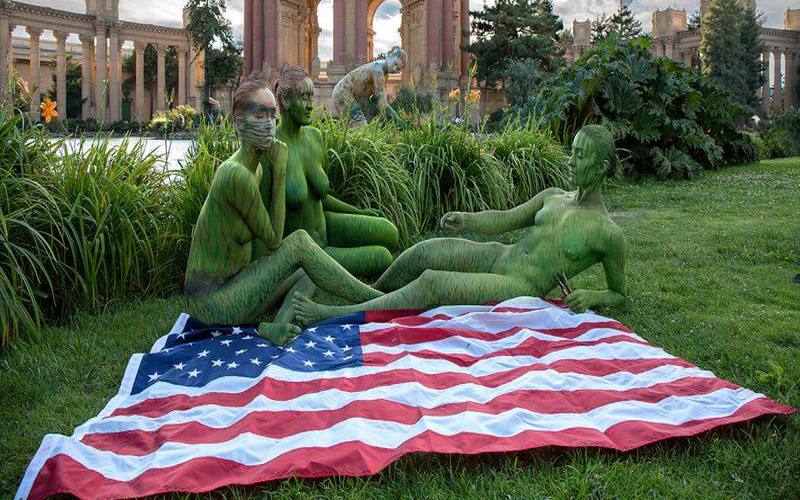} &
         \includegraphics[width=1.8cm, height=1.2cm]{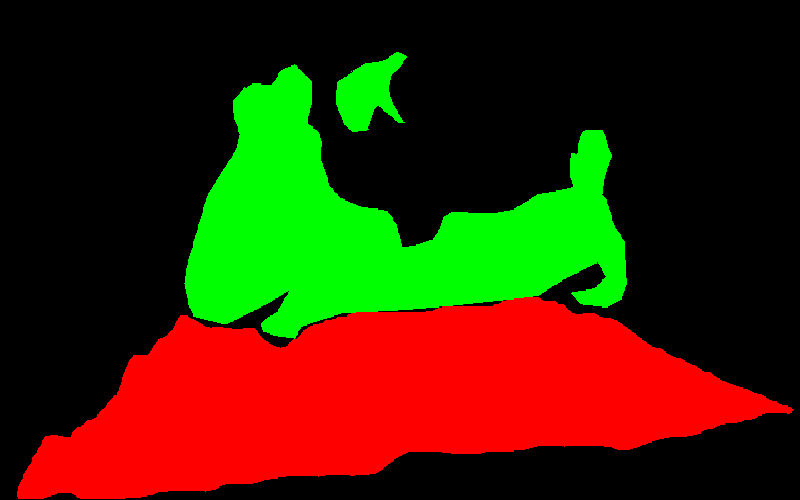} &
         \includegraphics[width=1.8cm, height=1.2cm]{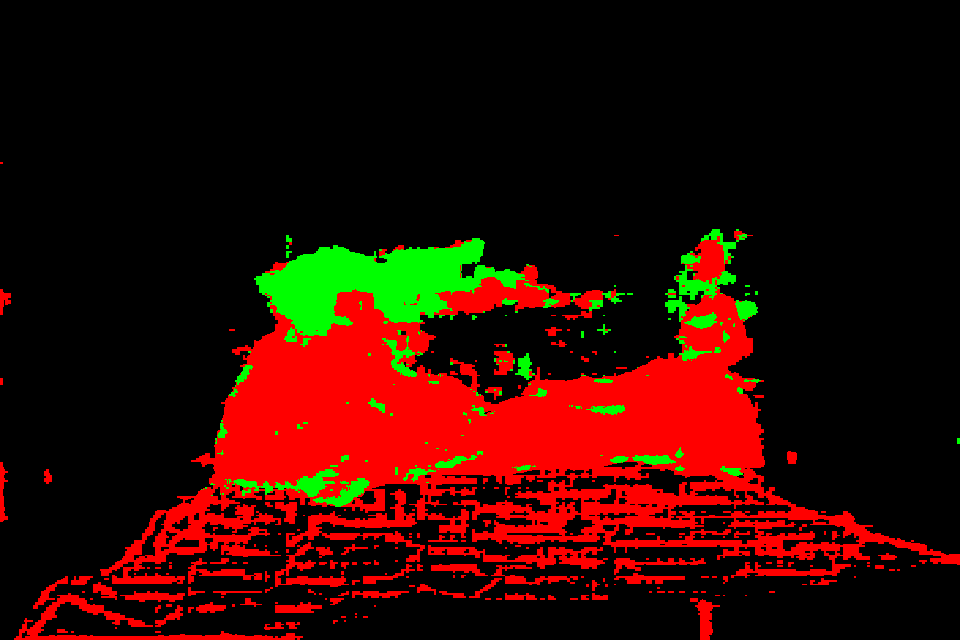} &
         \includegraphics[width=1.8cm, height=1.2cm]{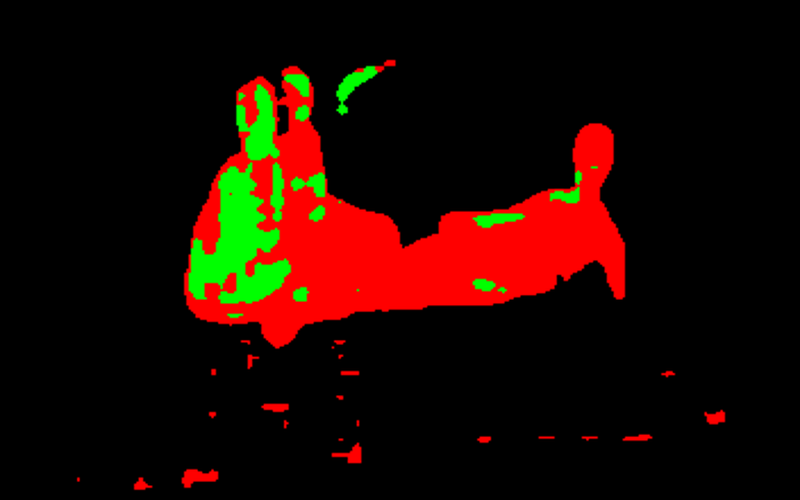} &
         \includegraphics[width=1.8cm, height=1.2cm]{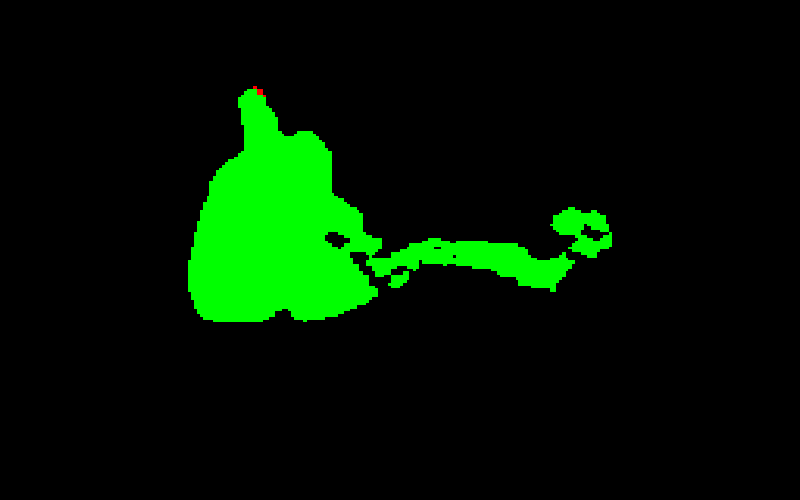} &
         \includegraphics[width=1.8cm, height=1.2cm]{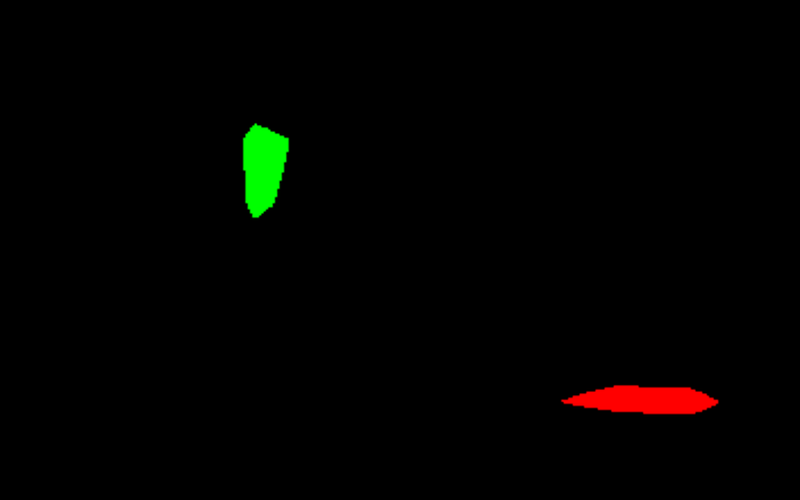} &
         \includegraphics[width=1.8cm, height=1.2cm]{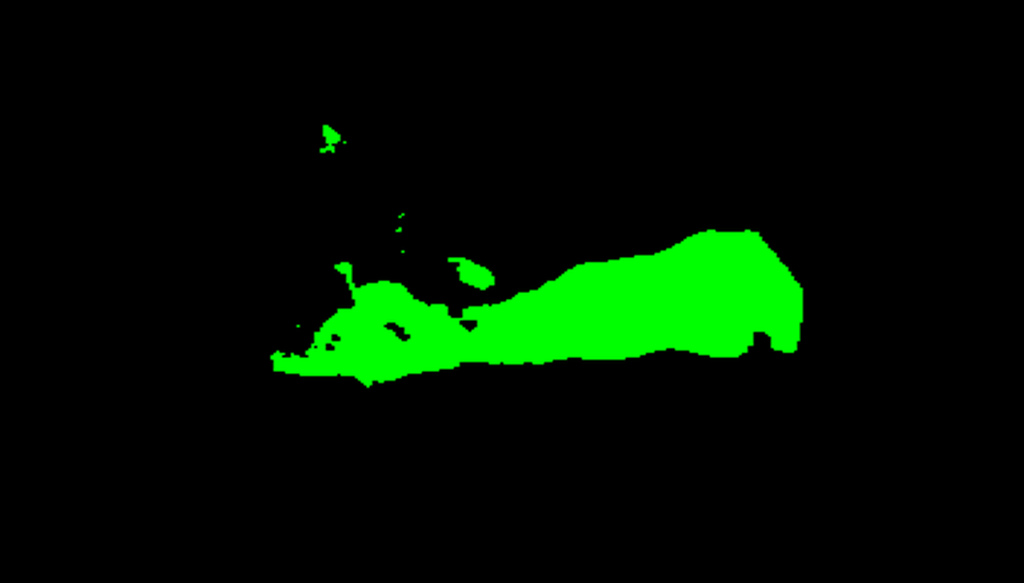}&
         \includegraphics[width=1.8cm, height=1.2cm]{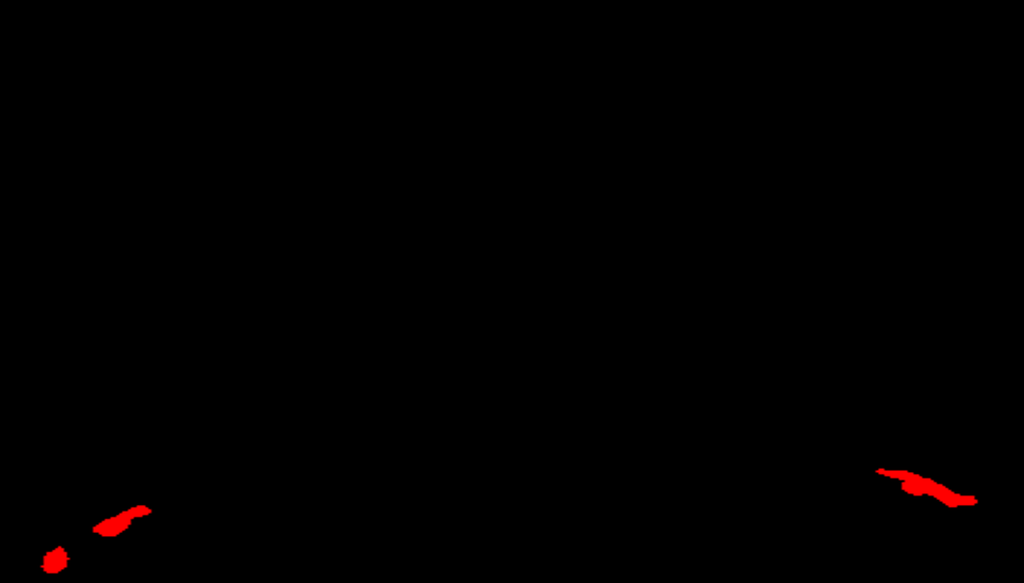}&
         \includegraphics[width=1.8cm, height=1.2cm]{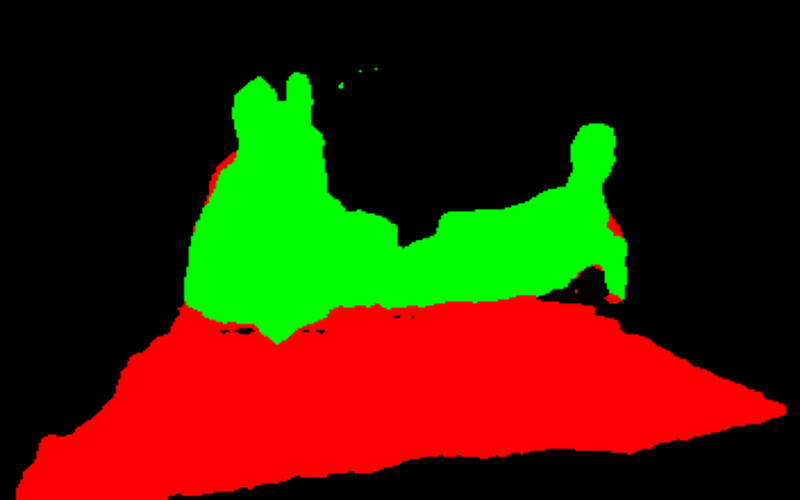} 
        \\

        \raisebox{0.10cm}{\makebox[0pt][c]{\rotatebox{90}{\small \textit{Scene D}}}} &
         \includegraphics[width=1.8cm, height=1.2cm]{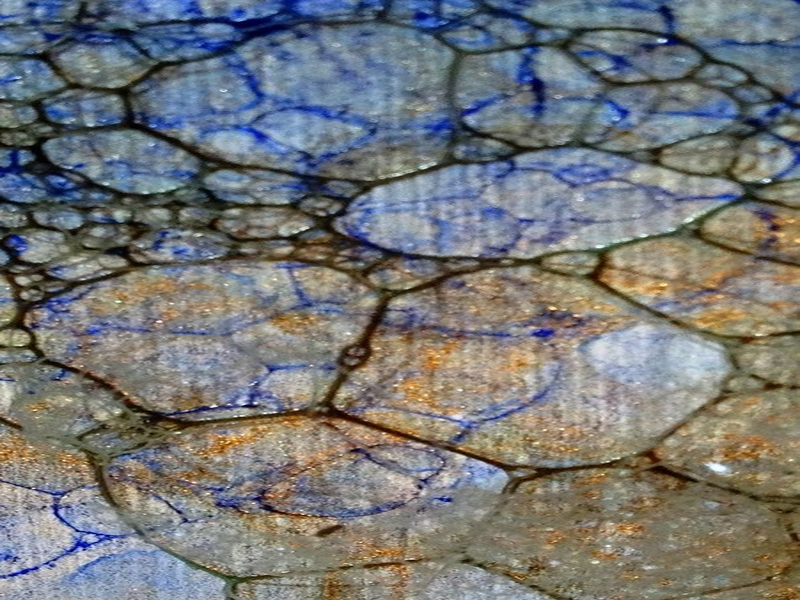} &
         \includegraphics[width=1.8cm, height=1.2cm]{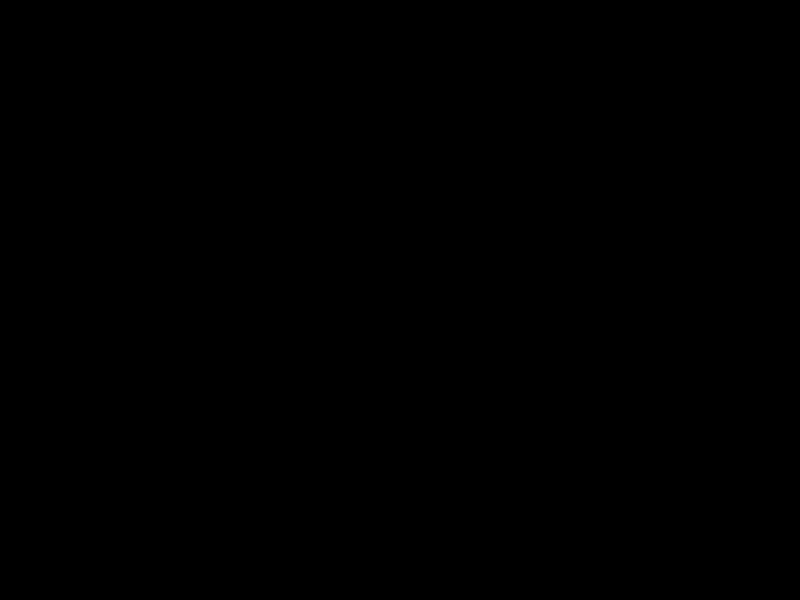} &
         \includegraphics[width=1.8cm, height=1.2cm]{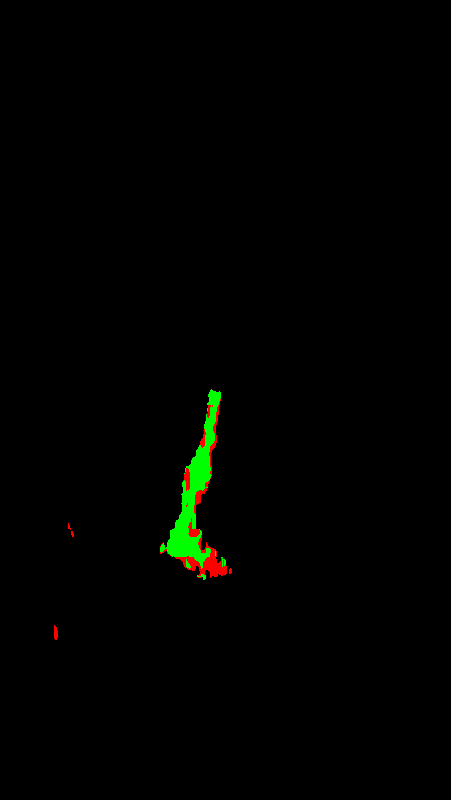} &
         \includegraphics[width=1.8cm, height=1.2cm]{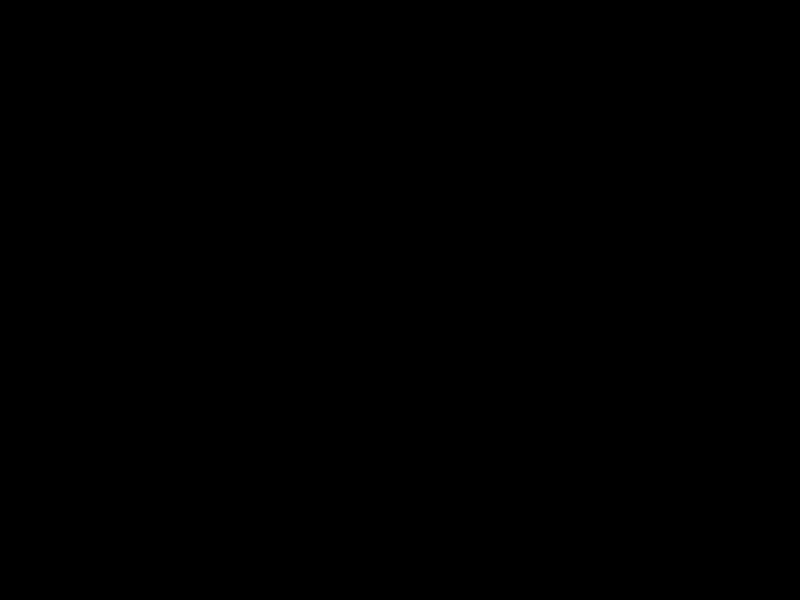} &
         \includegraphics[width=1.8cm, height=1.2cm]{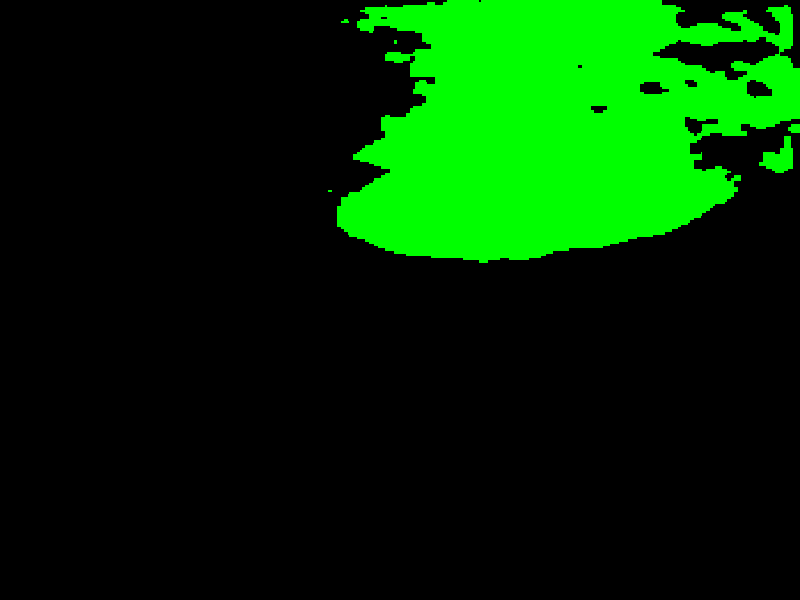} &
         \includegraphics[width=1.8cm, height=1.2cm]{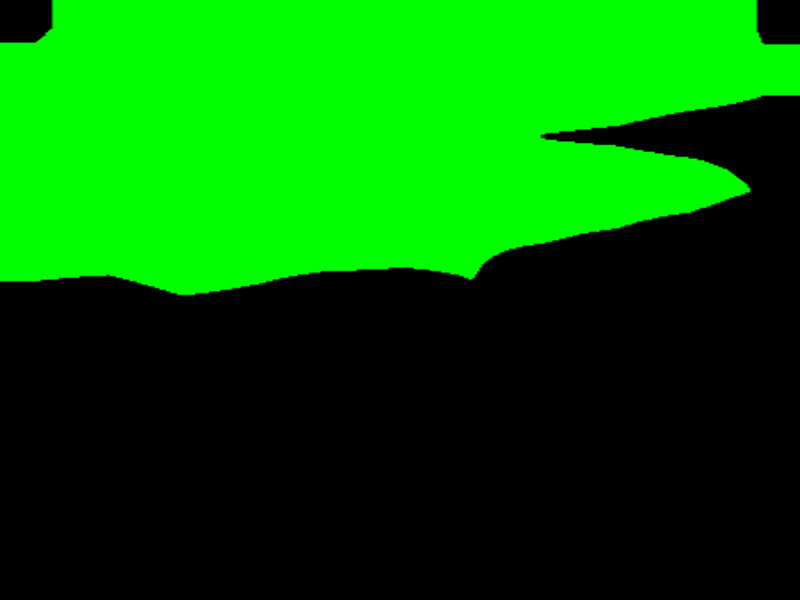} &
         \includegraphics[width=1.8cm, height=1.2cm]{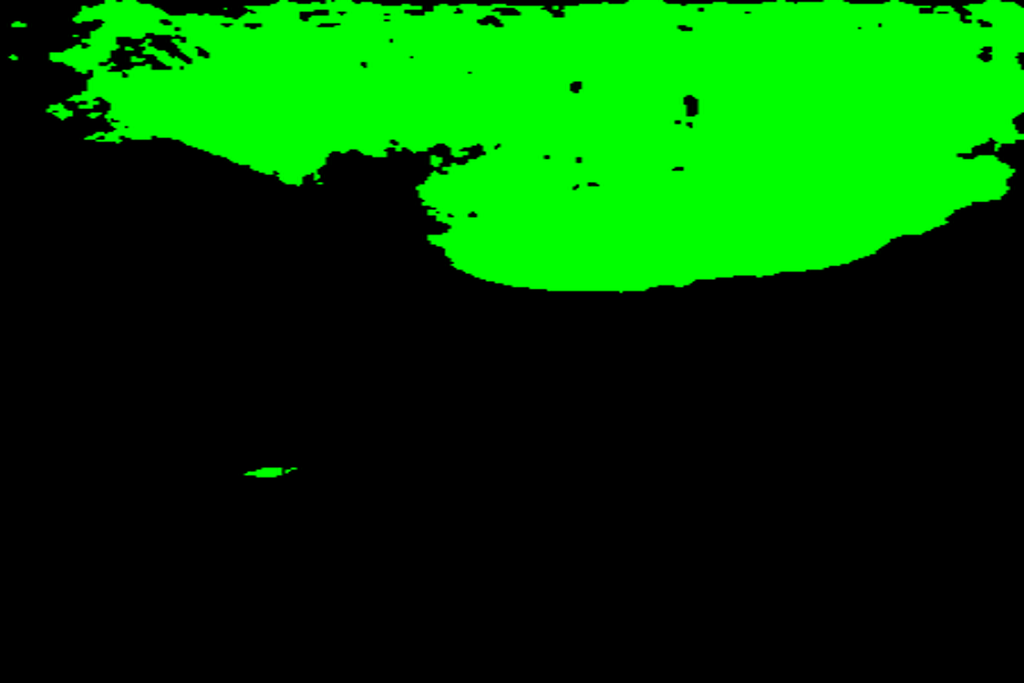} &
         \includegraphics[width=1.8cm, height=1.2cm]{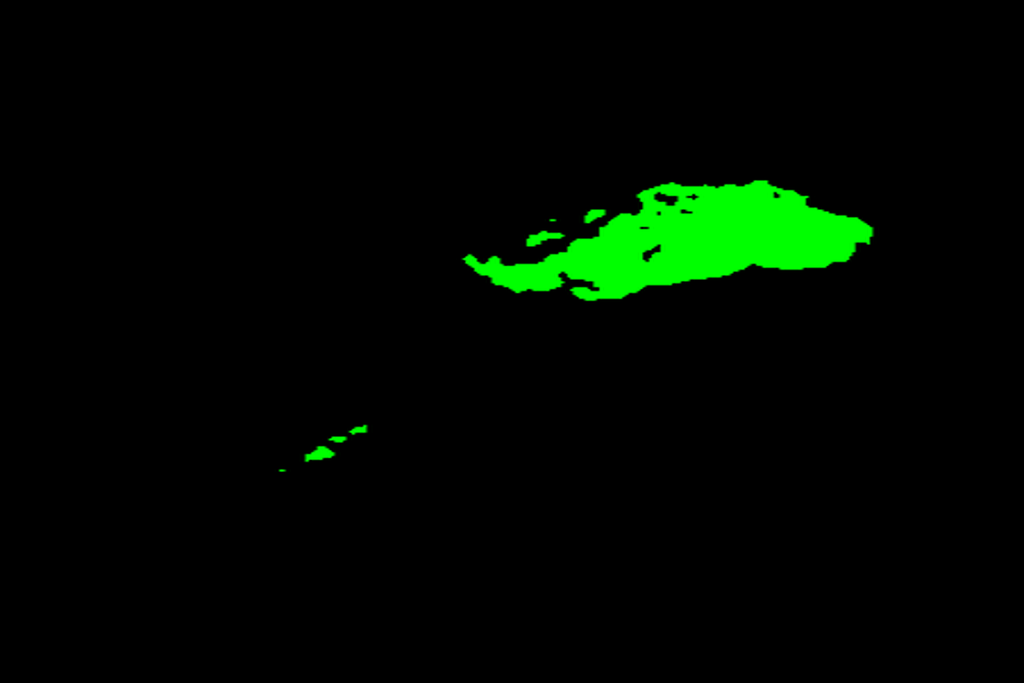} &
         \includegraphics[width=1.8cm, height=1.2cm]{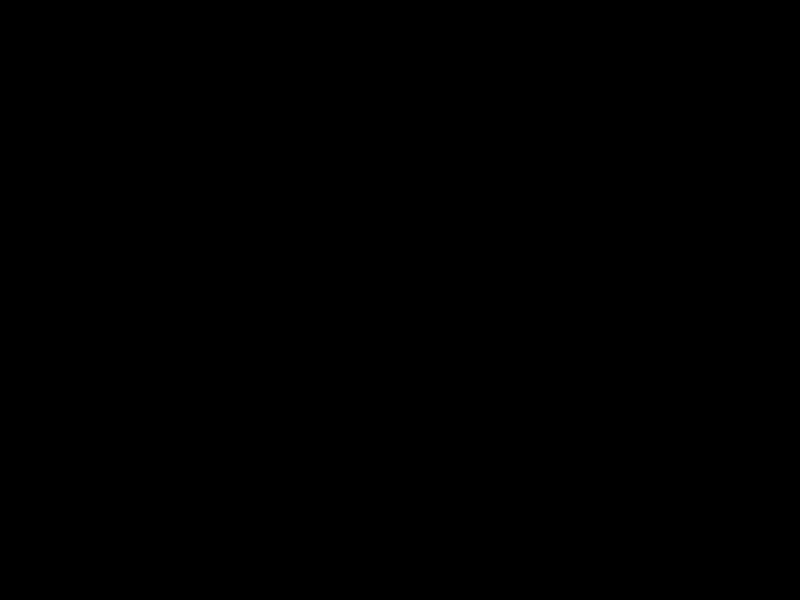} 
        \\

          \put(18,-7){{Image}}
       \put(75,-7){{GT}}
       \put(120,-7){VSCode}
        \put(165,-7){{SAM2-Ada.}}
        \put(230,-7){{ICON}}
        \put(280,-7){{ICEG}}
        \put(335,-7){{EDN}}
        \put(385,-7){{PFNet}}
         \put(435,-7){\textbf{Ours}}\\
    \end{tabular}
    \caption{Qualitative comparisons of~\ourmodel~with five models across overall scenes. More visualization can be seen in {\textcolor[rgb]{0.8588, 0.2666, 0.2156}{Appendix. \S 6}.}}
    \label{fig:visual_comp}
\end{figure*}

\subsection{Quantitative Evaluation}
\noindent\textbf{Performance on USC12K.}
We present in Table \ref{tab:my_label_sota} the performance of compared models on the USC12K benchmark. Comparing the results of the models in single-attribute scenes (refer to Scene A and Scene B) with those in multi-attribute scenes (refer to Scene C) reveals that all models achieve lower scores in Scene C than in Scene A and Scene B. This indicates that the simultaneous presence of both salient objects and camouflaged objects increases the difficulty for the models to recognize both. Our method achieves a greater lead in overall scenes, demonstrating that our model is more adaptable when faced with more challenging scenarios.
Furthermore,~\ourmodel~achieves the best performance in all scenarios compared to all other compared methods. Additionally, the evaluation results for other metrics, including AUC $\uparrow$, SI-AUC $\uparrow$, $F_m^{\beta}$ $\uparrow$, SI-$F_m^{\beta}$ $\uparrow$, $F_{\max}^{\beta}$ $\uparrow$, SI-$F_{\max}^{\beta}$ $\uparrow$, $E_m$ $\uparrow$, can be found in Table \textcolor{iccvblue}{1} and Table \textcolor{iccvblue}{2} in {\textcolor[rgb]{0.8588, 0.2666, 0.2156}{Appendix. \S 2}}.

\noindent\textbf{Misdetection performance.} To explore the impact of training with the USC12K dataset on the model's false detection rate, we test the SOD and COD models on the COD and SOD datasets. The results in Table \ref{tab:after_train} show that training with USC12K significantly reduced the model's false detection score compared to Table \ref{tab:experimental_phenomenon}.

\noindent\textbf{Generalization performance.} We evaluate the model’s performance across six widely used datasets. This includes COD datasets like COD10K, NC4K, and CAMO-TE, as well as SOD datasets like DUT-TE, HKU-IS, and DUT-OMRON. Our model demonstrates stronger generalization ability.
The results are shown in Table \ref{tab:supp_combined} and {\textcolor[rgb]{0.8588, 0.2666, 0.2156}{Appendix. \S 3}}.

\begin{table}[!t]
\centering
\caption{Effectiveness of different components in ARM. Intra-S.: intra-sample prompt query. Inter-S.: inter-sample prompt query.}
\label{tab:ablation}
\scriptsize
\setlength\tabcolsep{1pt}
\renewcommand{\arraystretch}{0.93}
\renewcommand{\tabcolsep}{0.4mm}

\begin{tabular}{c|c|cccc|c|cc|ccc}
\toprule
\multirow{1}{*}{Encoder} & 
\multirow{1}{*}{Decoder} & 
\multirow{1}{*}{Intra-S.} & 
\multirow{1}{*}{Inter-S.} & 
\multirow{1}{*}{Q2I} & 
\multirow{1}{*}{I2Q} & 
\multirow{1}{*}{Para.} & 
\multicolumn{1}{c|}{$\text{IoU}_S$} & 
\multicolumn{1}{c|}{$\text{IoU}_C$} & 
\multirow{1}{*}{mIoU} & 
\multirow{1}{*}{mAcc} & 
\multirow{1}{*}{CSCS} \\
\hline\hline
Frozen & Tuning & \textcolor[RGB]{219, 68, 55}{\XSolidBrush} & \textcolor[RGB]{219, 68, 55}{\XSolidBrush} & \textcolor[RGB]{219, 68, 55}{\XSolidBrush} & \textcolor[RGB]{219, 68, 55}{\XSolidBrush} & 4.22 & 66.42 & 44.02 & 68.78 & 77.65 & 11.58 \\
Tuning & Tuning & \textcolor[RGB]{219, 68, 55}{\XSolidBrush} & \textcolor[RGB]{219, 68, 55}{\XSolidBrush} & \textcolor[RGB]{219, 68, 55}{\XSolidBrush} & \textcolor[RGB]{219, 68, 55}{\XSolidBrush} & 4.36 & 71.42 & 56.71 & 74.98 & 84.74 & 9.12 \\
Tuning & Frozen & \textcolor[RGB]{219, 68, 55}{\XSolidBrush} & \textcolor[RGB]{15, 157, 88}{\CheckmarkBold} & \textcolor[RGB]{15, 157, 88}{\CheckmarkBold} & \textcolor[RGB]{15, 157, 88}{\CheckmarkBold} & 3.44 & 71.68 & 57.53 & 75.31 & 85.15 & 9.07 \\
Tuning & Frozen & \textcolor[RGB]{15, 157, 88}{\CheckmarkBold} & \textcolor[RGB]{219, 68, 55}{\XSolidBrush} & \textcolor[RGB]{15, 157, 88}{\CheckmarkBold} & \textcolor[RGB]{15, 157, 88}{\CheckmarkBold} & 4.03 & 74.32 & 58.91 & 76.96 & 85.80 & 7.98 \\
Tuning & Frozen & \textcolor[RGB]{15, 157, 88}{\CheckmarkBold} & \textcolor[RGB]{15, 157, 88}{\CheckmarkBold} & \textcolor[RGB]{219, 68, 55}{\XSolidBrush} & \textcolor[RGB]{219, 68, 55}{\XSolidBrush} & 0.75 & 70.97 & 56.56 & 74.77 & 84.43 & 9.85 \\
Tuning & Frozen & \textcolor[RGB]{15, 157, 88}{\CheckmarkBold} & \textcolor[RGB]{15, 157, 88}{\CheckmarkBold} & \textcolor[RGB]{15, 157, 88}{\CheckmarkBold} & \textcolor[RGB]{219, 68, 55}{\XSolidBrush} & 2.40 & 73.08 & 58.45 & 76.73 & 85.63 & 8.52 \\
\cellcolor{iccvblue!20}Tuning & \cellcolor{iccvblue!20}Frozen & \cellcolor{iccvblue!20}\textcolor[RGB]{15, 157, 88}{\CheckmarkBold} & \cellcolor{iccvblue!20}\textcolor[RGB]{15, 157, 88}{\CheckmarkBold} & \cellcolor{iccvblue!20}\textcolor[RGB]{15, 157, 88}{\CheckmarkBold} & \cellcolor{iccvblue!20}\textcolor[RGB]{15, 157, 88}{\CheckmarkBold} & \cellcolor{iccvblue!20}4.04 & \cellcolor{iccvblue!20}\textbf{75.57} & \cellcolor{iccvblue!20}\textbf{61.34} & \cellcolor{iccvblue!20}\textbf{78.03} & \cellcolor{iccvblue!20}\textbf{87.92} & \cellcolor{iccvblue!20}\textbf{7.49} \\
\bottomrule
\end{tabular}
\end{table}

\begin{figure}[ht]
  \centering
  \includegraphics[width=\columnwidth]{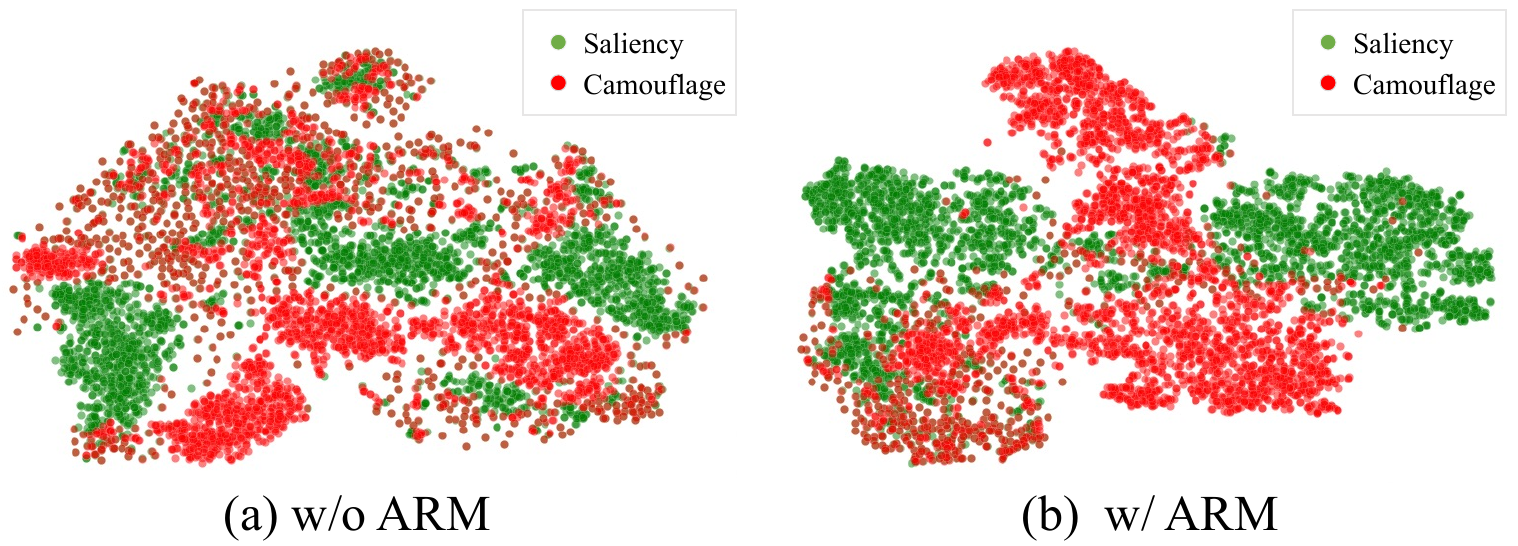}
   \caption{Embedding visualization of salient and camouflaged objects in USC12K test set images by our model w/o and w/ the ARM module. Dimensions reduced using t-SNE~\cite{van2008visualizing}.
   }
  \label{fig:t-SNE Distribution}
\end{figure}

\subsection{Qualitative Evaluation}
In Figure~\ref{fig:visual_comp}, we compare our model's qualitative results with six models. Our method exhibited a better detection capability for salient and camouflaged objects in all scenes.
Benefiting from the ARM module, our method better distinguished salient objects and camouflaged objects in the same image within Scene C of Figure~\ref{fig:visual_comp}.
For Scene D, SOD and COD methods become confused when encountering backgrounds, resulting in poor performance and unstable robustness, whereas our model demonstrates better performance.

\subsection{Ablation Study} 
\noindent\textbf{Effectiveness of Different Components in ARM.} As shown in Table \ref{tab:ablation}, ablation studies on overall scene validate the effectiveness of the ARM module's components.
From Line 3, 4, and 7, it is evident that both Intra-SPQ and Inter-SPQ improve the performance of the model, with Intra-SPQ providing a greater performance enhancement than Inter-SPQ when used together, achieving optimal results. Line 5, 6, and 7 demonstrate that Q2I and I2Q also facilitate the distinction of salient camouflaged objects, leading to reductions in CSCS of 2.36\% and 1.08\%, respectively. Additionally, compared to the original SAM2 (refer to Line 1) and SAM2-Adapter (refer to Line 2), the proposed~\ourmodel~enhances performance on the USC12K task across all metrics through a more efficient fine-tuning approach by incorporating the ARM module and frozen mask decoder.

\noindent\textbf{Impact of ARM on Embedding Separation.} We further demonstrate the effectiveness of the ARM module by visualizing embeddings of the salient and camouflaged objects. As shown in Figure~\ref{fig:t-SNE Distribution}, the plot without ARM module is homogeneous, mixing salient and camouflaged object embeddings. In contrast, our model with the ARM module produces a more distinct and well-clustered representation, better separating the embeddings of salient and camouflaged objects. More ablation can be seen in {\textcolor[rgb]{0.8588, 0.2666, 0.2156}{Appendix. \S 7}.}

\section{Conclusion}
We analyze the misdetection of salient and camouflaged objects in current SOD and COD methods, and pinpoint the constraints of existing datasets and the models. A large-scale dataset named USC12K to advance unconstrained salient and camouflaged object detection. Accordingly, we design a unified pipeline USCNet to explicitly models both inter-sample and intra-sample attribute relationships. In addition, we propose a new evaluation metric, CSCS, to assess the model's confusion between camouflage and saliency. Extensive experiments demonstrate that the proposed dataset alleviates the models' misdetection issue, and our method outperforms existing related models on the USC12K benchmark, showing better generalization across six SOD and COD datasets. We believe that the USC12K benchmark will promote further research in SOD and COD, helping models better capture saliency and camouflage patterns that align with the human visual system.

{
    \small
    \bibliographystyle{ieeenat_fullname}
    \bibliography{main}
}

\setcounter{section}{0} 
\section*{Appendix}
\label{sec:suppintro}
We summarize the supplementary material from the following aspects:
\paragraph{Table of contents:}
\begin{itemize}
\item \S\ref{CSCS_Metric}: CSCS Metric
\item \S\ref{Performance_of_models_in_detecting_objects_of_varying_sizes}:Performance in Detecting Objects of Different Sizes
\item \S\ref{Results_on_COD_and_SOD_Datasets}: Results on Popular COD and SOD Datasets
\item \S\ref{More_Technical_Details}: More Technical Details
\item \S\ref{More_USC12K_Dataset_Examples}: More USC12K Dataset Detail and Examples
\item \S\ref{Additional_Qualitative_Results}: Additional Qualitative Results
\item \S\ref{Additional_Ablation_Study}: Additional Ablation Study


\end{itemize}

\section{CSCS Metric}
\label{CSCS_Metric}

\begin{figure}[!t]
  \centering
  \includegraphics[width=0.5\textwidth]{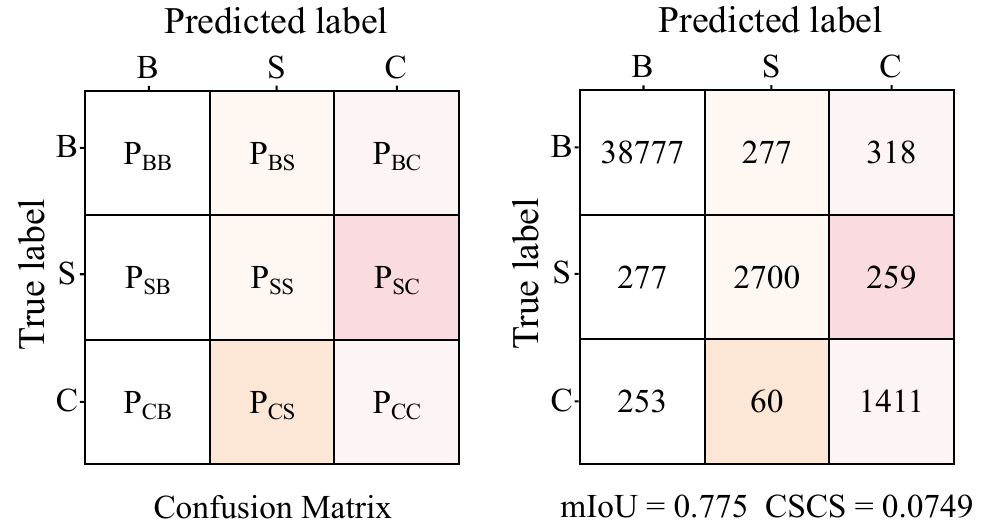}
  \put(-215,-10){{(a) Illustration}}
      \put(-90,-10){{(b)~\ourmodel}}
  \caption{Confusion matrix of our~\ourmodel~on the USC12K test set. The units of the values in the confusion matrix are in tens of thousands (\textbf{1E+04}).}
  \label{fig:supp_CSCS_confusion}
\end{figure}

Contrary to the Intersection over Union (IoU) that measures accuracy for a single class, the Camouflage-Saliency Confusion Score (CSCS) assesses the misclassification between two distinct classes. The CSCS, designed to evaluate the confusion between camouflaged and salient objects, is calculated as follows:

\begin{equation}\label{supp_CSCS_equa}\small
\text{CSCS} = \frac{1}{2} (\frac{\mathcal{P}_{CS}}{\mathcal{P}_{BS} + \mathcal{P}_{SS} + \mathcal{P}_{CS}} + \frac{\mathcal{P}_{SC}}{\mathcal{P}_{BC} + \mathcal{P}_{SC} + \mathcal{P}_{CC}}), 
\end{equation}
where $ \mathbb{P} = \left\{ \mathcal{P}_{\lambda \theta} \,|\, \lambda \in \Theta, \theta \in \Theta \right\},\ \Theta = \left\{B, C, S\right\} $, the B, C and S denote background, camouflage and saliency. A lower CSCS value indicates a stronger ability of the network to discriminate between salient and camouflaged objects. \textit{$ \mathcal{P}_{CS} $ } represents the label as camouflage but is predicted as saliency. We aim to minimize the misclassification of camouflaged pixels as salient, ensuring the network correctly distinguishes between camouflaged and salient objects. The same applies to $\mathcal{P}_{SC}$. As shown in Figure~\ref{fig:supp_CSCS_confusion}, we present the confusion matrix of the proposed~\ourmodel~on the USC12K test set. Our model balances improvements across all metrics, achieving a mIoU of 0.775 and a CSCS of 0.0749 (\textcolor{magenta}{see Table 3 in the manuscript}).
\begin{figure}[ht]
  \centering
  \includegraphics[width=\columnwidth]{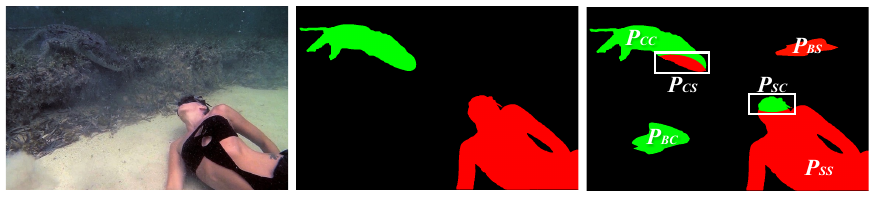}
  \put(-210,-8){{Image}}
    \put(-125,-8){{GT}}
      \put(-60,-8){{Prediction}}
  \caption{The illustration of $\mathcal{P}_{BS}$, $\mathcal{P}_{SS}$, $\mathcal{P}_{CS}$, $\mathcal{P}_{BC}$, $\mathcal{P}_{SC}$, and $\mathcal{P}_{CC}$ in the CSCS metric. The \textcolor[RGB]{219, 68, 55}{\textbf{red}} mask represents the salient regions, and the \textcolor[RGB]{15, 157, 88}{\textbf{green}} mask denotes the camouflaged regions.}
  \label{fig:CSCS}
\end{figure}

\begin{table*}[!t]
\begin{center}

\caption{Performance of different models detecting salient objects on USC12K testing set.}\label{tab:supp_sisod}
    \scriptsize
    \setlength\tabcolsep{100pt}
    \renewcommand{\arraystretch}{1.2}
    \renewcommand{\tabcolsep}{3.0mm}

\begin{tabular}{c|l|c|*{7}{p{1.12cm}<{\centering}}}
\toprule
\multirow{2}{*}{\textbf{Task}} &
\multirow{2}{*}{\textbf{Model}} & 
\ {\textbf{Update}} &
\multicolumn{7}{c}{\textbf{USC12K-SOD}}
\\
\cline{4-10}
&      &  {  } {\textbf{Params(M)}}
 & AUC $\uparrow$ & SI-AUC$\uparrow$ &$F_m^{\beta}$$\uparrow$&SI-$F_m^{\beta}$$\uparrow$
 & $F_{\max}^{\beta}$$\uparrow$ & SI-$F_{\max}^{\beta}$$\uparrow$ & $E_m$$\uparrow$  
\\ 
\hline\hline
\textbf{\multirow{6}{*}{SOD}}

& GateNet~\cite{zhao2020suppress} 
& 128                             
&.810
&.812
&.696
&.754
&.706
&.764
&.775
\\

&F3Net~\cite{wei2020f3net}
& 26
& .828 & .826 & .722 & .765 & .734 & .777 & .803
\\

& MSFNet~\cite{zhang2021auto}     
& 28                              
& .832 & .831 & .726 & .772 & .735 & .782 & .805
\\

& VST~\cite{liu2021visual}        
& 43          
& .777 & .777 & .642 & .732 & .650 & .741 & .742
\\

&EDN~\cite{wu2022edn}
&43
& .831 & .830 & .726 & .769 & .736 & .780 & .804
\\

& ICON~\cite{zhuge2022salient}    & 32    
& .821 & .832 & .702 & .764 & .711 & .774 & .795
\\
\hline
\textbf{\multirow{9}{*}{COD}}
& SINetV2~\cite{fan2021concealed} 
& 27                              
& .843 & .842 & .755 & .783 & .765 & .793 & .827
\\
& PFNet~\cite{mei2021camouflaged} 
& 47  
& .820 & .822 & .712 & .756 & .724 & .767 & .799
\\

& ZoomNet~\cite{pang2022zoom}     
& 33  
& .821 & .823 & .710 & .765 & .720 & .774 & .791
\\
                             

& FEDER~\cite{he2023camouflaged}  
& 44                    
& .841 & .842 & .742 & .784 & .750 & .796 & .820
\\   
& ICEG~\cite{he2024strategic}  
&    100
&.830 & .825& 734 & .762 & .743
&.770 & .831 
\\   

& PRNet~\cite{hu2024efficient}  
&   13& .851 & .845 & .742 & .779 & .750 & .792 & .832
\\   
& CamoFormer~\cite{yin2024camoformer}  &71& .844 & .843 & .750 & .782 & .758 & .790 & .821
\\   
& PGT~\cite{wang2024camouflaged}  &68
& .831 & .828 & .717 & .773 & .727 & .784 & .791
\\   
& SAM2-Adapter~\cite{chen2024sam2}  
& 4.36
& .847 & .847 & .741 & .783 & .751 & .794 & .816
\\
\hline
\textbf{\multirow{3}{*}{Unified}}  
& VSCode~\cite{luo2024vscode}  & 60  & .843 & .842 & .749 & .776 & .768 & .789 & .822
\\
& EVP~\cite{liu2023explicit}  &  4.95& .850 & .847 & .751 & .782 & .771 & .792 & .830
\\


&\cellcolor{iccvblue!20}\textbf{\ourmodel(Ours)} 
& \cellcolor{iccvblue!20}\textbf{4.04} 
& \cellcolor{iccvblue!20}\textbf{.853} 
& \cellcolor{iccvblue!20}\textbf{.850} 
& \cellcolor{iccvblue!20}\textbf{.761} 
& \cellcolor{iccvblue!20}\textbf{.787} 
& \cellcolor{iccvblue!20}\textbf{.772} 
& \cellcolor{iccvblue!20}\textbf{.798} 
& \cellcolor{iccvblue!20}\textbf{.833}  
\\

\bottomrule
\end{tabular}
\end{center}
\end{table*}

\begin{table*}[!t]
\begin{center}

\caption{Performance of different models detecting camouflaged objects on USC12K testing set.}\label{tab:supp_sicod}
    \scriptsize
    \setlength\tabcolsep{100pt}
    \renewcommand{\arraystretch}{1.2}
    \renewcommand{\tabcolsep}{3.0mm}

\begin{tabular}{c|l|c|*{7}{p{1.12cm}<{\centering}}}
\toprule
\multirow{2}{*}{\textbf{Task}} &
\multirow{2}{*}{\textbf{Model}} & 
\ {\textbf{Update}} &
\multicolumn{7}{c}{\textbf{USC12K-COD}}
\\
\cline{4-10}
&      &  {  } {\textbf{Params(M)}}
 & AUC$\uparrow$ & SI-AUC$\uparrow$ &$F_m^{\beta}$$\uparrow$&SI-$F_m^{\beta}$$\uparrow$
 & $F_{\max}^{\beta}$$\uparrow$ & SI-$F_{\max}^{\beta}$$\uparrow$ & $E_m$$\uparrow$    
\\ 
\hline\hline
\textbf{\multirow{6}{*}{SOD}}

& GateNet~\cite{zhao2020suppress} 
& 128                             
& .692 & .687 & .443 & .558 & .453 & .569 & .651
\\

&F3Net~\cite{wei2020f3net}
& 26
& .695 & .687 & .449 & .564 & .458 & .574 & .649
\\

& MSFNet~\cite{zhang2021auto}     
& 28                              
& .698 & .691 & .455 & .565 & .465 & .576 & .659
\\

& VST~\cite{liu2021visual}        
& 43          
& .626 & .625 & .303 & .536 & .312 & .546 & .524
\\

&EDN~\cite{wu2022edn}
&43
& .709 & .703 & .476 & .575 & .485 & .585 & .670
\\

& ICON~\cite{zhuge2022salient}    & 32    
& .663 & .663 & .384 & .549 & .394 & .560 & .587
\\
\hline
\textbf{\multirow{9}{*}{COD}}
& SINetV2~\cite{fan2021concealed} 
& 27                              
& .715 & .705 & .505 & .588 & .514 & .598 & .690
\\
& PFNet~\cite{mei2021camouflaged} 
& 47  
& .678 & .672 & .429 & .544 & .440 & .555 & .630
\\

& ZoomNet~\cite{pang2022zoom}     
& 33  
& .657 & .653 & .394 & .545 & .405 & .556 & .588
\\
                             

& FEDER~\cite{he2023camouflaged}  
& 44 & .710 & .703 & .486 & .567 & .497 & .578 & .689
\\   
& ICEG~\cite{he2024strategic}  
&    100
&.730 & .717 & .525 & .601 & .532
&.609 & .719 
\\   

& PRNet~\cite{hu2024efficient}  
&   13
& .705 & .695 & .454 & .569 & .464 & .579 & .652
\\   
& CamoFormer~\cite{yin2024camoformer}  &71& .756 & .745 & .565 & .626 & .575 & .636 & .743
\\   
& PGT~\cite{wang2024camouflaged}  &68
& .746 & .734 & .527 & .596 & .539 & .607 & .715
\\   
& SAM2-Adapter~\cite{chen2024sam2}  
& 4.36& .770 & .761 & .575 & .637 & .585 & .647 & .746
\\

\hline
\textbf{\multirow{3}{*}{Unified}}  
& VSCode~\cite{luo2024vscode}  & 60  & .735 & .727 & .519 & .601 & .525 & .597 & .722
\\
& EVP~\cite{liu2023explicit}  &  4.95& .695 & .684 & .485 & .577 & .494 & .587 & .650
\\

&\cellcolor{iccvblue!20}\textbf{\ourmodel(Ours)} 
& \cellcolor{iccvblue!20}\textbf{4.04} 
& \cellcolor{iccvblue!20}\textbf{.801} 
& \cellcolor{iccvblue!20}\textbf{.794} 
& \cellcolor{iccvblue!20}\textbf{.610} 
& \cellcolor{iccvblue!20}\textbf{.658} 
& \cellcolor{iccvblue!20}\textbf{.619} 
& \cellcolor{iccvblue!20}\textbf{.667} 
& \cellcolor{iccvblue!20}\textbf{.795}  \\

\bottomrule
\end{tabular}
\end{center}
\end{table*}

\section{Performance in Detecting Objects of Different Sizes}
\label{Performance_of_models_in_detecting_objects_of_varying_sizes}
To evaluate the model's ability to detect objects of varying sizes, we employ several metrics: AUC $\uparrow$, SI-AUC $\uparrow$, $F_m^{\beta}$ $\uparrow$, SI-$F_m^{\beta}$ $\uparrow$, $F_{\max}^{\beta}$ $\uparrow$, SI-$F_{\max}^{\beta}$ $\uparrow$, $E_m$ $\uparrow$. From \tabref{tab:supp_sisod} and \tabref{tab:supp_sicod}, it can be observed that, compared to the size-sensitive(e.g., AUC $\uparrow$ and $F_m^{\beta}$ $\uparrow$) and size-invariance metrics(e.g., SI-AUC $\uparrow$ and SI-$F_m^{\beta}$ $\uparrow$), our method exhibits smaller performance fluctuations, demonstrating its robustness to variations in object size and number in the scene.


\begin{table*}[!t]
\begin{center}
\caption{Generalization performance of related methods on the DUTS, HKU-IS, and DUT-OMRON test sets. $\uparrow$ / $\downarrow$ represents the higher/lower the score, the better.}\label{tab:supp_sod}
    \scriptsize

    \setlength\tabcolsep{100pt}
    \renewcommand{\arraystretch}{1.2}
    \renewcommand{\tabcolsep}{1.2mm}

\begin{tabular}{c|l|c|ccccc|ccccc|ccccc}
\toprule
\multirow{2}{*}{Task} &
\multirow{2}{*}{Model} & 
\ Update &
\multicolumn{5}{c|}{DUTS} & 
\multicolumn{5}{c|}{HKU-IS} &
\multicolumn{5}{c}{DUT-OMRON}
\\
\cline{4-8}
\cline{9-13}
\cline{14-18}
&      &  {  } Params(M) 

 & $F_\beta^\text{max} \uparrow$ &  $F_\beta^\omega \uparrow$ & $M \downarrow$ &
    $S_\alpha \uparrow$ & $E_{\phi}^\text{m} \uparrow$  &
    
    $F_\beta^\text{max} \uparrow$ &  $F_\beta^\omega \uparrow$ & $M \downarrow$ &
    $S_\alpha \uparrow$ & $E_{\phi}^\text{m} \uparrow$  & 
$F_\beta^\text{max} \uparrow$ &  $F_\beta^\omega \uparrow$ & $M \downarrow$ &
    $S_\alpha \uparrow$ & 
    
    $E_{\phi}^\text{m} \uparrow$  

\\ 
\hline\hline
\textbf{\multirow{6}{*}{SOD}}

& GateNet~\cite{zhao2020suppress} 
& 128                             

&.666
&.644
&.062
&.755
&.765

&.804
&.785
&.049
&.841
&.857

&.634
&.603
&.079
&.747
&.751

\\

&F3Net~\cite{wei2020f3net}
&26

&.703
&.683
&.055
&.783
&.794

&.832
&.816
&.044
&.853
&.881

&.638
&.615
&.073
&.747
&.758

\\

& MSFNet~\cite{zhang2021auto}     & 28          

&.651
&.638
&.063
&.749
&.758           

&.824
&.806
&.045
&.853
&.877

&.641
&.611
&.076
&.751
&.764
\\

& VST~\cite{liu2021visual}        
& 43                             
&.630
&.610
&.061
&.744
&.749

&.777
&.760
&.052
&.820
&.851

&.580
&.560
&.073
&.720
&.715
\\

&EDN~\cite{wu2022edn}
&43

&.692
&.676
&.053
&.784
&.785

&.820
&.806
&.043
&.852
&.873

&.616
&.597
&.071
&.742
&.735

\\

& ICON~\cite{zhuge2022salient}    & 32                              
&.679
&.647
&.069
&.769
&.785

&.814
&.787
&.051
&.843
&.874

&.615
&.576
&.099
&.728
&.738

\\
\hline
\textbf{\multirow{9}{*}{COD}}
& SINetV2~\cite{fan2021concealed} & 27                              
&.732
&.710
&.052
&.801
&.821

&.838
&.822
&.046
&.847
&.884

&.665
&.642
&.068
&.763
&.786

\\
& PFNet~\cite{mei2021camouflaged} & 47                              
&.691
&.668
&.060
&.775
&.790

&.818
&.801
&.048
&.843
&.876

&.643
&.614
&.075
&.747
&.764

\\

& ZoomNet~\cite{pang2022zoom}     & 33   

&.729
&.709
&.053
&.801
&.813

&.785
&.774
&.051
&.830
&.842

&.623
&.601
&.075
&.742
&.735

\\             

& FEDER~\cite{he2023camouflaged}  & 44                    
& .736 & .714 & .052 & .808 & .821

&.839
&.827
&.045
&.869
&.881

&.645
&.615
&.077
&.755
&.760

\\   
& PRNet~\cite{hu2024efficient}  &13
&.773
&.756
&.043
&.830
&.849

&.840
&.833
&.044
&.857
&.880

&.708
&.685
&.057
&\textbf{.796}
&.808
\\

& ICEG~\cite{he2024strategic}  &
 100         &
 .719 & 
 .700 & 
 .050 & 
 .789 & 
 .820 &
 .832 &
 .815 &
 .045 &
 .848 &
 \textbf{.896} &
 .664 & 
 .645 & 
 .061 & 
 .762 & 
 .785
\\

& CamoFormer~\cite{yin2024camoformer}   &
 71 &
 .733 & 
 .715 & 
 .049 & 
 .813 & 
 .819 &

 .838 &
 .817 & 
 .046 & 
 .857 & 
 .884 &

 .687 & 
 .661 & 
 .066 & 
 .783 & 
 .793
\\   

& PGT~\cite{wang2024camouflaged}  &
 68 &
 .686 & 
 .670 & 
 .053 & 
 .786 & 
 .779 &

 .819 & 
 .802 & 
 .044 & 
 .855 & 
 .871 &

 .642 & 
 .619 & 
 .068 & 
 .758 & 
 .754
\\   

& SAM-Adapter~\cite{chen2023sam}  &
 4.11 &
 .761 & 
 .746 & 
 .048 & 
 .834 & 
 .796 &
 .822 & 
 .806 & 
 .043 & 
 .836 & 
 .869 &
 .708 & 
 .685 & 
 .059 & 
 .793 & 
 .802

\\   
& SAM2-Adapter~\cite{chen2024sam2}  &4.36
& .776 & .762 & .041 & .831 & .848 & .831 & .828 & \textbf{.042} & .849 & .881 & .706 & .692 & \textbf{.056} & .790 & .810 \\
\hline
\textbf{\multirow{3}{*}{Unified}} 
& VSCode~\cite{luo2024vscode}  & 60  
& .724 
& .706 
& .060 
& .795
& .812

&.834
&.830
&.043
&.851
&.885

&.636
&.608
&.075
&.748
&.753

\\

& EVP~\cite{liu2023explicit}  &  4.95
 & .769 & .750 & .045 & .833 & .836 
 
 &.835  &.832  &.043  &.852  &.878  
 
 & \textbf{.710} & .692 & .057 & .794 & .810 \\


&\cellcolor{iccvblue!20}\textbf{\ourmodel(Ours)} 
& \cellcolor{iccvblue!20}\textbf{4.04} 
& \cellcolor{iccvblue!20}\textbf{.784} 
& \cellcolor{iccvblue!20}\textbf{.780} 
& \cellcolor{iccvblue!20}\textbf{.040} 
& \cellcolor{iccvblue!20}\textbf{.835} 
& \cellcolor{iccvblue!20}\textbf{.852} 
& \cellcolor{iccvblue!20}\textbf{.844} 
& \cellcolor{iccvblue!20}\textbf{.840} 
& \cellcolor{iccvblue!20}\textbf{.042} 
& \cellcolor{iccvblue!20}\textbf{.860} 
& \cellcolor{iccvblue!20}.886
& \cellcolor{iccvblue!20}\textbf{.710} 
& \cellcolor{iccvblue!20}\textbf{.697} 
& \cellcolor{iccvblue!20}\textbf{.056} 
& \cellcolor{iccvblue!20}\textbf{.796} 
& \cellcolor{iccvblue!20}\textbf{.814} \\

\bottomrule
\end{tabular}
\end{center}
\end{table*}

\begin{table*}[h]
\begin{center}

\caption{Generalization performance of related methods on CAMO, COD10K, and NC4K test set. $\uparrow$ / $\downarrow$ represents the higher/lower the score, the better.}\label{tab:supp_cod}
    \scriptsize
    \setlength\tabcolsep{100pt}
    \renewcommand{\arraystretch}{1.2}
    \renewcommand{\tabcolsep}{1.2mm}
\begin{tabular}{c|l|c|ccccc|ccccc|ccccc}
\toprule
\multirow{2}{*}{Task} &
\multirow{2}{*}{Model} & 
\ Update &
\multicolumn{5}{c|}{CAMO} & 
\multicolumn{5}{c|}{NC4K} &
\multicolumn{5}{c}{COD10K}
\\
\cline{4-8}
\cline{9-13}
\cline{14-18}
&      &  {  } Params(M) 

 & $F_\beta^\text{max} \uparrow$ &  $F_\beta^\omega \uparrow$ & $M \downarrow$ &
    $S_\alpha \uparrow$ & $E_{\phi}^\text{m} \uparrow$  &
    
    $F_\beta^\text{max} \uparrow$ &  $F_\beta^\omega \uparrow$ & $M \downarrow$ &
    $S_\alpha \uparrow$ & $E_{\phi}^\text{m} \uparrow$  & 
$F_\beta^\text{max} \uparrow$ &  $F_\beta^\omega \uparrow$ & $M \downarrow$ &
    $S_\alpha \uparrow$ & $E_{\phi}^\text{m} \uparrow$  

\\ 
\hline\hline
\textbf{\multirow{6}{*}{SOD}}

& GateNet~\cite{zhao2020suppress} 
& 128                             

&.573
&.542
&.109
&.666
&.680

&.562
&.529
&.047
&.707
&.724

&.675
&.645
&.066
&.752
&.777

\\

&F3Net~\cite{wei2020f3net}
& 26
&.538
&.506
&.117
&.643
&.657

&.576
&.539
&.047
&.712
&.744

&.661
&.633
&.070
&.738
&.773

\\

& MSFNet~\cite{zhang2021auto}     
& 28                              

&.568
&.535
&.113
&.661
&.682
&.543
&.534
&.052
&.692
&.719

&.671
&.645
&.067
&.747
&.778

\\

& VST~\cite{liu2021visual}        
& 43          

&.484
&.455
&.109
&.636
&.631

&.468
&.430
&.055
&.661
&.670

&.597
&.567
&.072
&.710
&.732

\\

&EDN~\cite{wu2022edn}
&43

&.573
&.542
&.109 
&.666
&.680

&.595
&.562
&.044
&.727
&.756

&.688
&.660
&.063 
&.761
&.795

\\

& ICON~\cite{zhuge2022salient}    & 32

&.520
&.481
&.125
&.641
&.648

&.540
&.502
&.053
&.695
&.715

&.631
&.596
&.076
&.724
&.752

\\
\hline
\textbf{\multirow{9}{*}{COD}}
& SINetV2~\cite{fan2021concealed} 
& 27                              

&.590
&.562
&.102
&.681
&.694

&.609
&.577
&.043
&.729
&.763

&.662
&.639
&.066
&.740
&.769

\\
& PFNet~\cite{mei2021camouflaged} 
& 47  

&.535
&.505
&.110
&.652
&.661

&.556
&.524
&.049
&.699
&.730

&.660
&.633
&.068
&.737
&.769

\\

& ZoomNet~\cite{pang2022zoom}     
& 33  

&.494
&.472
&.113
&.635
&.612

&.520
&.496
&.048
&.488
&.671

&.596
&.576
&.074
&.708
&.706

\\
                             

& FEDER~\cite{he2023camouflaged}  
& 44                    

&.567
&.538
&.106
&.669
&.687

&.636
&.598
&.042
&.749
&.793

&.688
&.664
&.063
&.758
&.790

\\   
& PRNet~\cite{hu2024efficient}  
&   13

&.648
&.607
&.096
&.716
&.766

&.709
&.672
&.059
&.772
&.820

&.650
&.603
&.038
&.756
&.815
\\
   
& ICEG~\cite{he2024strategic}  &
 100 &

 .728 & 
 .697 & 
 .066 & 
 .769 & 
 .820 &

 .735 & 
 .708 & 
 .051 & 
 .786 & 
 .840 &

 .645 & 
 .610 & 
 .035 & 
 .753 &
 .807

\\

& CamoFormer~\cite{yin2024camoformer}  &
 71 &
 .645 & 
 .618 & 
 .078 & 
 .732 & 
 .750 &
 .729 & 
 .707 & 
 .054 & 
 .789 & 
 .822 &
 .668 & 
 .639 & 
 .035 & 
 .770 & 
 .811
\\   

& PGT~\cite{wang2024camouflaged} &
 68 &
 .635 & 
 .612 & 
 .089 & 
 .718 & 
 .730 &
 .729 & 
 .706 & 
 .052 & 
 .791 & 
 .819 &
 .642 & 
 .612 & 
 .036 & 
 .758 &
 .786
\\   

& SAM-Adapter~\cite{chen2023sam}  & 
 4.11 &
 .661 &
 .638 &
 .080 &
 .744 &
 .753 &
 .688 &
 .667 &
 .037 &
 .788 &
 .808 &
 .727 &
 .710 &
 .051 &
 .794 &
 .809 
\\
& SAM2-Adapter~\cite{chen2024sam2}  
& 4.36

&.717
&.692
&.074
&.779
&.807

&.724
&.694
&.044
&.809
&.847

&.735
&.694
&.045
&.819
&.845

\\

\hline
\textbf{\multirow{3}{*}{Unified}}  
& VSCode~\cite{luo2024vscode}  & 60  
&.562
&.532
&.109
&.658
&.678

&.626
&.591
&.043
&.744
&.787

&.684
&.662
&.067
&.753
&.783

\\

& EVP~\cite{liu2023explicit}  &  4.95
& .636 & .637 & .085 & .701 & .718 
& .693 & .694 & .040 & .742 & .775 
& .615 & .614 & .069 & .724 & .749 \\


&\cellcolor{iccvblue!20}\textbf{\ourmodel(Ours)} 
& \cellcolor{iccvblue!20}\textbf{4.04} 
& \cellcolor{iccvblue!20}\textbf{.829} 
& \cellcolor{iccvblue!20}\textbf{.790} 
& \cellcolor{iccvblue!20}\textbf{.049} 
& \cellcolor{iccvblue!20}\textbf{.845} 
& \cellcolor{iccvblue!20}\textbf{.886} 

& \cellcolor{iccvblue!20}\textbf{.794} 
& \cellcolor{iccvblue!20}\textbf{.768} 
& \cellcolor{iccvblue!20}\textbf{.039} 
& \cellcolor{iccvblue!20}\textbf{.839} 
& \cellcolor{iccvblue!20}\textbf{.877} 

& \cellcolor{iccvblue!20}\textbf{.743} 
& \cellcolor{iccvblue!20}\textbf{.700} 
& \cellcolor{iccvblue!20}\textbf{.030} 
& \cellcolor{iccvblue!20}\textbf{.821} 
& \cellcolor{iccvblue!20}\textbf{.869} \\

\bottomrule
\end{tabular}
\end{center}
\end{table*}

\section{Results on Popular COD and SOD Datasets}
\label{Results_on_COD_and_SOD_Datasets}
To further validate the effectiveness and robustness of our method regarding generalizability, we conduct tests on popular SOD datasets (DUTS~\cite{wang2017learning}, HKU-IS~\cite{li2015visual}, and DUT-OMRON~\cite{yang2013saliency}) and COD datasets (CAMO~\cite{le2019anabranch}, COD10K~\cite{fan2020camouflaged}, and NC4K~\cite{lv2021simultaneously}), with all methods uniformly trained using our USC12K dataset. We adopt five metrics that are widely used in COD and SOD tasks~\cite{wang2021salient, fan2021concealed}. These metrics include maximal F-measure ($F_\beta^\text{max}\uparrow$)~\cite{achanta2009frequency}, weighted F-measure ($F_\beta^\omega\uparrow$)~\cite{margolin2014evaluate}, Mean Absolute Error (MAE, $M\downarrow$)~\cite{perazzi2012saliency}, Structural measure (S-measure, $S_\alpha\uparrow$)~\cite{fan2017structure}, and mean Enhanced alignment measure (E-measure, $E_{\phi}^\text{m}\uparrow$)~\cite{fan2018enhanced}. As shown in \tabref{tab:supp_sod} and \tabref{tab:supp_cod}, our~\ourmodel~achieves state-of-the-art performance on these datasets through parameter-efficient fine-tuning. This further confirms the strong capability of our method to accurately identify both salient and camouflaged objects in unconstrained environments. This achievement is attributed to the exceptional versatility of SAM in class-agnostic segmentation tasks and the discriminative ability of our specially designed ARM for distinguishing between salient and camouflaged objects.

\section{More Technical Details}
\label{More_Technical_Details}
All models are retrained using the training set of USC12K with an input image resolution of 352$\times$352. Horizontal flipping and random cropping are applied for data augmentation. The experiments are conducted in PyTorch on one NVIDIA L40 GPU. For our model, we use the hiera-large version of SAM2 following the SAM2-Adapter~\cite{chen2024sam2}. AdamW optimizer is used with a warm-up strategy and linear decay strategy. The initial learning rate is set to 0.0001. The batch size is set to 24, and the maximum number of epochs is set to 90.

\noindent\textbf{Backbone of models.}
The models compared can be divided into two categories based on their papers: one is full-tuning models, and the other is parameter-efficient fine-tuning (PEFT) models. (i)Full Tuning models: Include all SOD and COD methods and VSCode in the Unified Method. For fairness, the models compared are all trained according to the configurations specified in their original papers. (ii)PEFT models: SAM-Adapter, SAM2-Adapter, EVP in the Unified Method and our model.
The backbone architectures across various models consist of several types. For full tuning, VST employs a transformer encoder based on T2T-ViT~\cite{yuan2021tokens}, while SINet-V2 utilizes Res2Net-50~\cite{gao2019res2net}. VSCode uses Swin-T~\cite{liu2021swin}, and ICEG adopts Swin-B~\cite{liu2021swin}. PRNet is based on the SMT backbone~\cite{lin2023scale}, and both CamoDiffusion, CamoFormer, and PGT use PVTv2-b4~\cite{wang2021pvtv2}. Other models generally rely on ResNet-50~\cite{he2016deep} with pre-trained weights from ImageNet~\cite{deng2009imagenet}. In the case of PEFT models, EVP uses SegFormer-B4~\cite{xie2021segformer} as its base, SAM-Adapter uses the default ViT-H version of SAM~\cite{kirillov2023segment}, and both SAM2-Adapter and our model employ the  hiera-large version of SAM2~\cite{ravi2024sam}.

\noindent\textbf{Training and Inference.} For traditional SOD and COD models: USC12K is defined by three attributes: saliency, camouflage, and background. Conventional methods for COD and SOD are crafted for dichotomous mapping tasks and don't seamlessly transition to the nuanced demands of the USC12K benchmark. Inspired by seminal works in semantic segmentations~\cite{long2015fully,strudel2021segmenter}, we retool the output layers of our models to yield a tripartite representation for saliency, camouflage, and background. This is achieved by harnessing a softmax layer to generate a predictive mapping. We employ a cross-entropy loss function to refine the model, which is congruent with our overarching methodological framework. For unified models: VSCode and EVP, which require task-specific prompts for each dataset, we create two copies of the USC12K training set. One copy is used for SOD, with the ground truth being the SOD-only mask, and is used to train the prompts corresponding to the SOD task. The other copy is used for COD, with the ground truth being the COD-only mask, and is used to train the prompts corresponding to the COD task. VSCode is trained once using all 16,800 images (two copies of 8,400 images), while EVP is trained twice on the two separate training sets (each containing 8,400 images) to obtain the two task-specific prompts.
During inference, all unified models perform inference on the testing set of USC12K twice, with the corresponding prompt enabled for each task. The first inference run generates the SOD results, and the second inference run generates the COD results. The final prediction is obtained by merging the SOD and COD predictions. For overlapping pixels, the attribute with the higher prediction value between the two tasks is chosen as the final attribute for that pixel.

\section{More USC12K Dataset Detail and Examples}
\label{More_USC12K_Dataset_Examples}

\begin{figure}[t]
\centering
\includegraphics[width=0.85\linewidth]{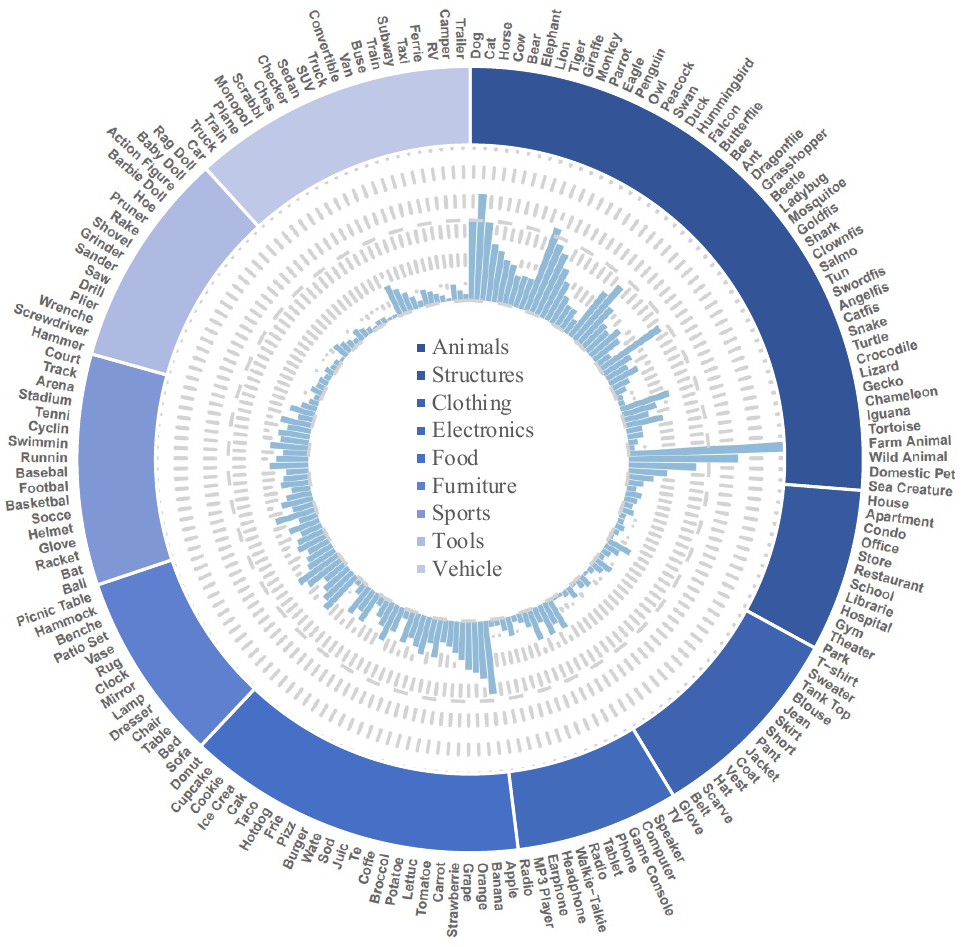}
\caption{The data source and distribution of different data types.
}
\label{fig:data_category}
\end{figure}

\noindent\textbf{Object category distribution.} We obtain an initial coarse classification using CLIP~\cite{radford2021learning}, followed by manual verification and refinement. Except for images collected from COD10K~\cite{fan2020camouflaged}, which already include camouflage object category labels, all other objects require classification. Then we assign category labels to each image, covering 9 super-classes and 179 sub-classes. Figure~\ref{fig:data_category} illustrates the class breakdown of our USC12K dataset.

\noindent\textbf{Object number distribution.} Our USC12K dataset contains images with different numbers of objects. For clarify, we have counted the distribution of images with different numbers of objects in USC12K, as shown in the following \tabref{tab:supp_USC12K}.

\begin{table}[h!]
    \centering
    \small
    \caption{\small
    Distribution of Images with Different Numbers of Objects in USC12K.
    }\label{tab:supp_USC12K}
    \renewcommand{\arraystretch}{1.0}
    \setlength\tabcolsep{9.0pt}
    \begin{tabular}{l|cccc}
        \toprule
         \textbf{Number of objects} & \textbf{0} & \textbf{1} & \textbf{2} & \textbf{$>$2} \\
         \hline
         Number of images & 3000 & 4197 & 2335 & 2468 \\

        \bottomrule  
    \end{tabular}
\end{table}

\noindent\textbf{Detail of annotation process.} For Scene A and B, we retained their original annotations, while Scene D did not require additional annotation. Therefore, we focus here on detailing the annotation process for Scene C.
\begin{itemize}
\item \textbf{Initial Determination of Object Attributes:} We invited 7 observers to perform the initial identification of salient and camouflaged objects in the images. A voting process was used to determine the salient and camouflaged objects in each image, with objects and their attributes receiving more than half of the votes being retained. We then used Photoshop to apply red boxes for salient objects and green boxes for camouflaged objects, which served as the reference for the subsequent mask annotation step.
\item \textbf{Mask Annotation:} We invited 9 volunteers to perform detailed mask annotation for the dataset using the ISAT interactive annotation tool~\cite{ISAT_with_segment_anything}, which supports SAM semi-automatic labeling.
\item \textbf{Annotation Quality Control:} After annotation, we invited an additional 3 observers to review and refine the results. Masks with imprecise or incorrect annotations were manually corrected.
\end{itemize}

\begin{figure*}[h]
\centering
\includegraphics[width=0.95\textwidth]{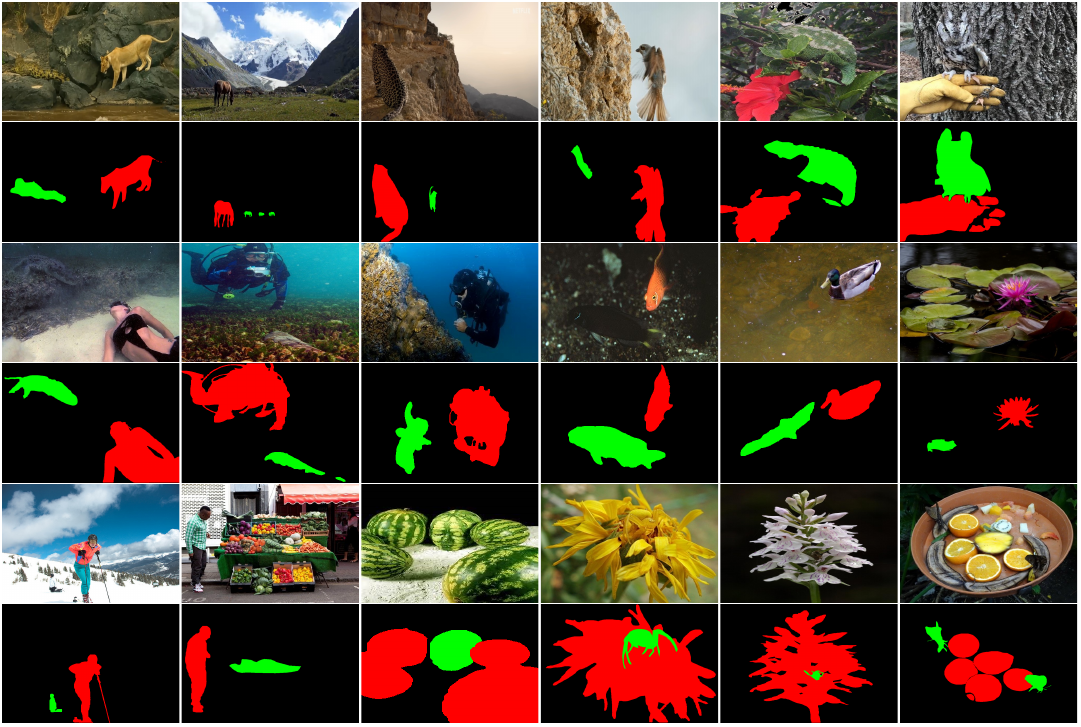}
    \put(-480,25){\rotatebox{90}{\footnotesize{GT
}}}
    \put(-480,70){\rotatebox{90}{\footnotesize{Image
}}}
    \put(-480,130){\rotatebox{90}{\footnotesize{GT
}}}
        \put(-480,170){\rotatebox{90}{\footnotesize{Image
}}}
    \put(-480,230){\rotatebox{90}{\footnotesize{GT
}}}
        \put(-480,280){\rotatebox{90}{\footnotesize{Image
}}}

\caption{Additional Example images where exist both camouflaged and salient objects from the USC12K dataset. Our collection comprises 3,000 carefully curated and annotated images, encompassing a diverse range of scenes and categories. \textbf{Please zoom in for an better view}.}\label{fig:supp-add-examples}
\end{figure*}

\noindent\textbf{More USC12K examples.}
In Figure~\ref{fig:supp-add-examples}, we illustrate a selection of images from the USC12K dataset, each featuring both salient and camouflaged objects. The main difference between our USC12K dataset and existing SOD and COD datasets is that it includes a curated subset of 3,000 images, each featuring both salient and camouflaged objects. We invest significant time and effort in finding and annotating these images. Our dataset spans an extensive variety of environments, including, but not limited to, terrestrial, aquatic, alpine, sylvan, and urban ecosystems, and encompasses a broad spectrum of categories, such as lion, flower and various fruit species. This dataset is designed to assist the SOD and COD research communities in advancing the state-of-the-art in discerning more sophisticated saliency and camouflage patterns.

\section{Additional Qualitative Results}
\label{Additional_Qualitative_Results}
We present additional predictive results of our~\ourmodel~model compared to other COD and SOD models in the USC12K test set. As illustrated in Figure~\ref{fig:supp_add_visual_comp}, our model outperforms its competitors. Specifically, across four different scenes, our model demonstrates a high degree of consistency with the ground truth, especially in distinguishing between salient and camouflaged objects. Our model is adept at learning distinctive features of saliency and camouflage. For instance, it can accurately identify patterns such as camouflaged humans (refer to the fifth column of Figure~\ref{fig:supp_add_visual_comp}). Moreover, in scenes devoid of salient or camouflaged objects, our model remains unaffected by complex backgrounds (refer to the sixth column of Figure~\ref{fig:supp_add_visual_comp}). This further underscores the robustness and accuracy of our~\ourmodel~model.

\begin{table}[!t]
\centering
\caption{Performance of different base models. *In the original SAM or SAM2, we only fine-tune the mask decoder.}
\label{tab:base_model_ablation}
\scriptsize
\setlength\tabcolsep{1pt}
\renewcommand{\arraystretch}{1.2}
\renewcommand{\tabcolsep}{2.1mm}

\begin{tabular}{l|l|c|c|c|ccc}
\toprule
\multirow{1}{*}{Method} &
\multirow{1}{*}{Base} &
\multirow{1}{*}{Para.} &
\multirow{1}{*}{$\text{IoU}_S$} & 
\multirow{1}{*}{$\text{IoU}_C$} & 
\multirow{1}{*}{mIoU} &
\multirow{1}{*}{mAcc} &
\multirow{1}{*}{CSCS}
\\ 
\hline\hline
SAM*  & SAM & 3.92 & 51.07 & 33.00 & 59.56  & 68.73  & 18.66  \\
\cellcolor{iccvblue!20}\ourmodel & \cellcolor{iccvblue!20}SAM & \cellcolor{iccvblue!20}4.08  & \cellcolor{iccvblue!20}73.93 & \cellcolor{iccvblue!20}56.50 & \cellcolor{iccvblue!20}75.87 & \cellcolor{iccvblue!20}83.86 & \cellcolor{iccvblue!20}8.24 \\
\hline
SAM2*  & SAM2 & 4.22  & 66.42 & 44.02 & 68.78  & 77.65  & 11.58  \\
\cellcolor{iccvblue!20}\textbf{\ourmodel} & \cellcolor{iccvblue!20}SAM2 & \cellcolor{iccvblue!20}4.04 & \cellcolor{iccvblue!20}\textbf{75.57} & \cellcolor{iccvblue!20}\textbf{61.34} & \cellcolor{iccvblue!20}\textbf{78.03} & \cellcolor{iccvblue!20}\textbf{87.92} & \cellcolor{iccvblue!20}\textbf{7.49} \\
\bottomrule
\end{tabular}
\end{table}

\section{Additional Ablation Study}
\label{Additional_Ablation_Study}

\textbf{Performance of Different Base Models.} We conducted ablation experiments to evaluate the performance of different base models, as presented in Table \ref{tab:base_model_ablation}. First, as shown in the first two and last two rows of the table, our model demonstrates significant performance improvements on the USC12K benchmark, regardless of whether SAM~\cite{kirillov2023segment}
(default vit-huge version) or SAM2~\cite{ravi2024sam} (default hiera-large version) is used as the base model. For instance, when using SAM as the base model, our method achieves a 16.31\% gain in mIoU compared to the original SAM, while utilizing SAM2 results in a 9.25\% improvement in mIoU over the original SAM2. Additionally, transitioning from SAM to SAM2 (as shown in rows 2 and 4) results in performance gains across all metrics with fewer fine-tuned parameters.

\begin{figure*}[!t]
    \centering
    \renewcommand{\arraystretch}{0.2}
    \begin{tabular}{@{}c@{\hskip 5pt}c@{\hskip 0.9pt}c@{\hskip 0.9pt}c@{\hskip 0.9pt}c@{\hskip 0.9pt}c@{\hskip 0.9pt}c@{}}

        \raisebox{0.3cm}{\makebox[0pt][c]{\rotatebox{90}{\footnotesize \textit{Image}}}} &
        \includegraphics[width=2.25cm, height=1.53cm]{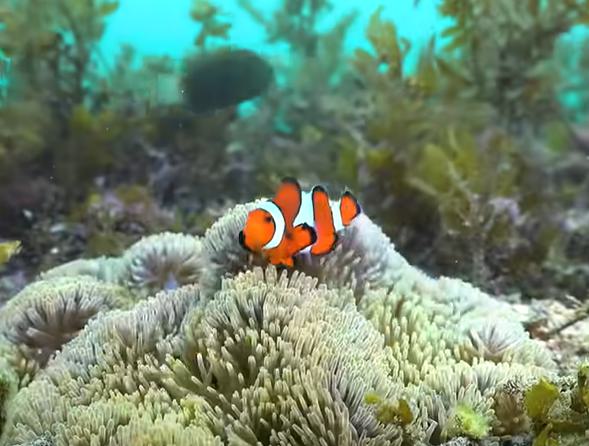} &
        \includegraphics[width=2.25cm, height=1.53cm]{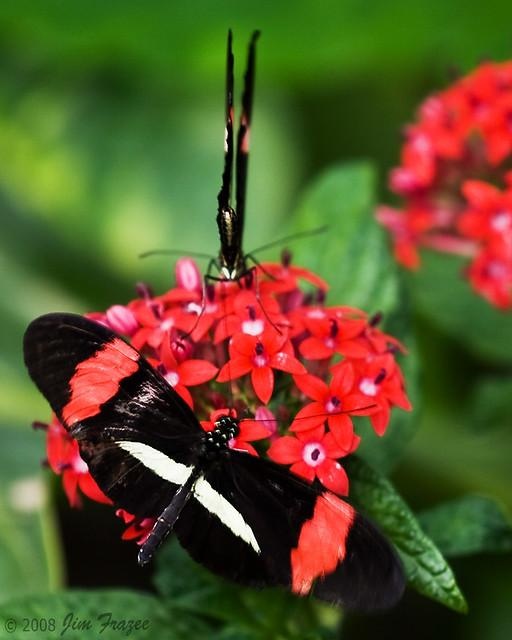} &
        \includegraphics[width=2.25cm, height=1.53cm]{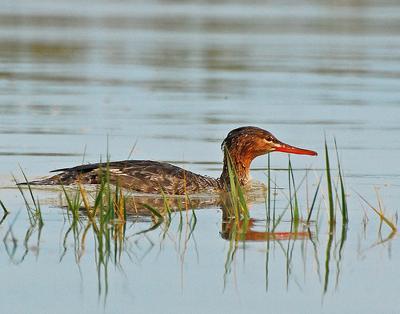} &
        \includegraphics[width=2.25cm, height=1.53cm]{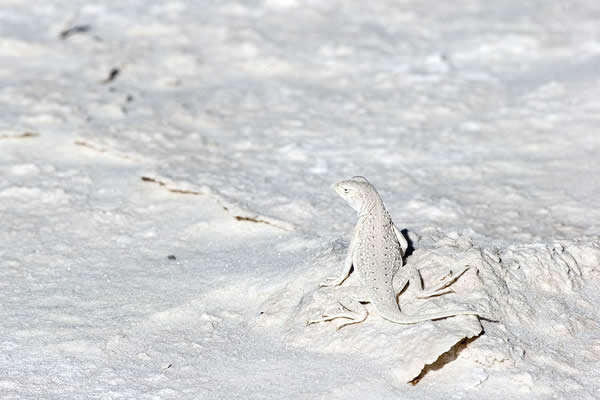} &
        \includegraphics[width=2.25cm, height=1.53cm]{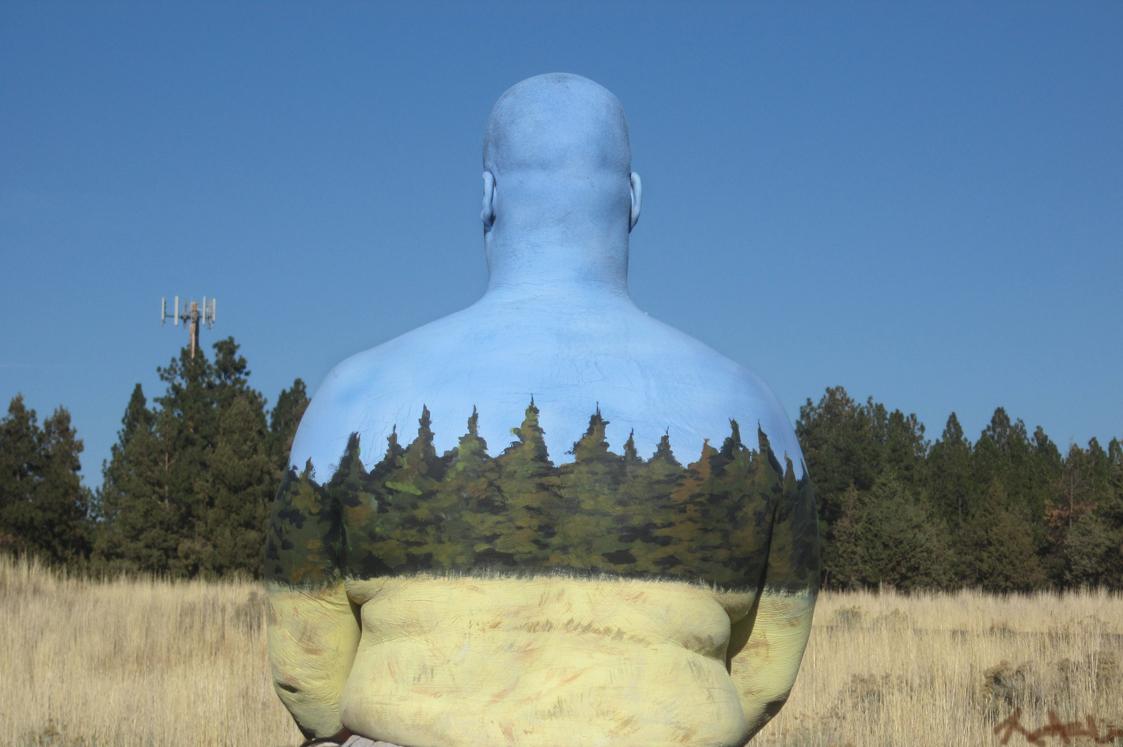} &
        \includegraphics[width=2.25cm, height=1.53cm]{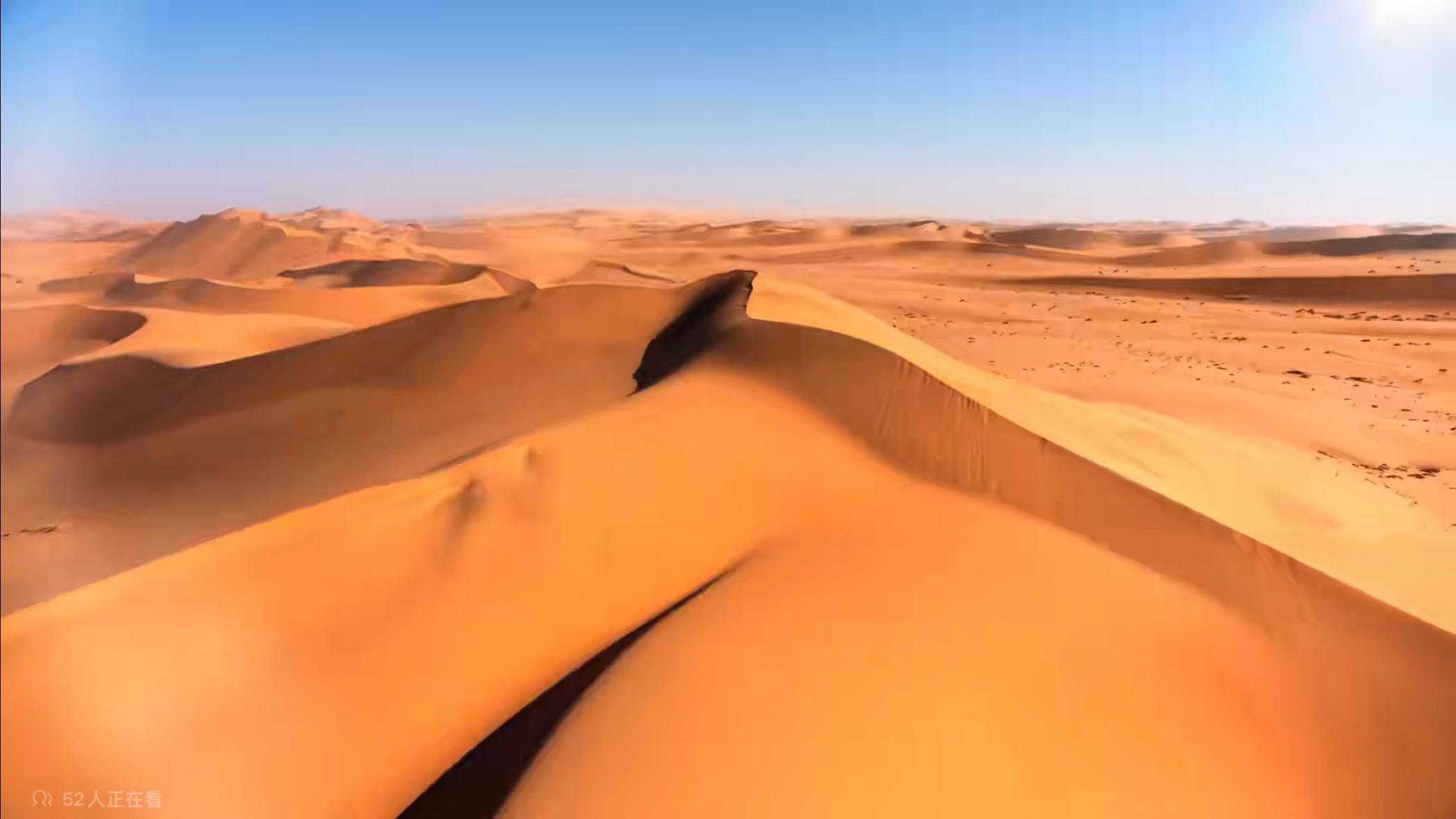}  \\
        
        \raisebox{0.5cm}{\makebox[0pt][c]{\rotatebox{90}{\footnotesize \textit{GT}}}} &
        \includegraphics[width=2.25cm, height=1.53cm]{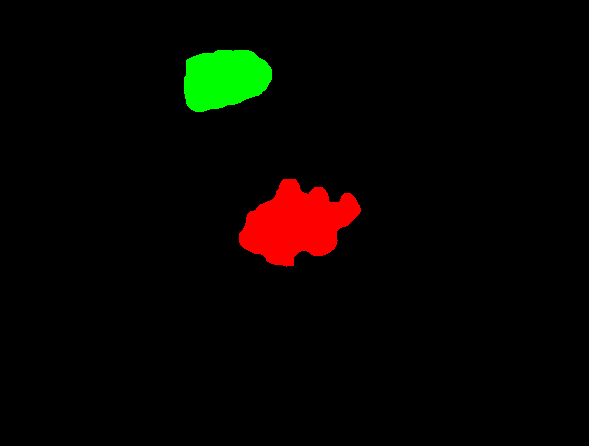} &
        \includegraphics[width=2.25cm, height=1.53cm]{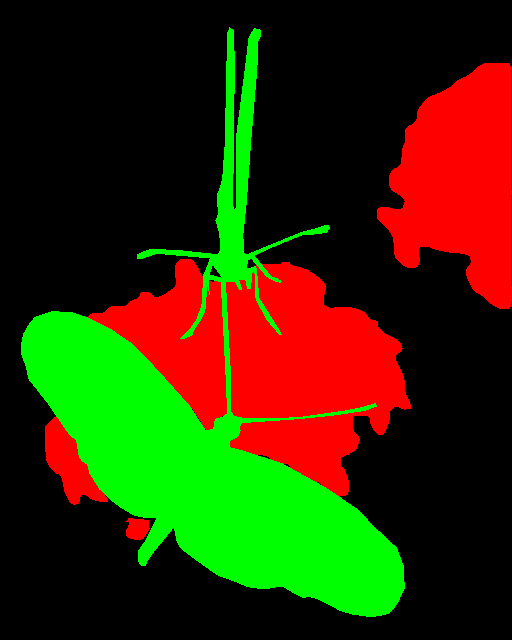} &
        \includegraphics[width=2.25cm, height=1.53cm]{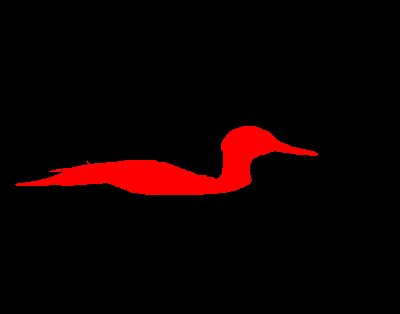} &
        \includegraphics[width=2.25cm, height=1.53cm]{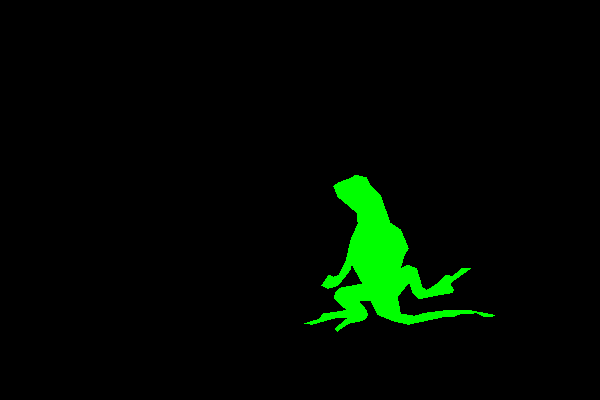} &
        \includegraphics[width=2.25cm, height=1.53cm]{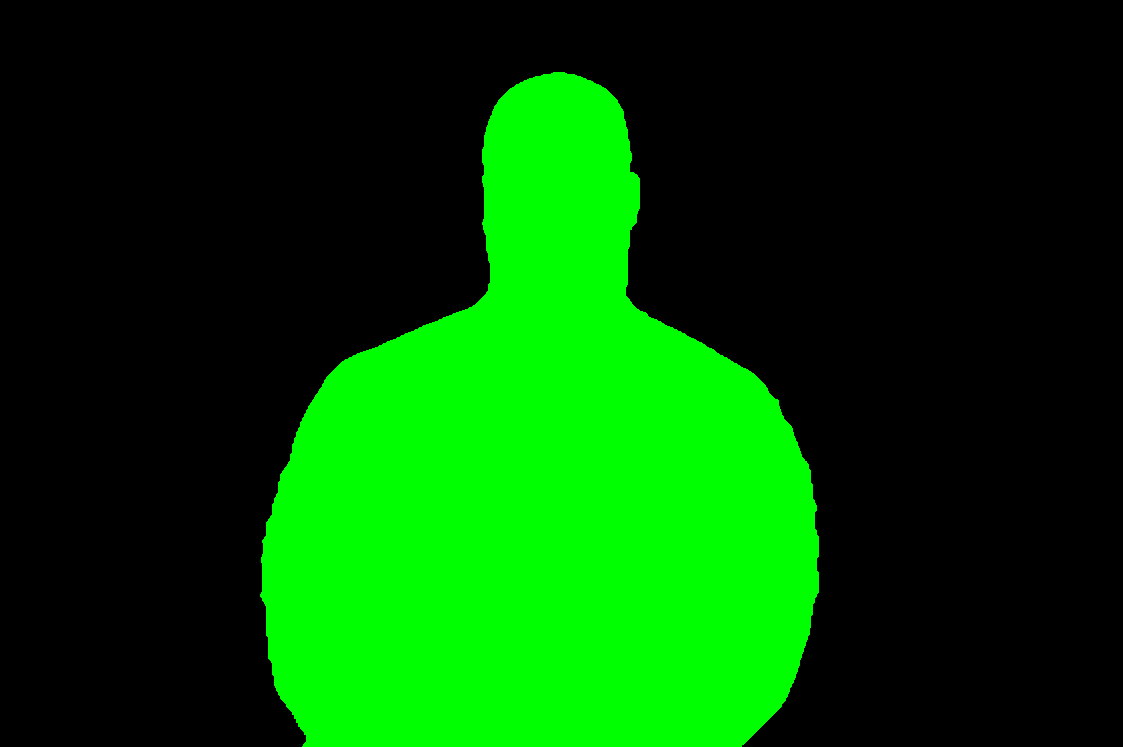} &
        \includegraphics[width=2.25cm, height=1.53cm]{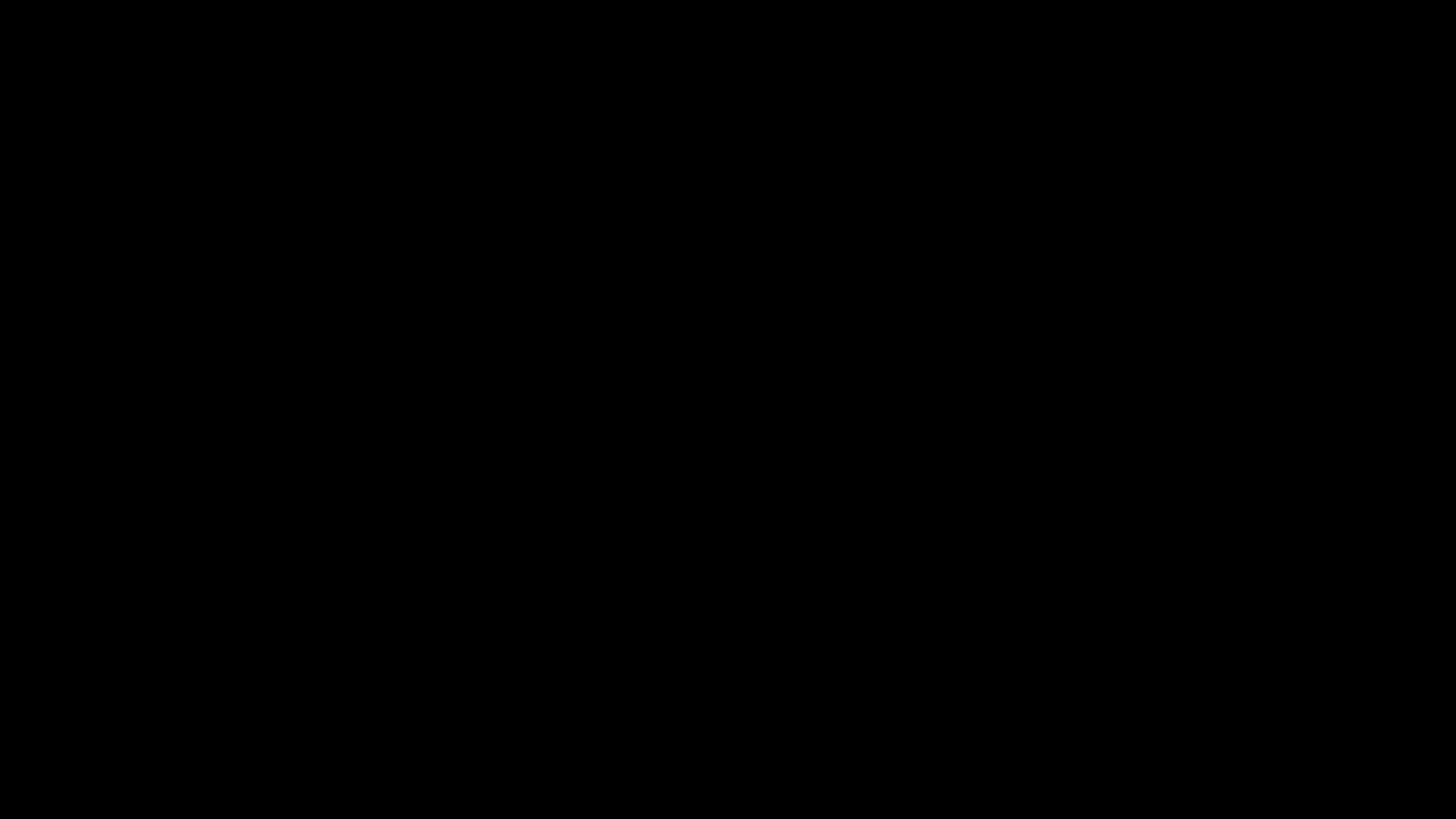} \\

        \raisebox{0.2cm}{\makebox[0pt][c]{\rotatebox{90}{\footnotesize \textbf{\textit{\ourmodel}}}}} &
        \includegraphics[width=2.25cm, height=1.53cm]{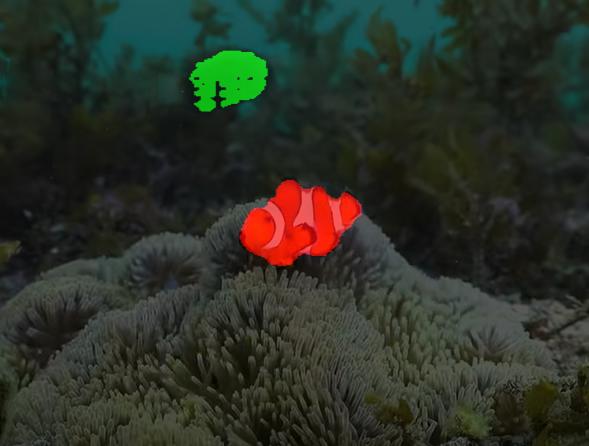} &
        \includegraphics[width=2.25cm, height=1.53cm]{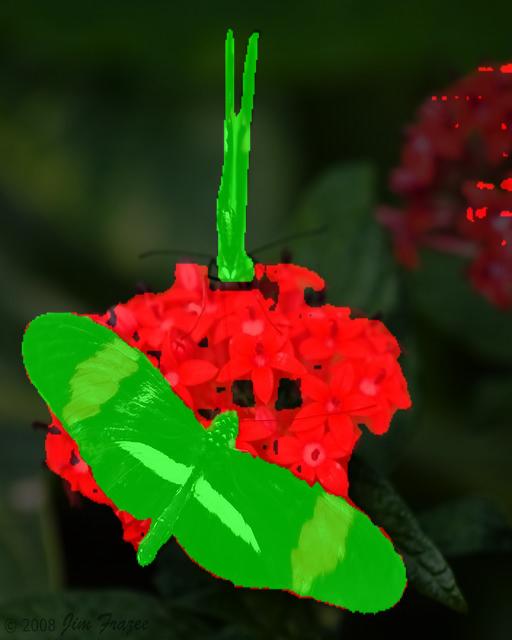} &
        \includegraphics[width=2.25cm, height=1.53cm]{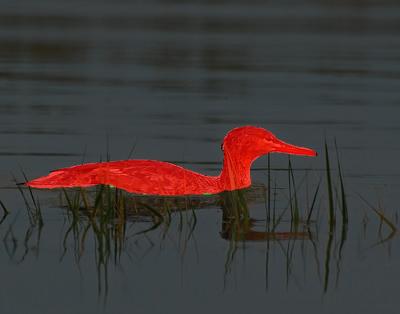} &
        \includegraphics[width=2.25cm, height=1.53cm]{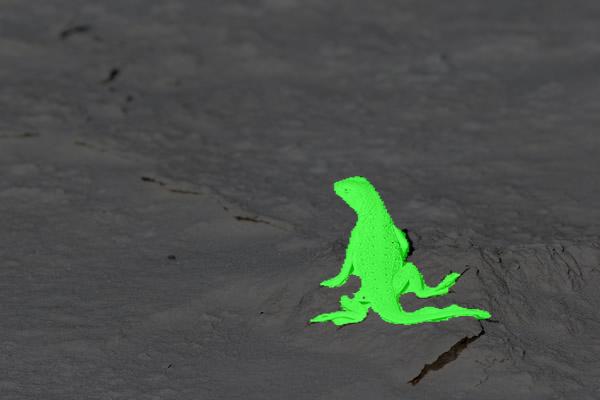} &
        \includegraphics[width=2.25cm, height=1.53cm]{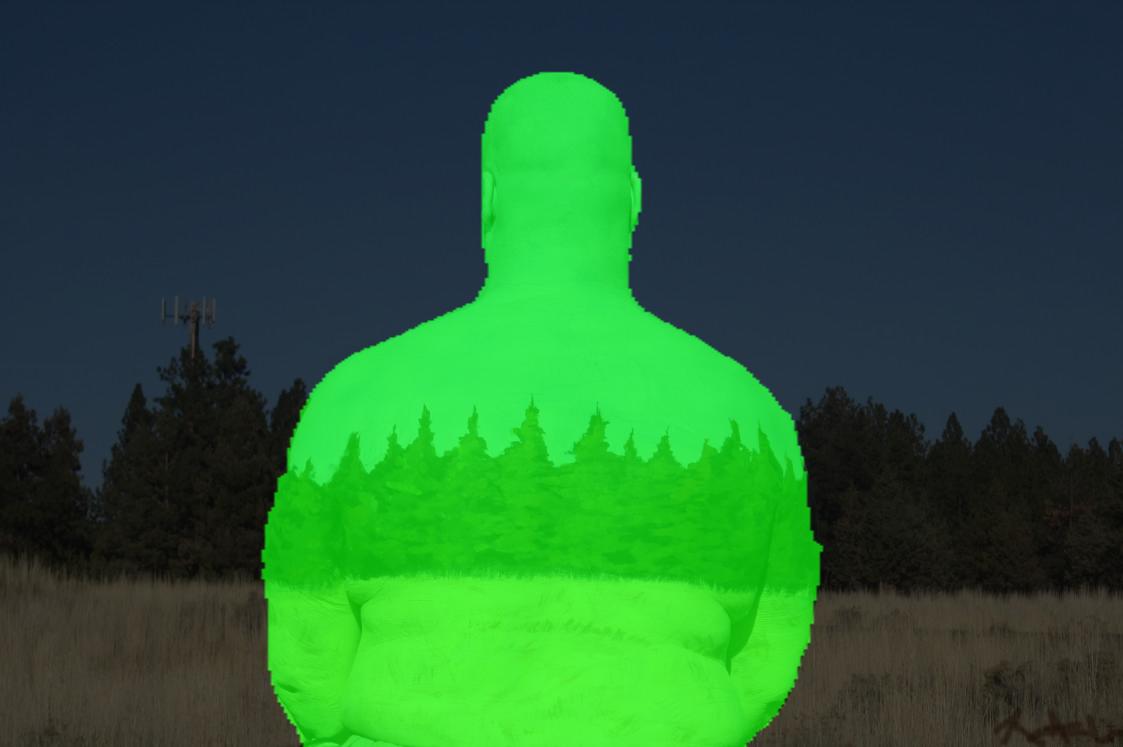} &
        \includegraphics[width=2.25cm, height=1.53cm]{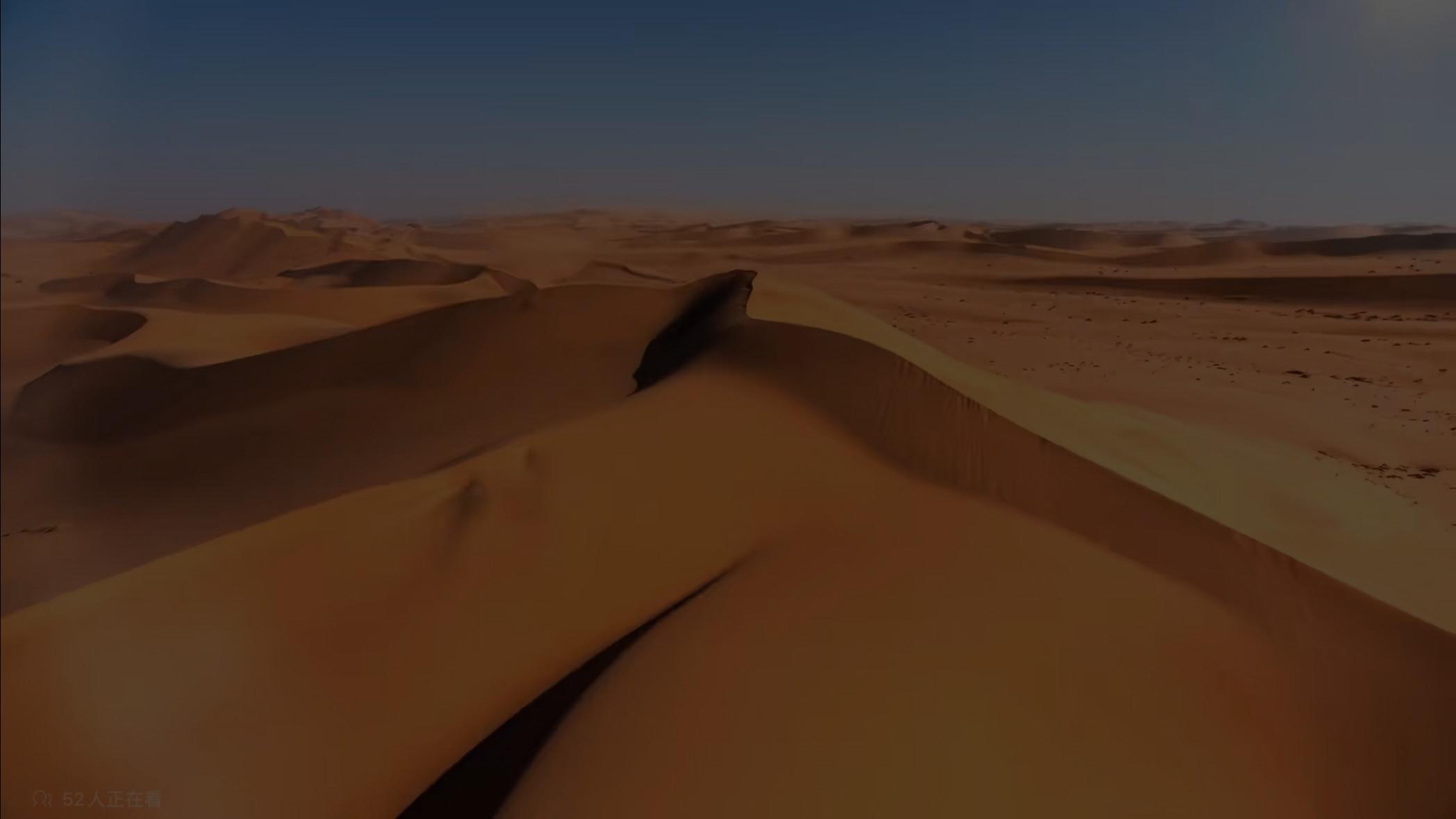} \\

        \raisebox{0.3cm}{\makebox[0pt][c]{\rotatebox{90}{\footnotesize \textit{PRNet}}}} &
        \includegraphics[width=2.25cm, height=1.53cm]{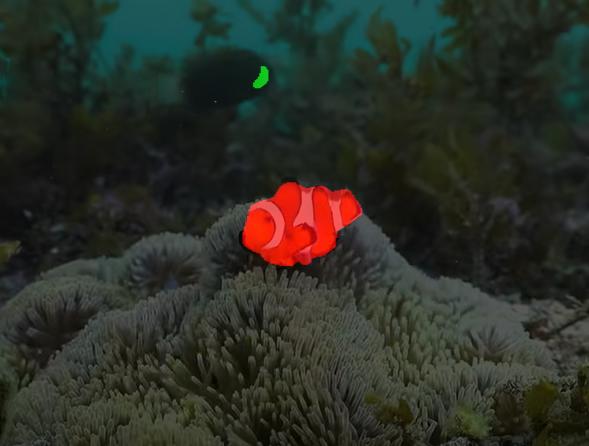} &

        \includegraphics[width=2.25cm, height=1.53cm]{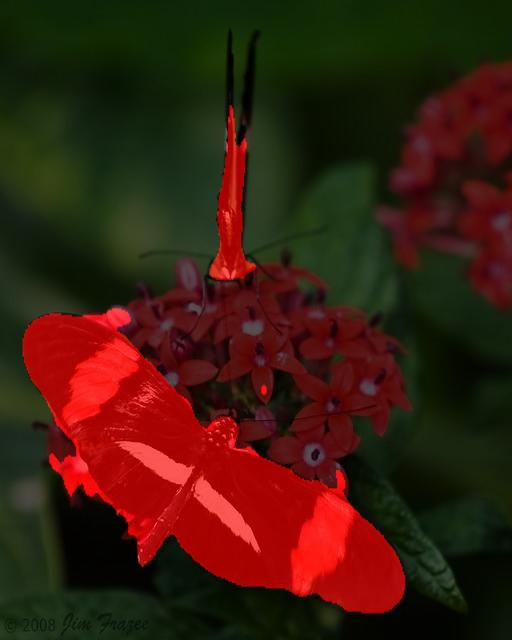} &
                \includegraphics[width=2.25cm, height=1.53cm]{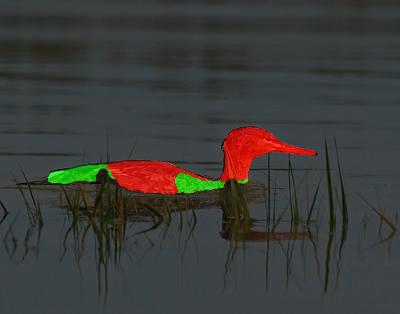} &
        \includegraphics[width=2.25cm, height=1.53cm]{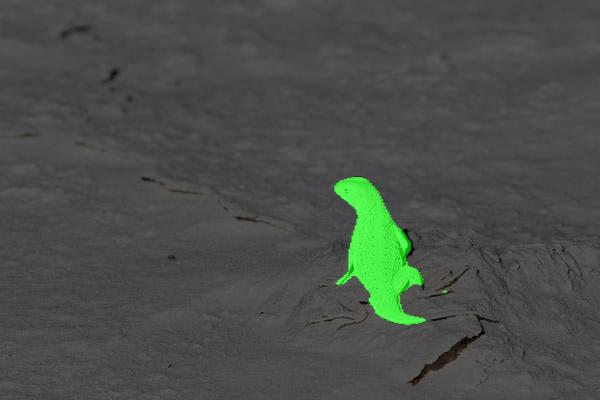} &
        \includegraphics[width=2.25cm, height=1.53cm]{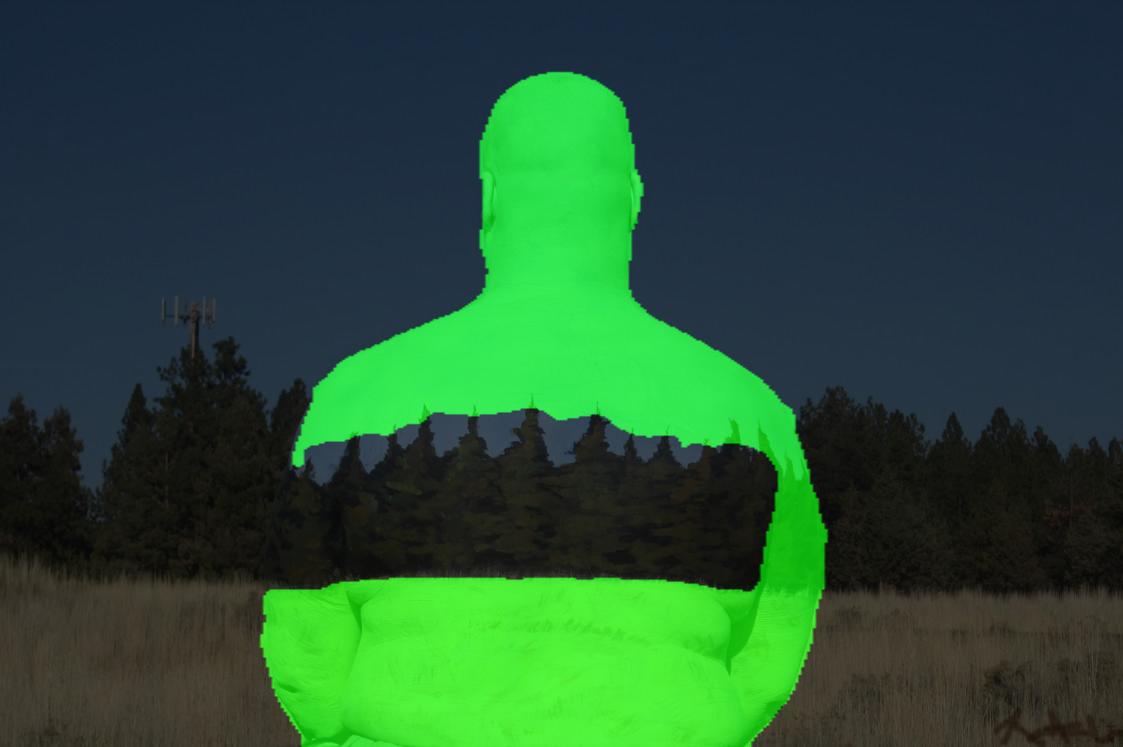} &
        \includegraphics[width=2.25cm, height=1.53cm]{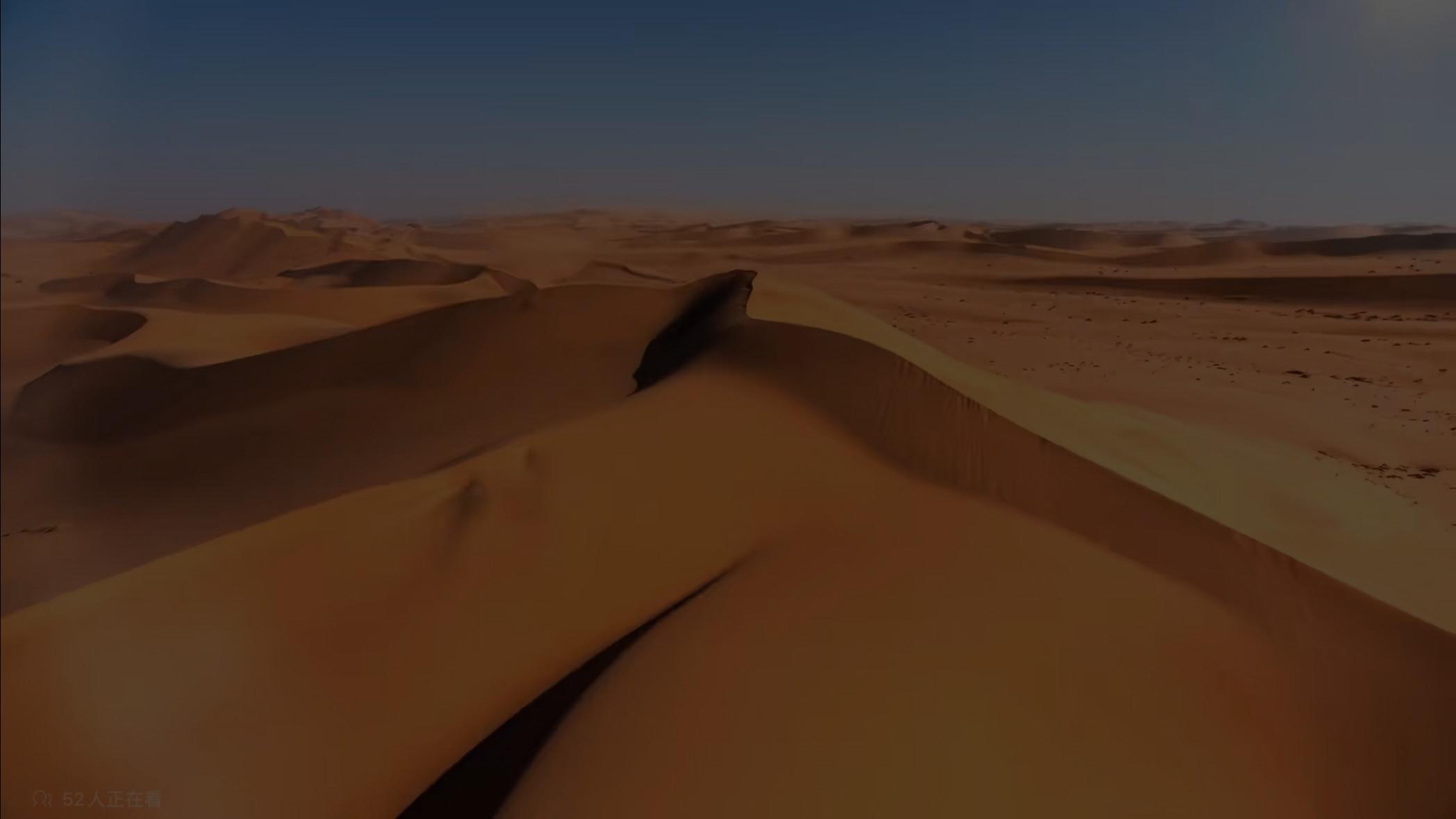} \\

        \raisebox{0.4cm}{\makebox[0pt][c]{\rotatebox{90}{\footnotesize \textit{ICON}}}} &
        \includegraphics[width=2.25cm, height=1.53cm]{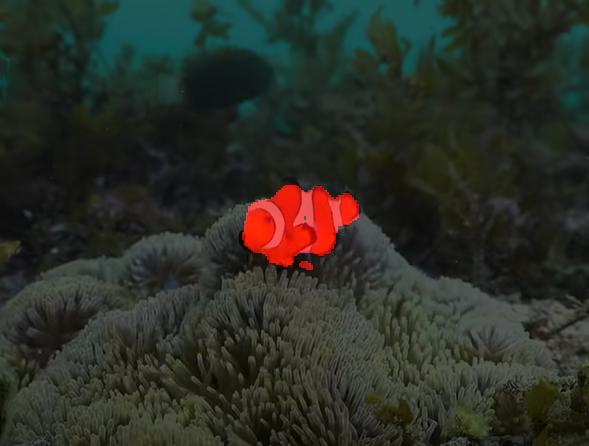} &
        \includegraphics[width=2.25cm, height=1.53cm]{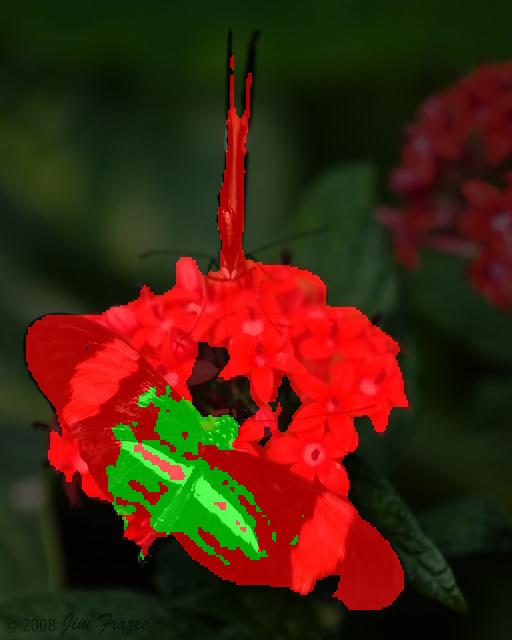} &
         \includegraphics[width=2.25cm, height=1.53cm]{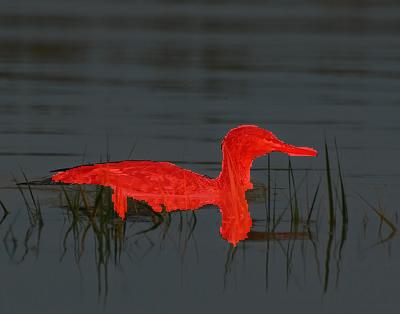} &
        \includegraphics[width=2.25cm, height=1.53cm]{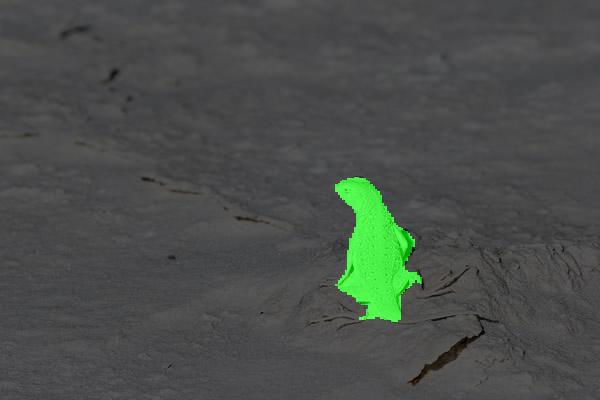} &
        \includegraphics[width=2.25cm, height=1.53cm]{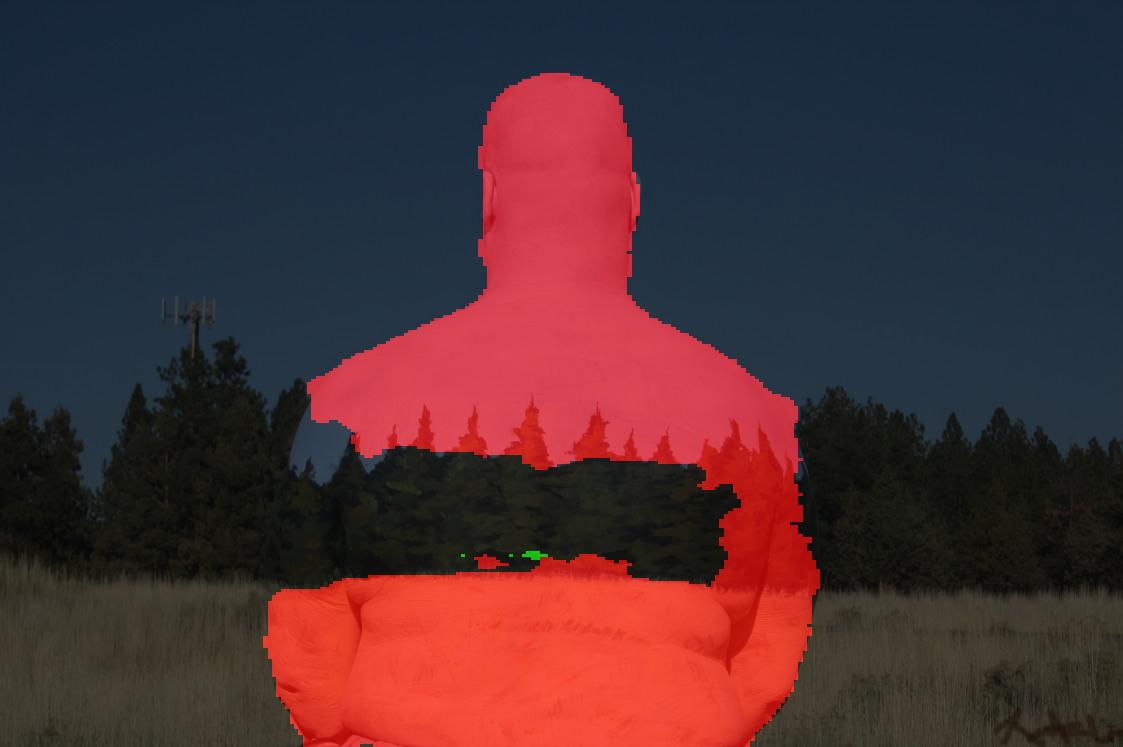} &
        \includegraphics[width=2.25cm, height=1.53cm]{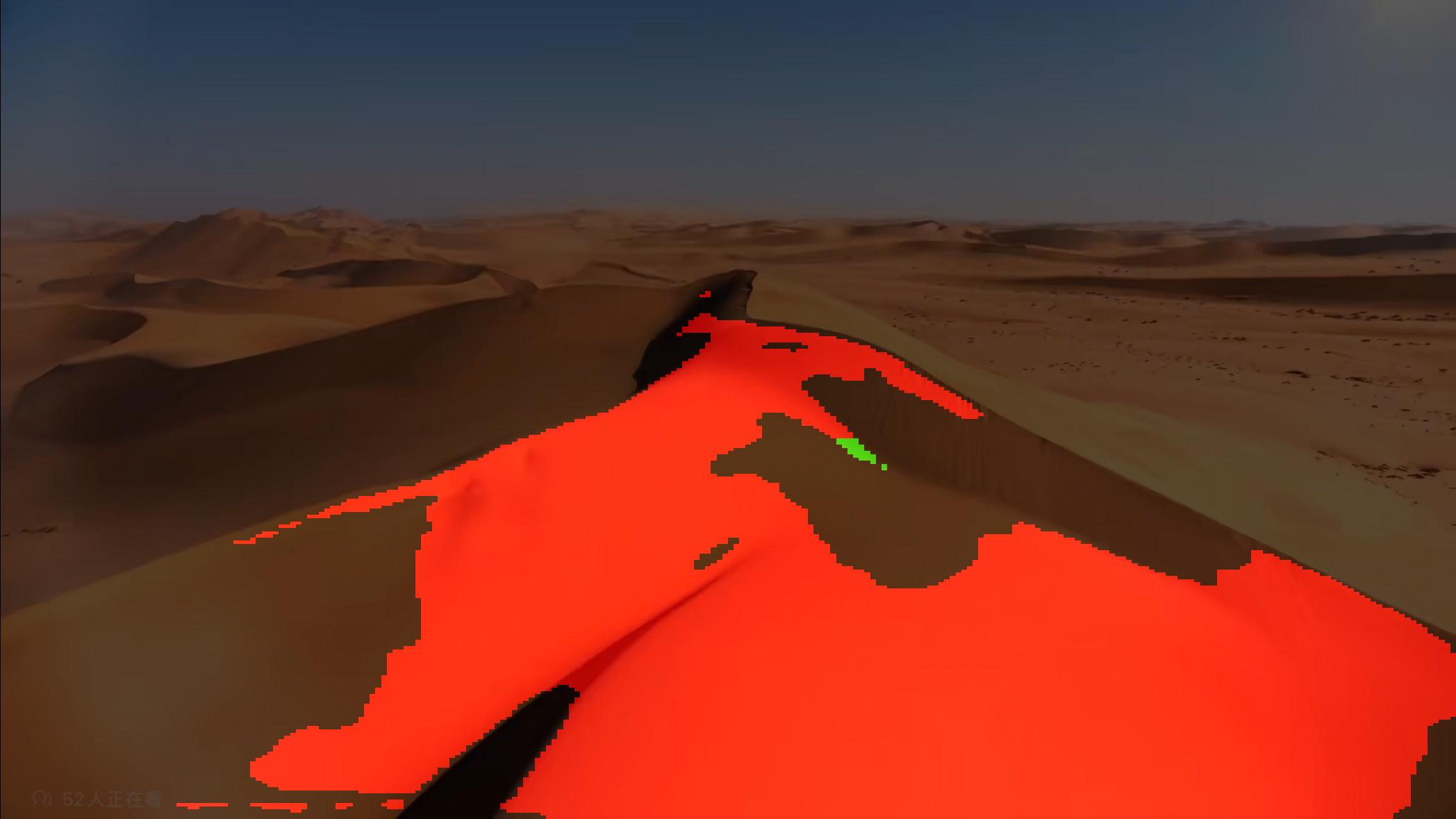} \\
        
        \raisebox{0.4cm}{\makebox[0pt][c]{\rotatebox{90}{\footnotesize \textit{ICEG}}}} &
        \includegraphics[width=2.25cm, height=1.53cm]{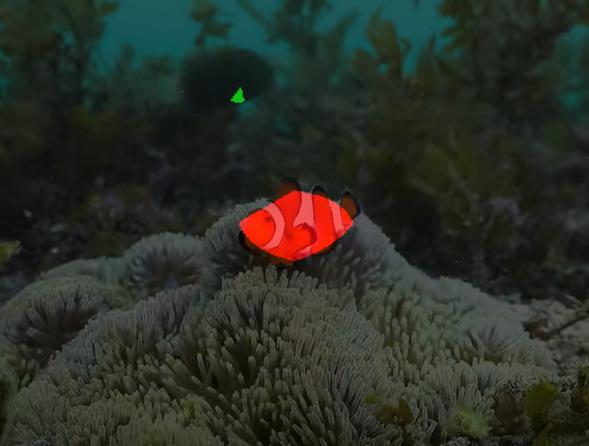} &

        \includegraphics[width=2.25cm, height=1.53cm]{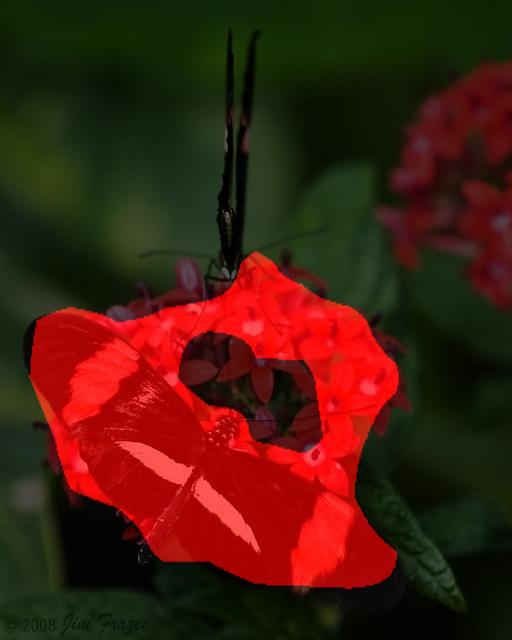} &
               \includegraphics[width=2.25cm, height=1.53cm]{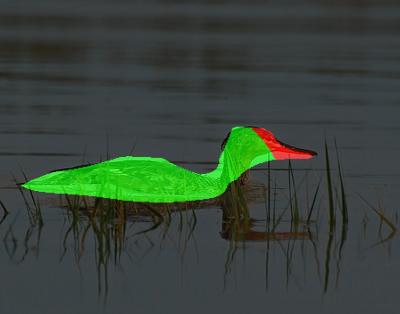} & 
        \includegraphics[width=2.25cm, height=1.53cm]{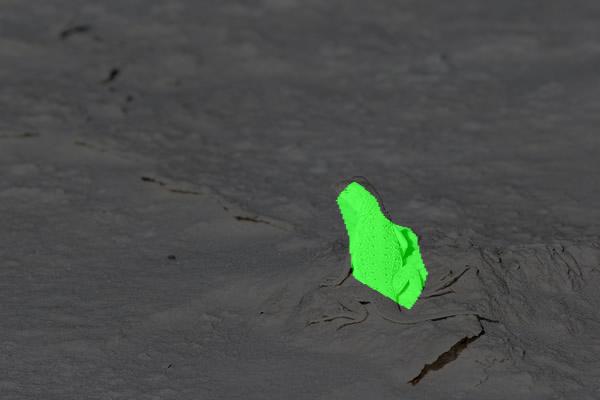} &
        \includegraphics[width=2.25cm, height=1.53cm]{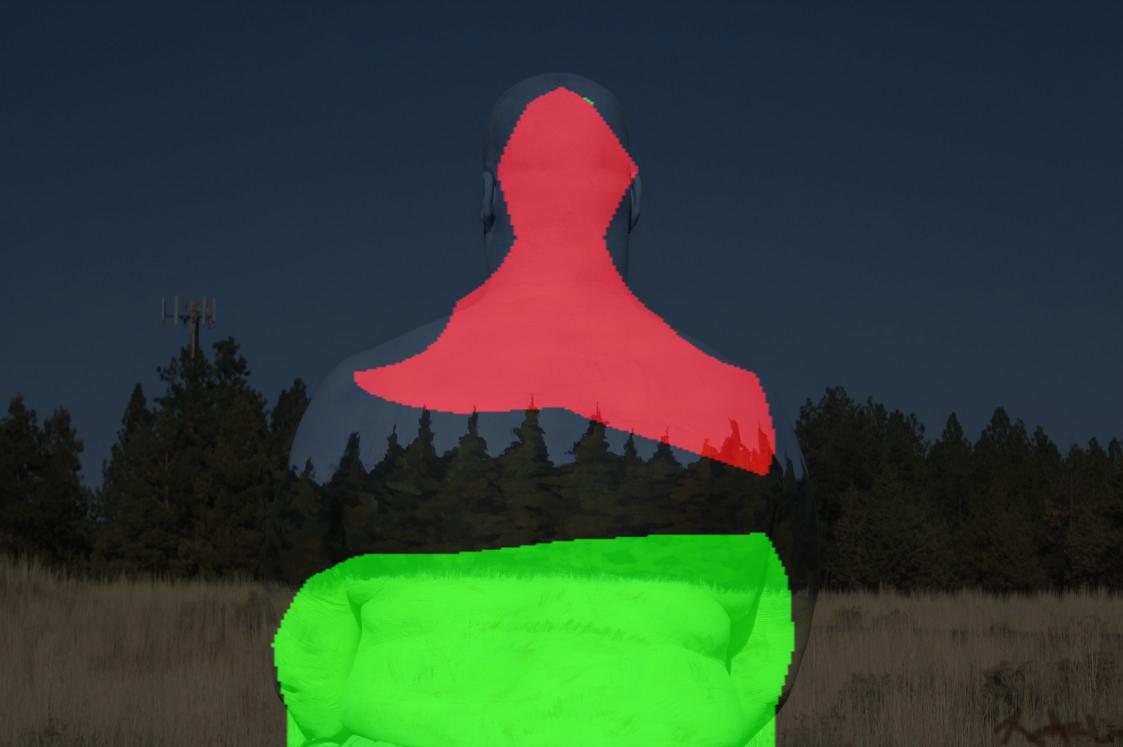} &
        \includegraphics[width=2.25cm, height=1.53cm]{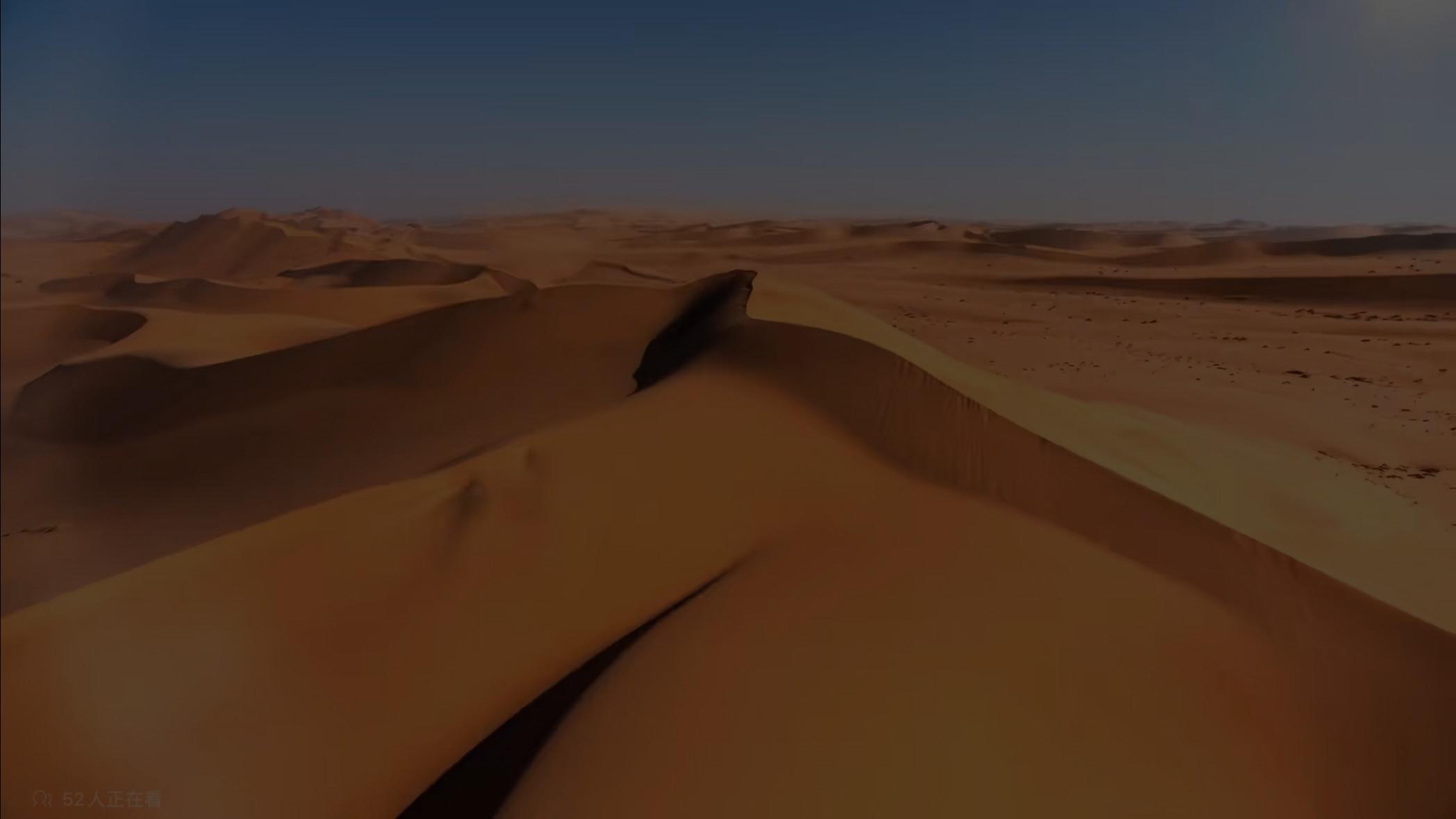} \\

        \raisebox{0.3cm}{\makebox[0pt][c]{\rotatebox{90}{\footnotesize \textit{FEDER}}}} &
        \includegraphics[width=2.25cm, height=1.53cm]{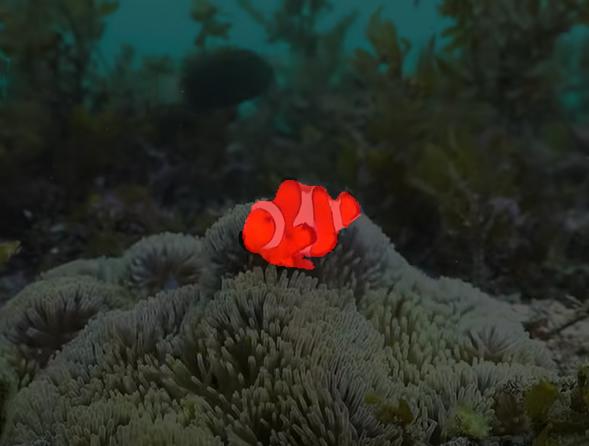} &

        \includegraphics[width=2.25cm, height=1.53cm]{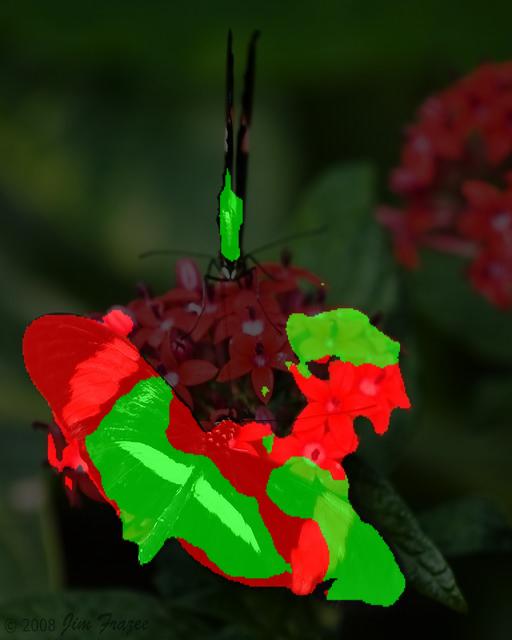} &
                \includegraphics[width=2.25cm, height=1.53cm]{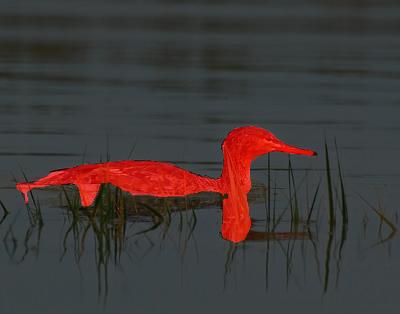} &
        \includegraphics[width=2.25cm, height=1.53cm]{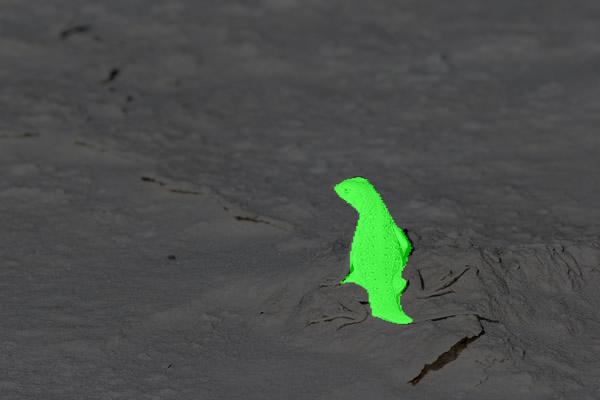} &
        \includegraphics[width=2.25cm, height=1.53cm]{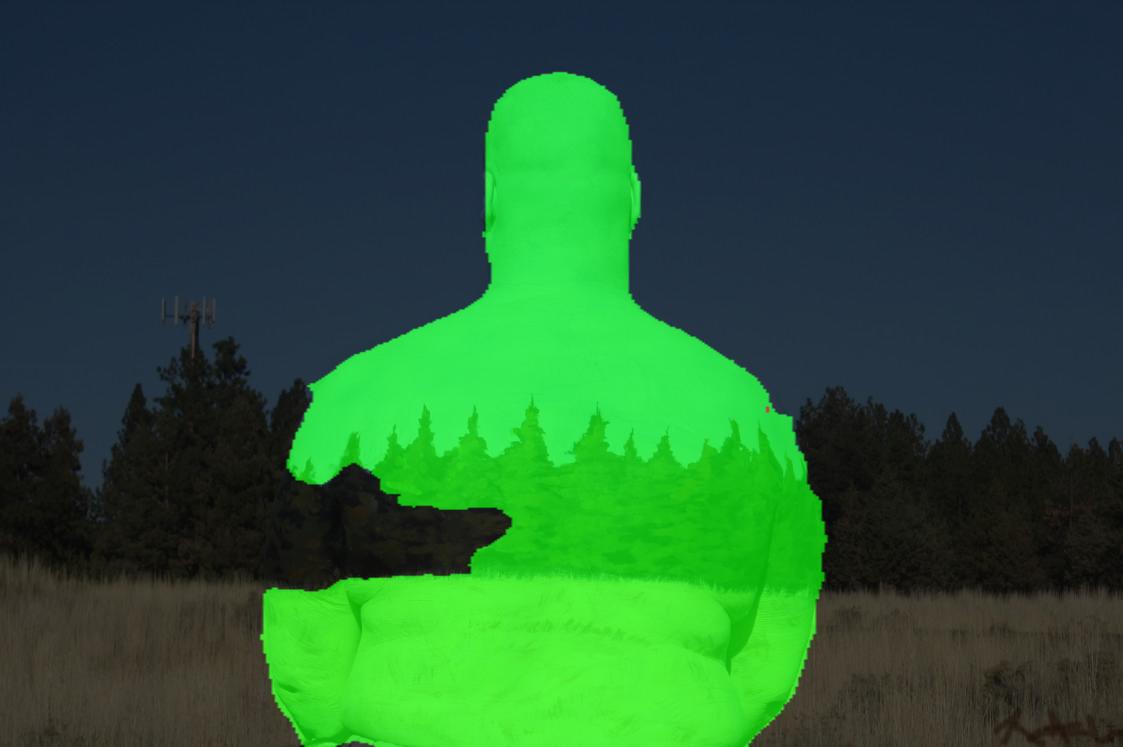} &
        \includegraphics[width=2.25cm, height=1.53cm]{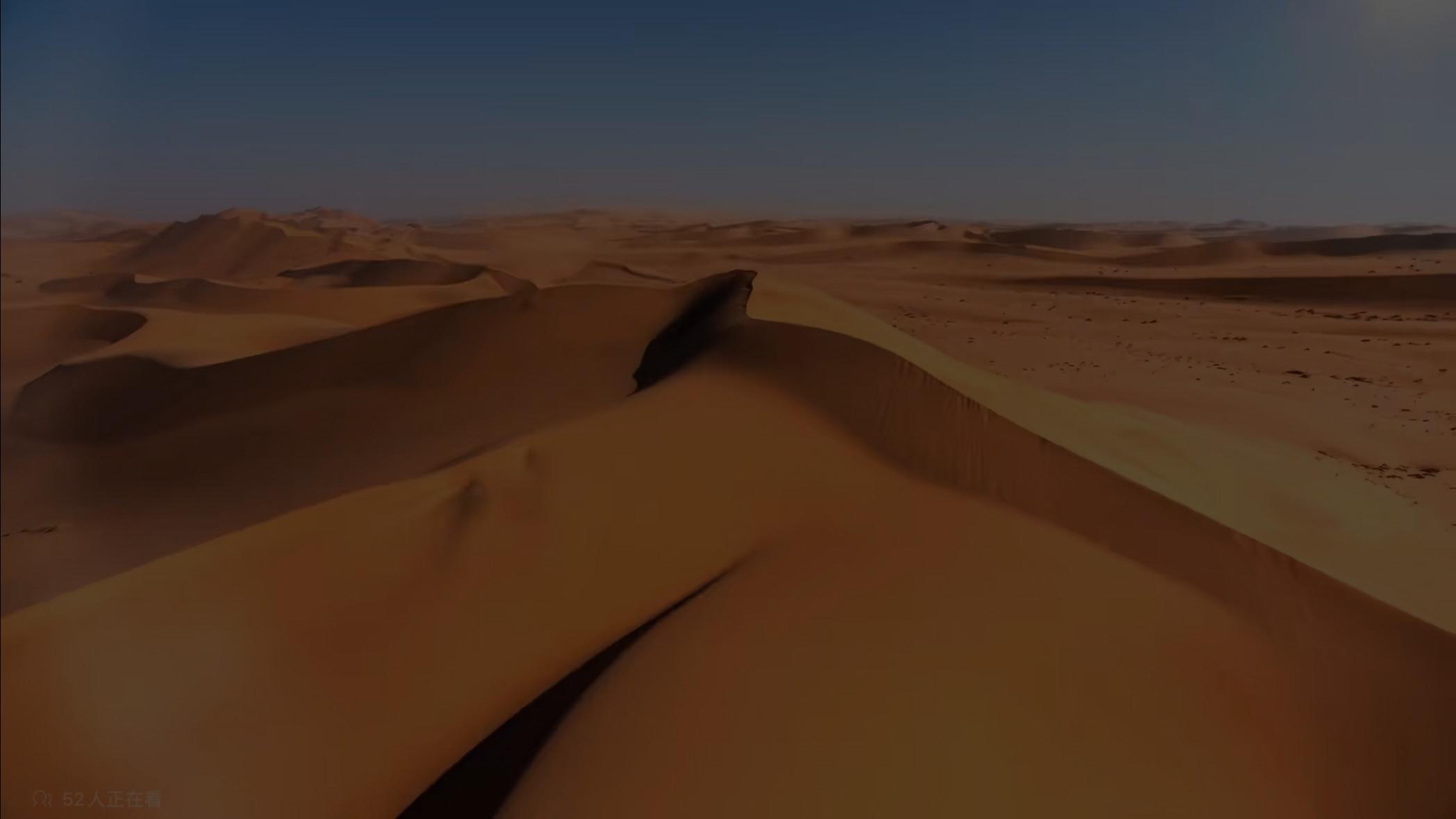} \\

        \raisebox{0.2cm}{\makebox[0pt][c]{\rotatebox{90}{\footnotesize \textit{ZoomNet}}}} &
        \includegraphics[width=2.25cm, height=1.53cm]{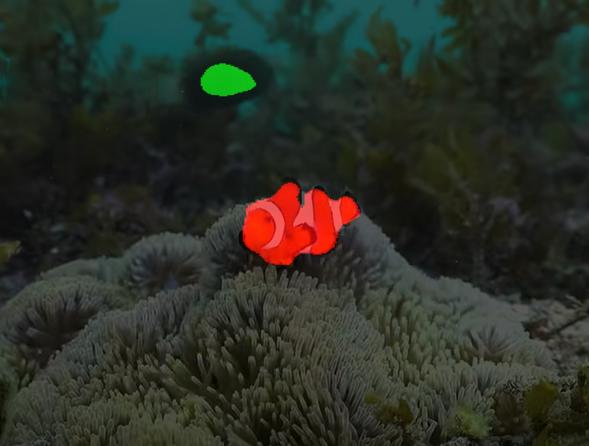} &

        \includegraphics[width=2.25cm, height=1.53cm]{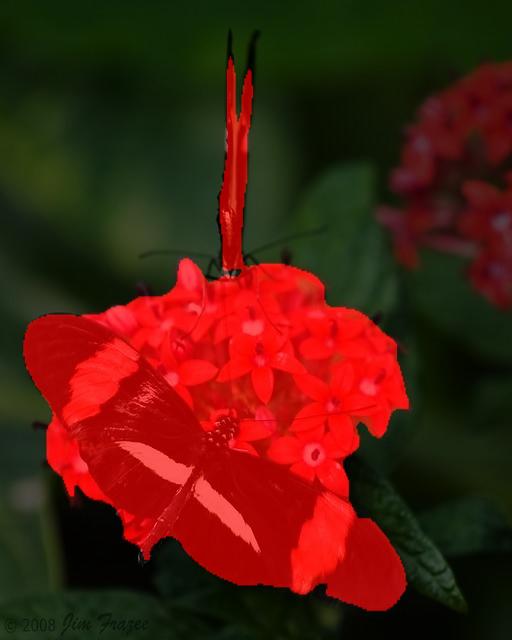} &
                \includegraphics[width=2.25cm, height=1.53cm]{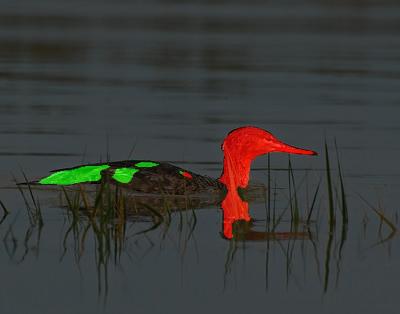} &
        \includegraphics[width=2.25cm, height=1.53cm]{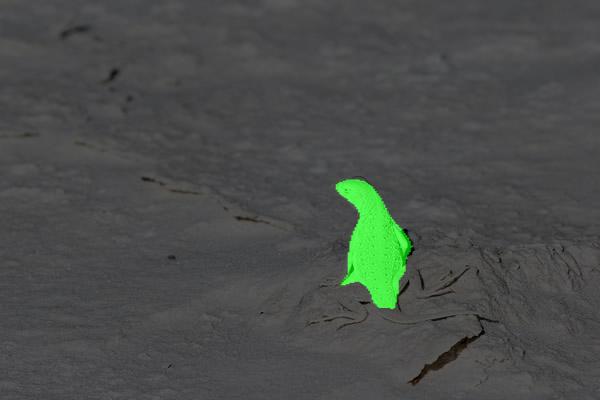} &
        \includegraphics[width=2.25cm, height=1.53cm]{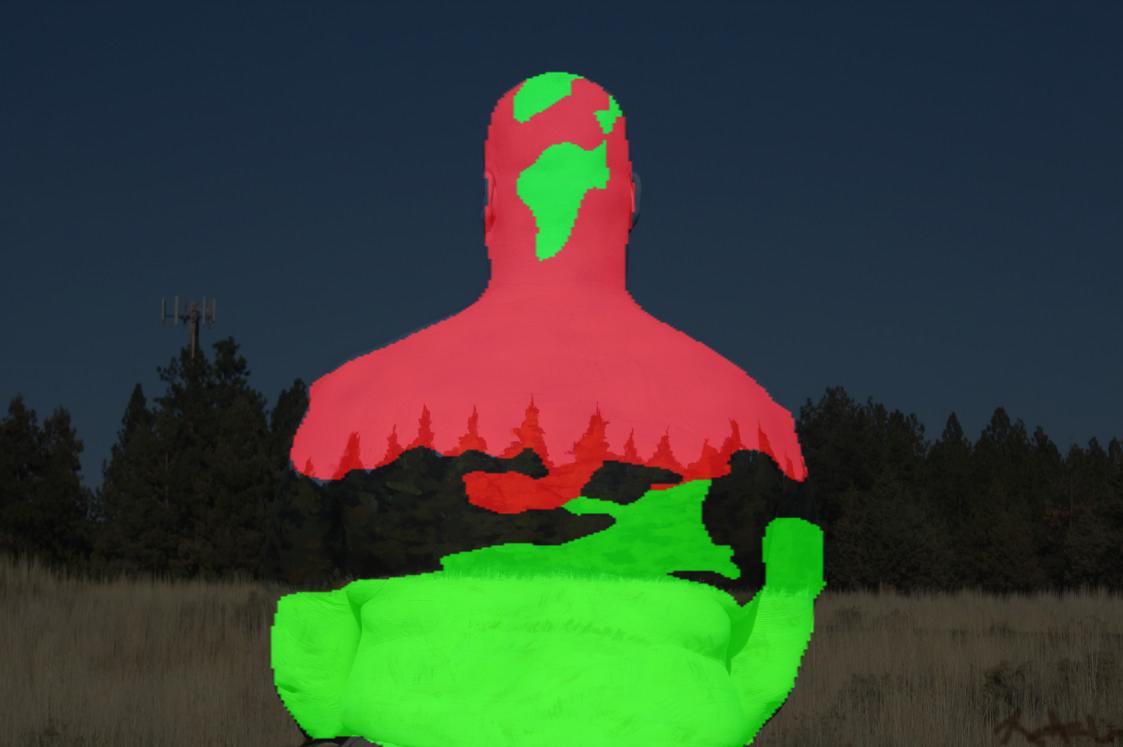} &
        \includegraphics[width=2.25cm, height=1.53cm]{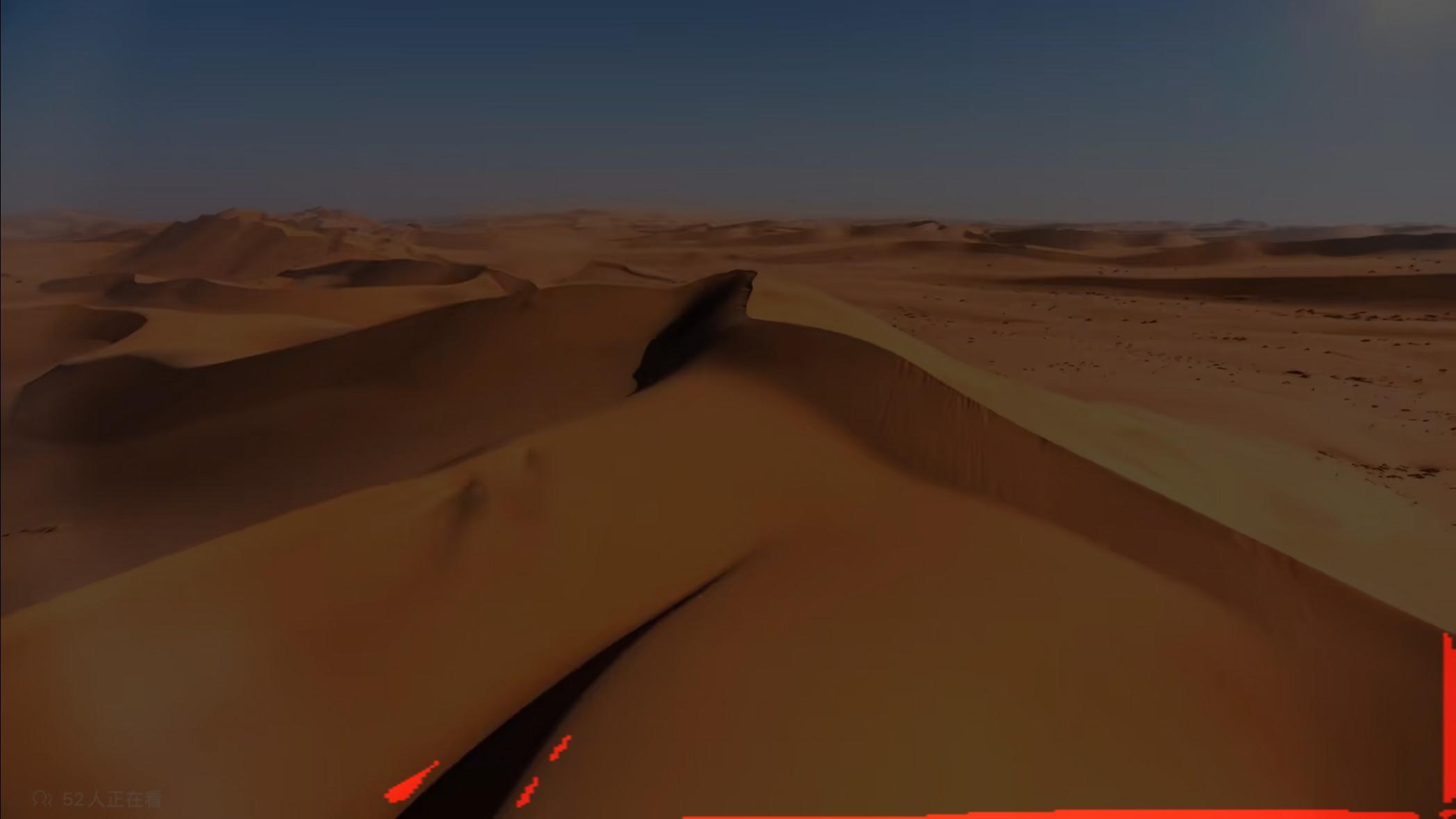} \\

        \raisebox{0.2cm}{\makebox[0pt][c]{\rotatebox{90}{\footnotesize \textit{SINet-V2}}}} &
        \includegraphics[width=2.25cm, height=1.53cm]{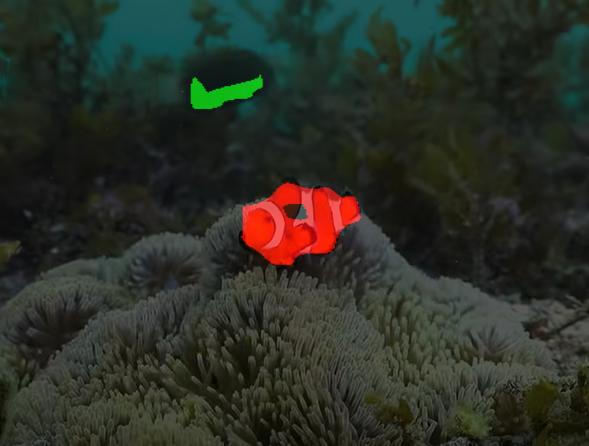} &

        \includegraphics[width=2.25cm, height=1.53cm]{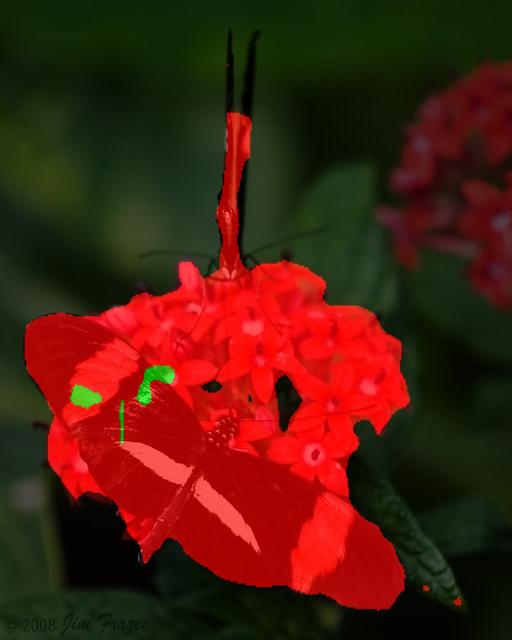} &
                \includegraphics[width=2.25cm, height=1.53cm]{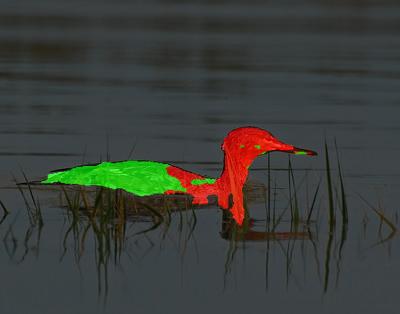} &
        \includegraphics[width=2.25cm, height=1.53cm]{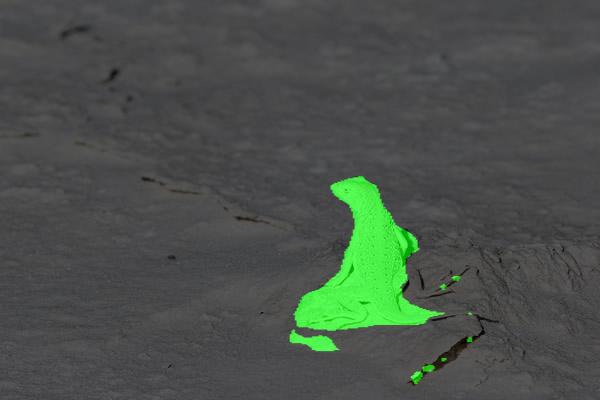} &
        \includegraphics[width=2.25cm, height=1.53cm]{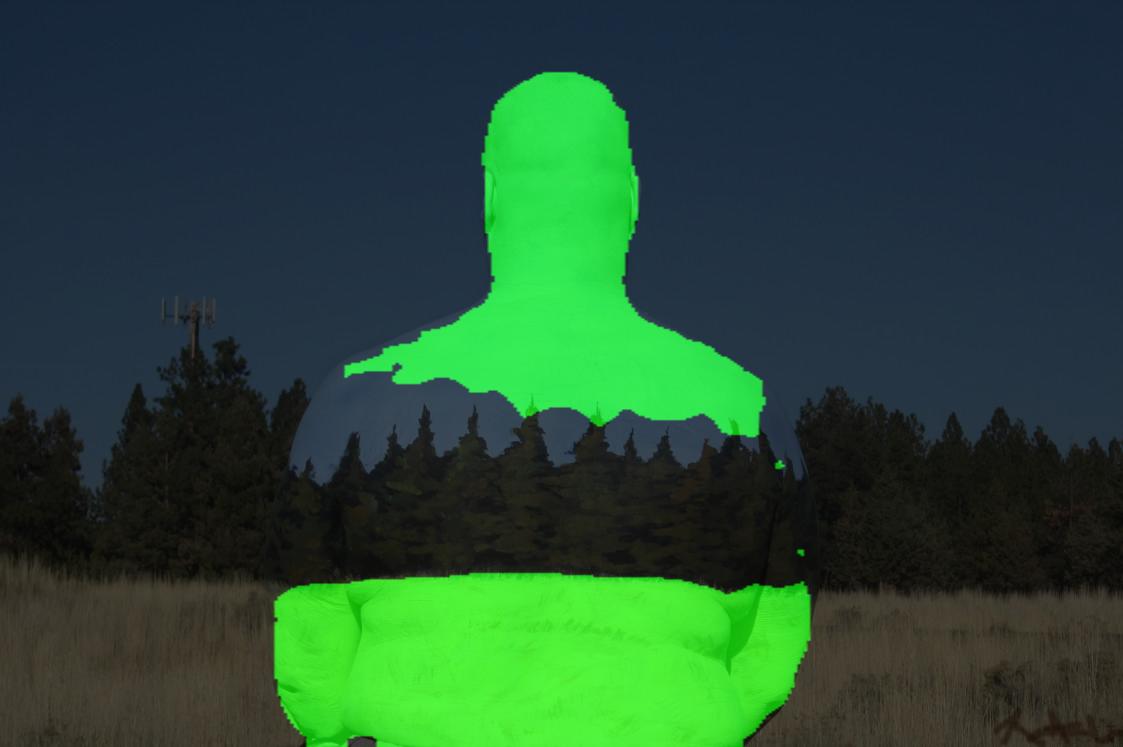} &
        \includegraphics[width=2.25cm, height=1.53cm]{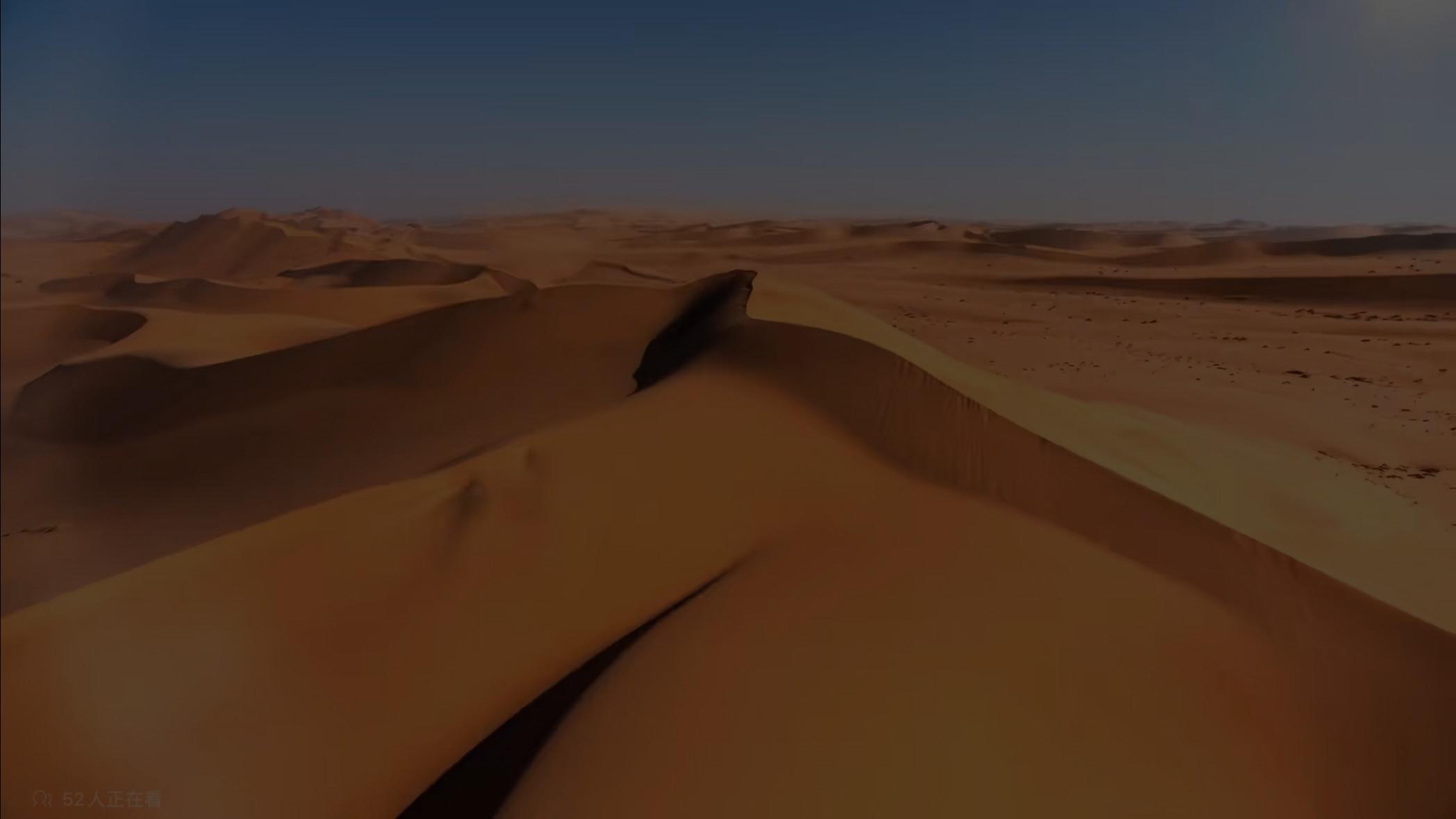} \\
        
        \raisebox{0.5cm}{\makebox[0pt][c]{\rotatebox{90}{\footnotesize \textit{EDN}}}} &
        \includegraphics[width=2.25cm, height=1.53cm]{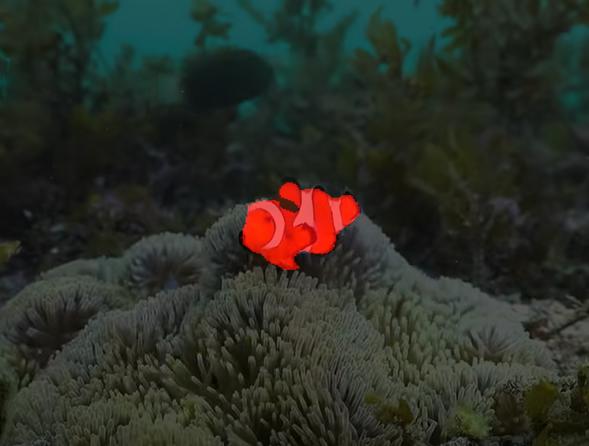} &
        \includegraphics[width=2.25cm, height=1.53cm]{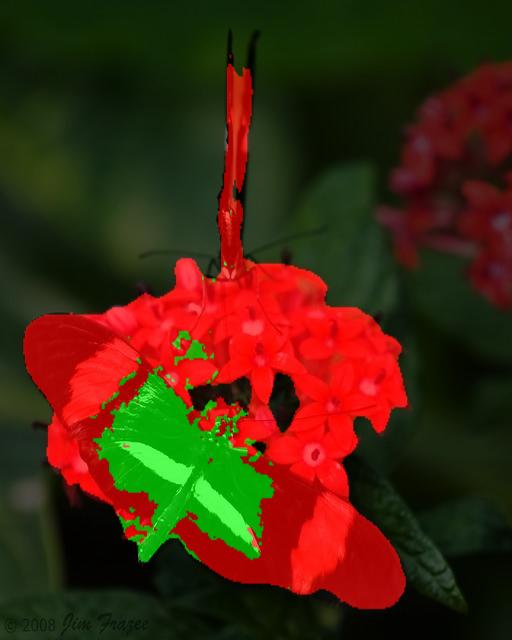} &
        \includegraphics[width=2.25cm, height=1.53cm]{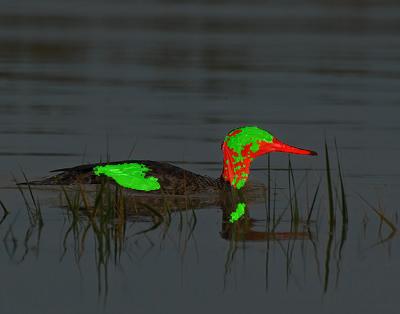} &
        \includegraphics[width=2.25cm, height=1.53cm]{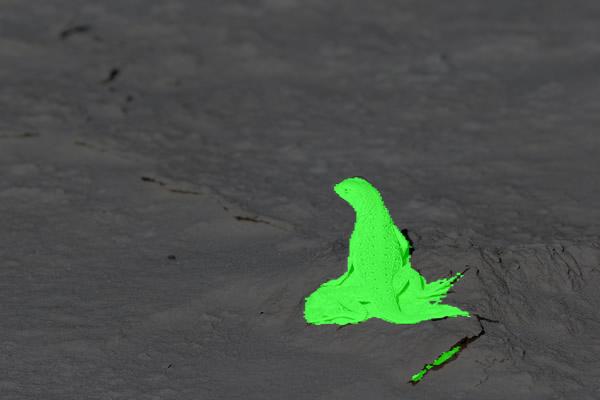} &
        \includegraphics[width=2.25cm, height=1.53cm]{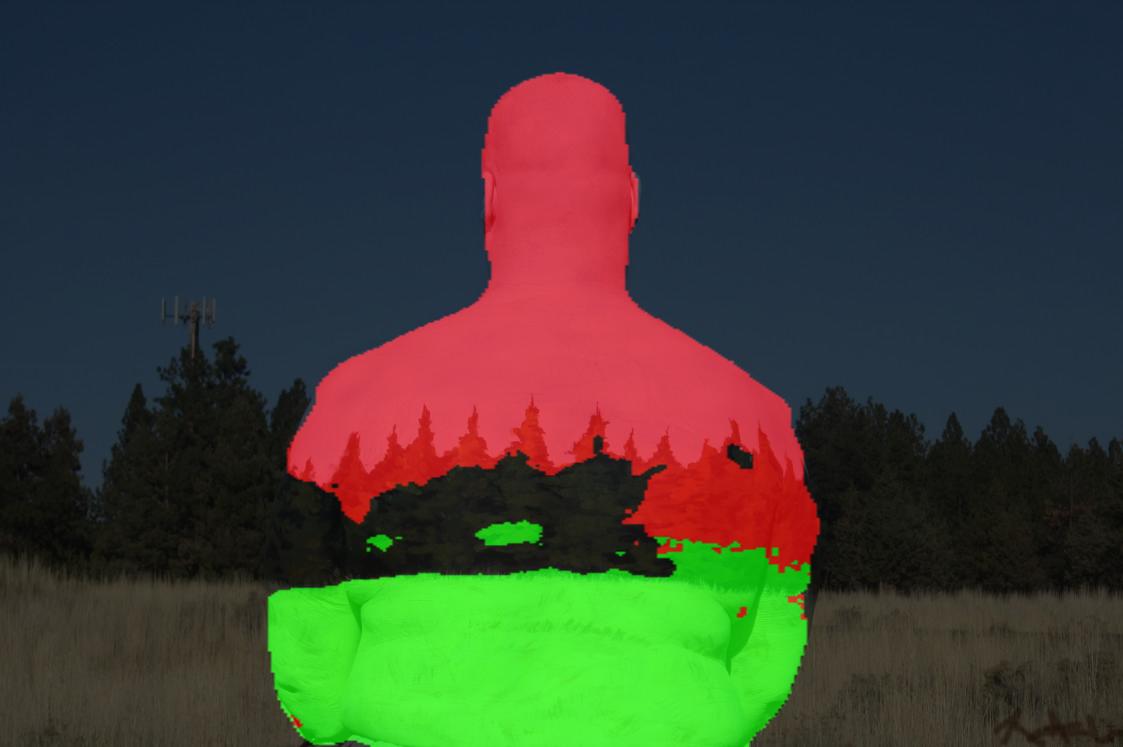} &
        \includegraphics[width=2.25cm, height=1.53cm]{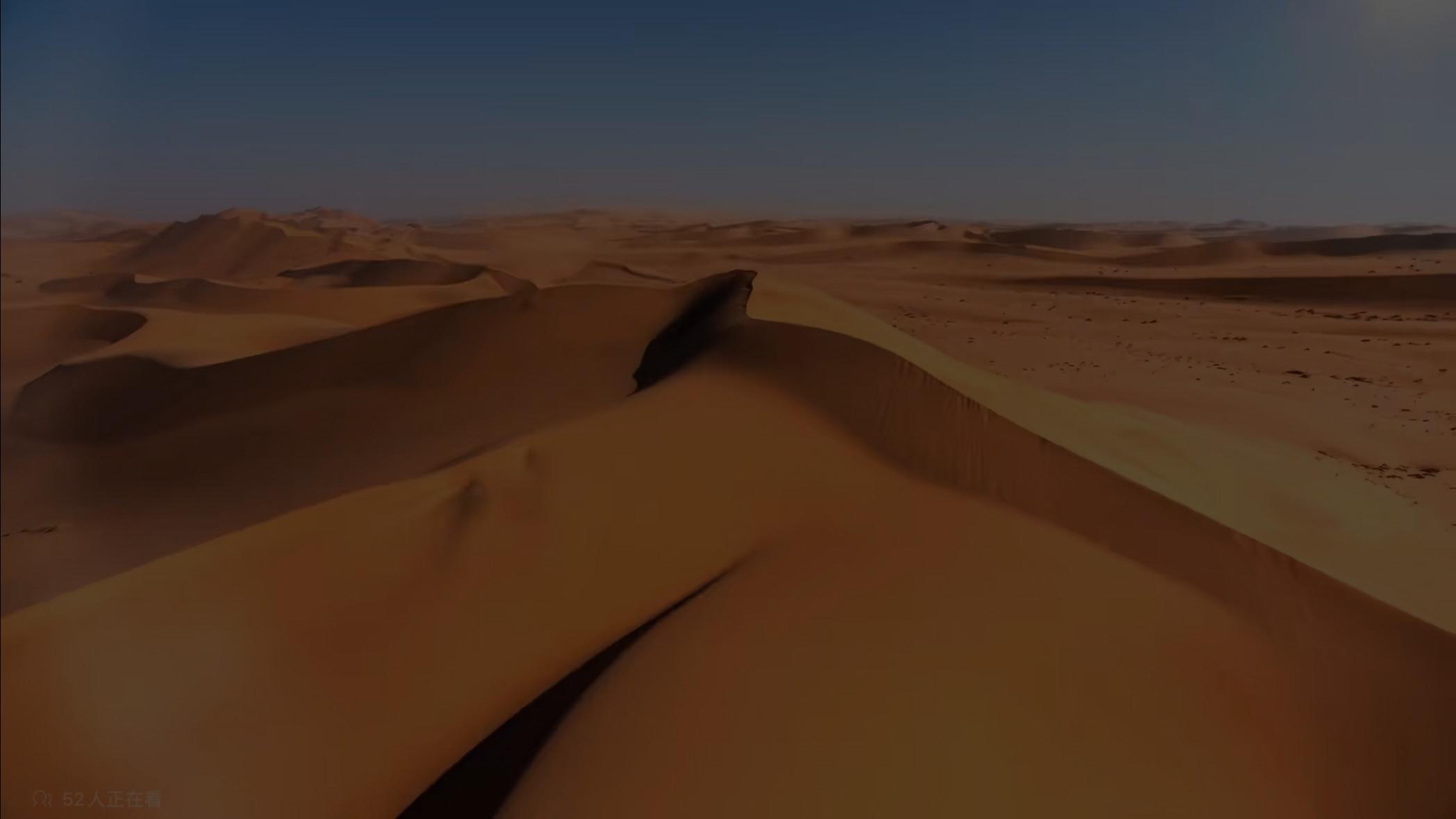} \\

        \raisebox{0.3cm}{\makebox[0pt][c]{\rotatebox{90}{\footnotesize \textit{GateNet}}}} &
        \includegraphics[width=2.25cm, height=1.53cm]{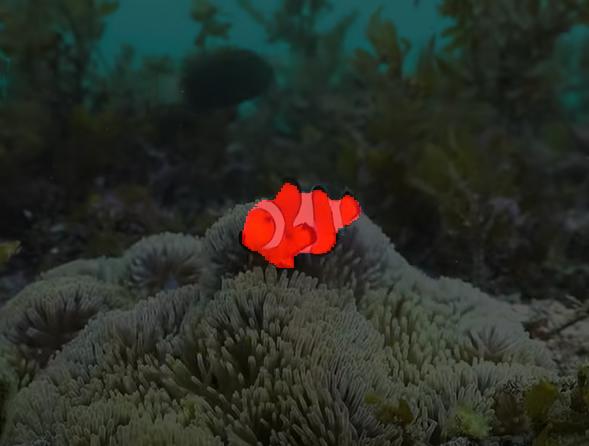} &

        \includegraphics[width=2.25cm, height=1.53cm]{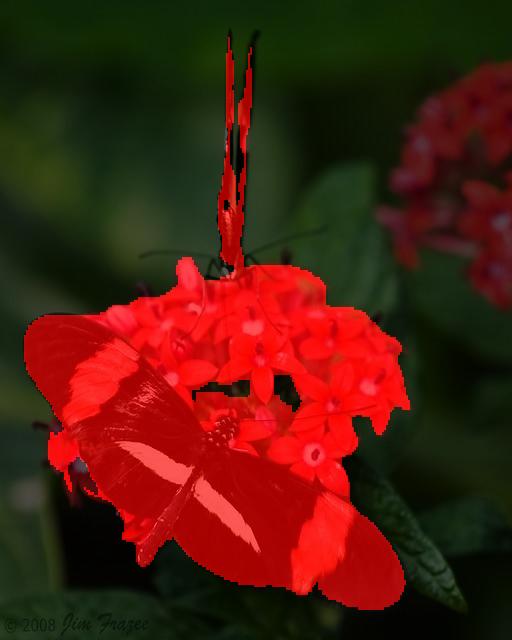} &
                \includegraphics[width=2.25cm, height=1.53cm]{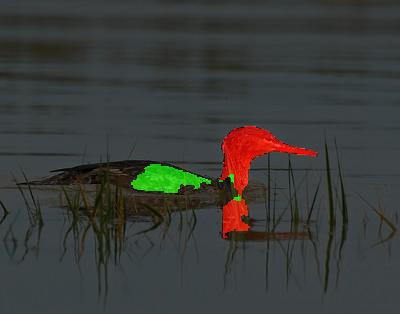} &
        \includegraphics[width=2.25cm, height=1.53cm]{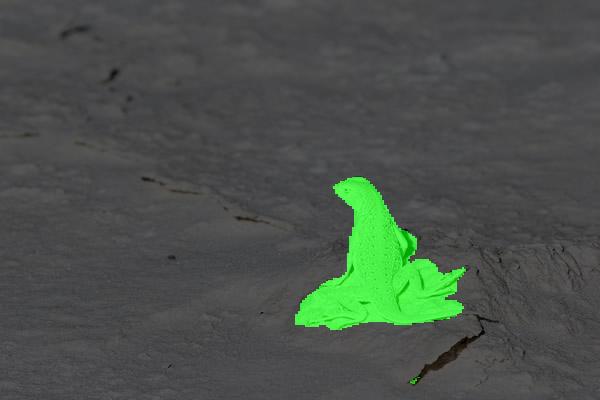} &
        \includegraphics[width=2.25cm, height=1.53cm]{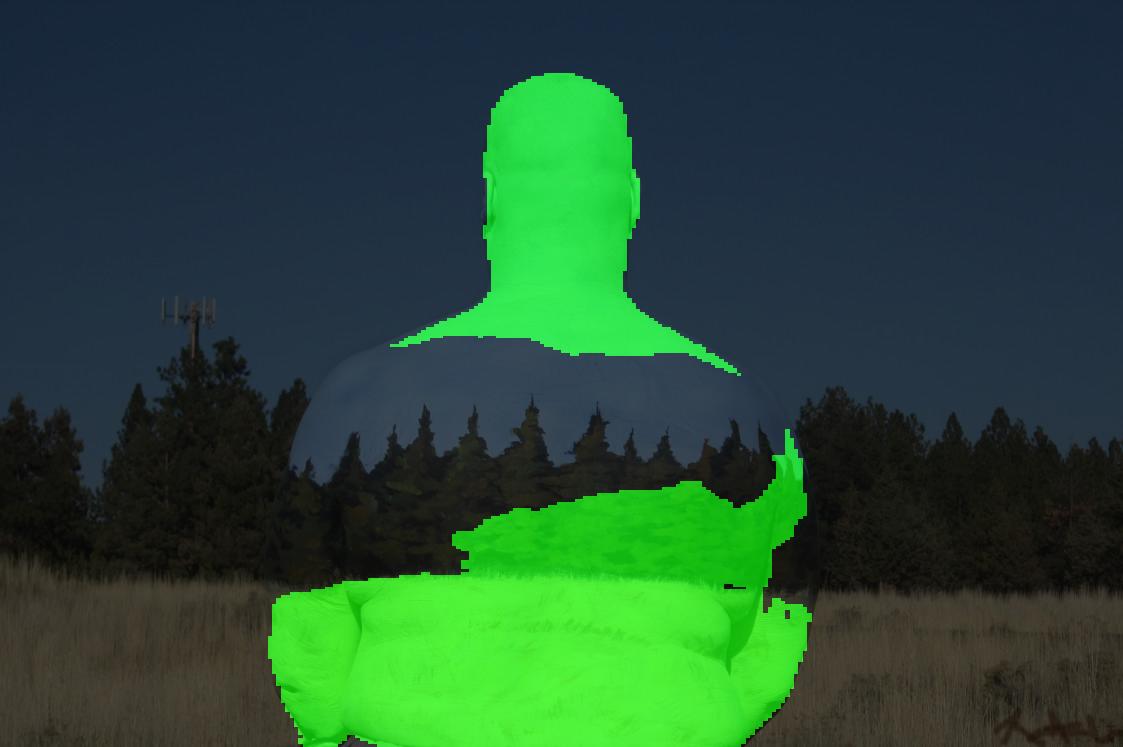} &
        \includegraphics[width=2.25cm, height=1.53cm]{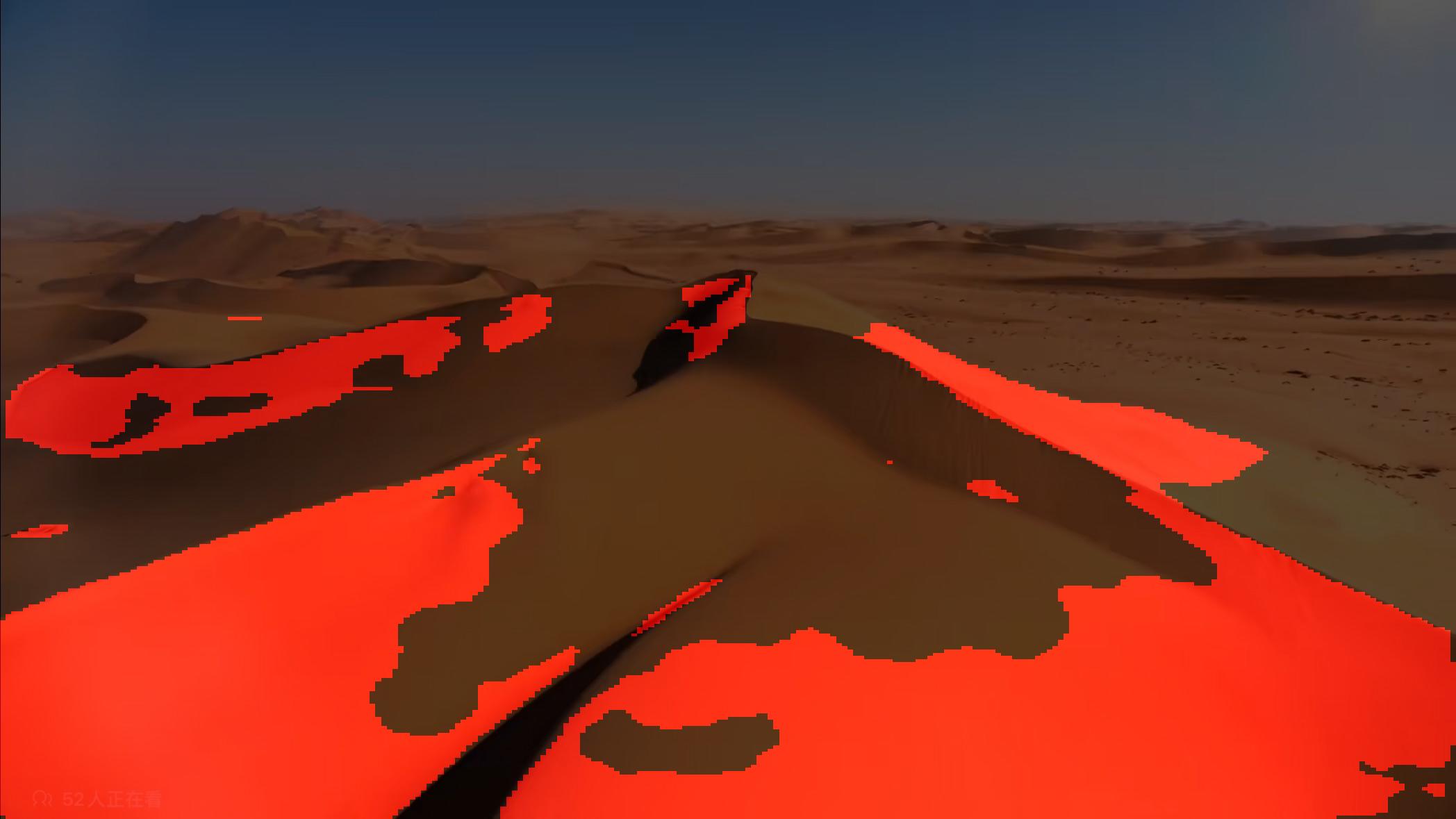} \\

        \raisebox{0.5cm}{\makebox[0pt][c]{\rotatebox{90}{\footnotesize \textit{VST}}}} &
        \includegraphics[width=2.25cm, height=1.53cm]{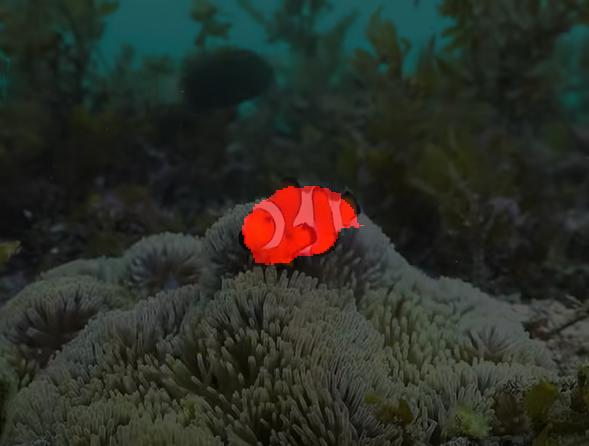} &

        \includegraphics[width=2.25cm, height=1.53cm]{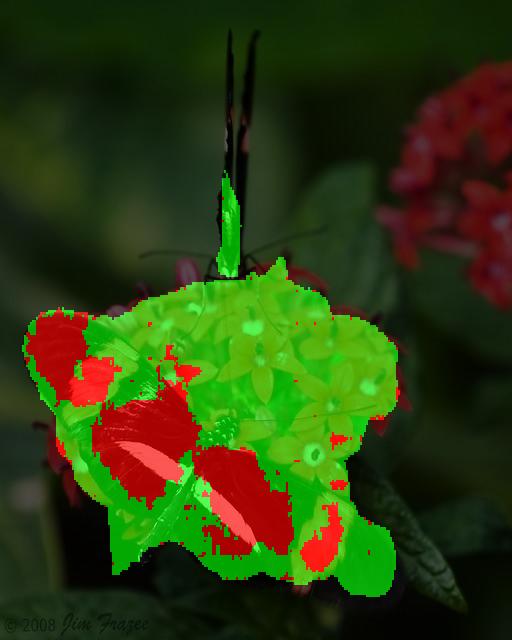} &
                \includegraphics[width=2.25cm, height=1.53cm]{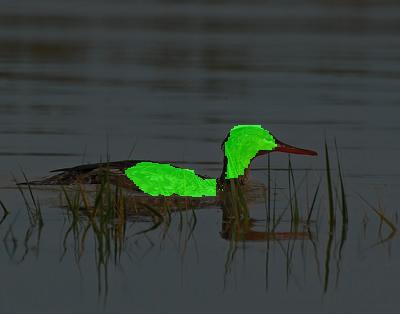} &
        \includegraphics[width=2.25cm, height=1.53cm]{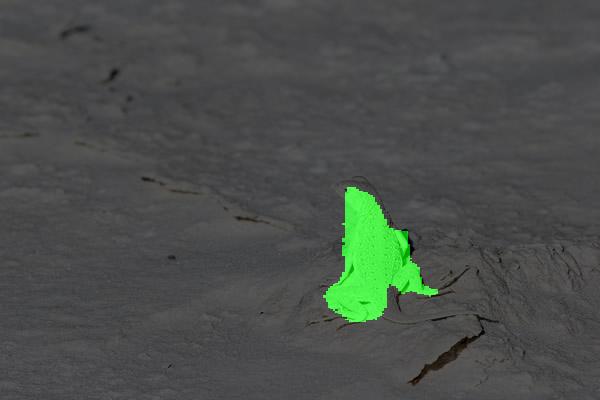} &
        \includegraphics[width=2.25cm, height=1.53cm]{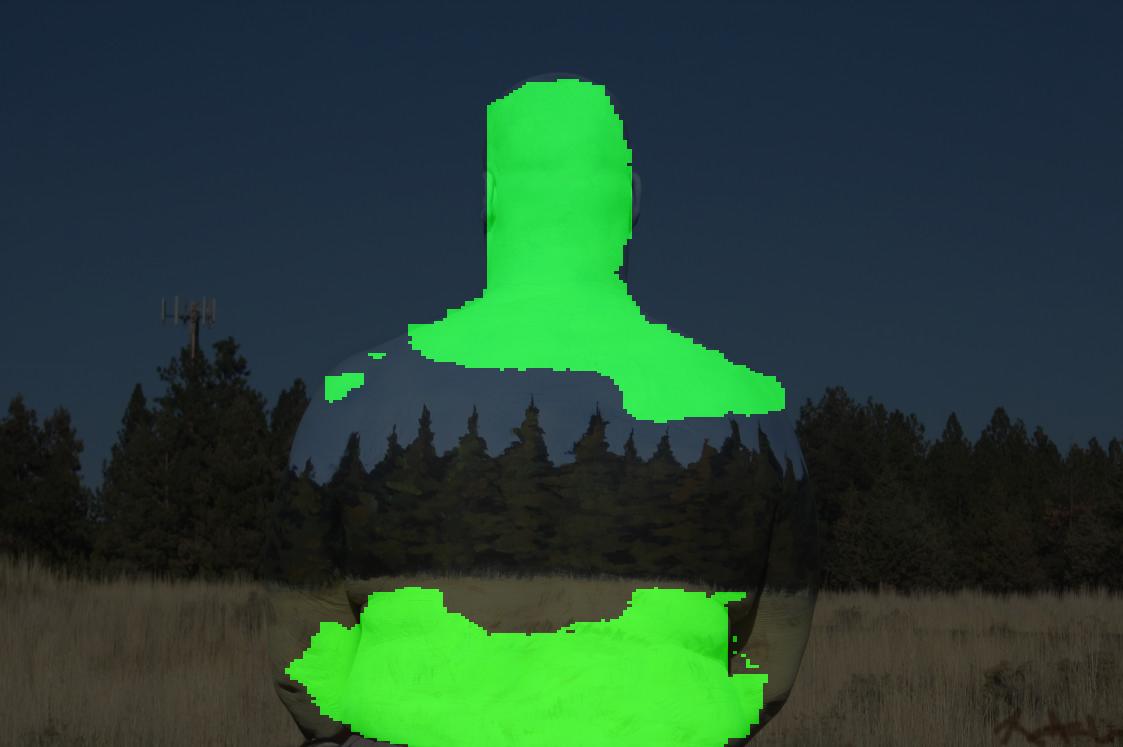} &
        \includegraphics[width=2.25cm, height=1.53cm]{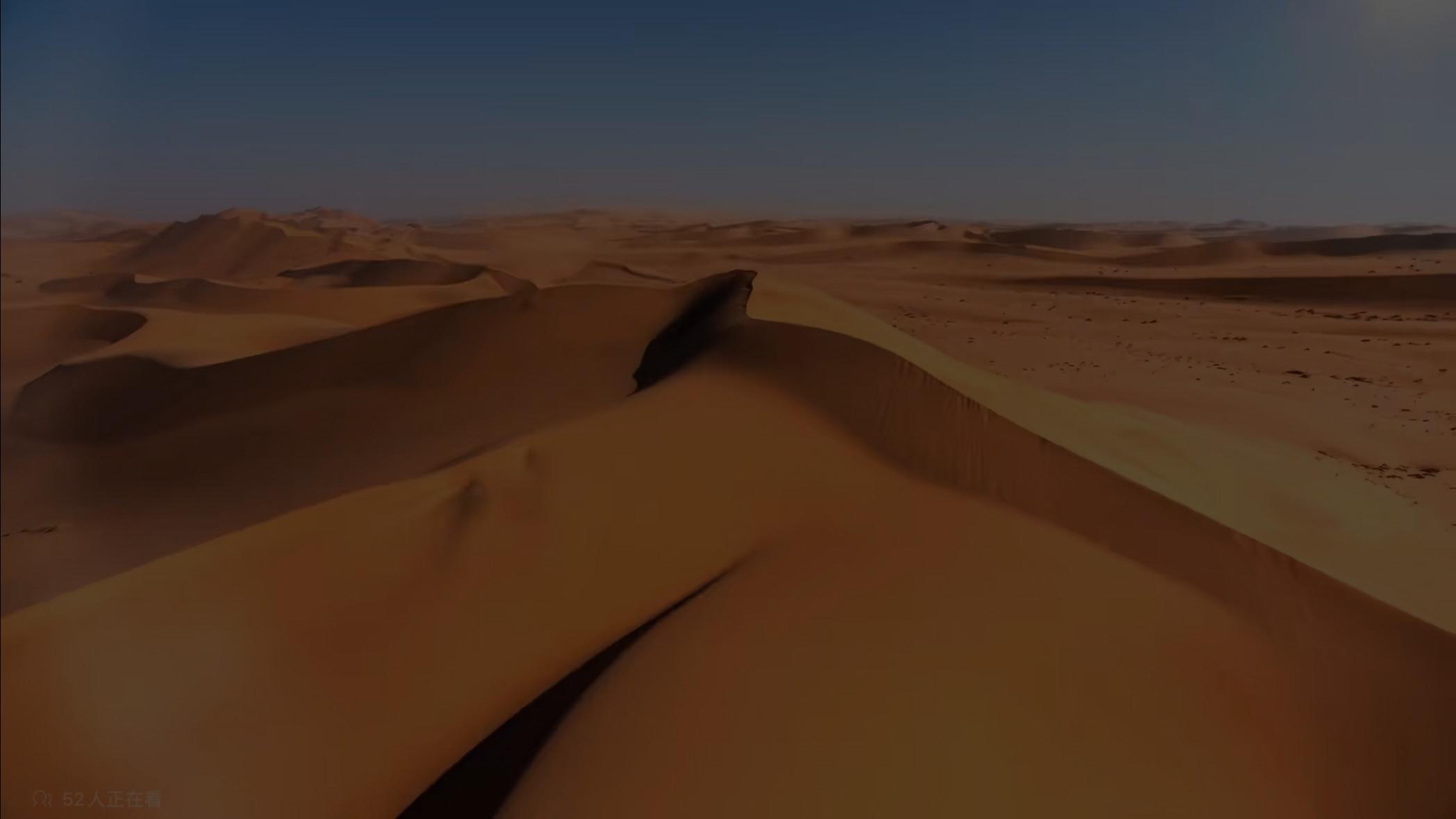} \\

        \raisebox{0.3cm}{\makebox[0pt][c]{\rotatebox{90}{\footnotesize \textit{MSFNet}}}} &
        \includegraphics[width=2.25cm, height=1.53cm]{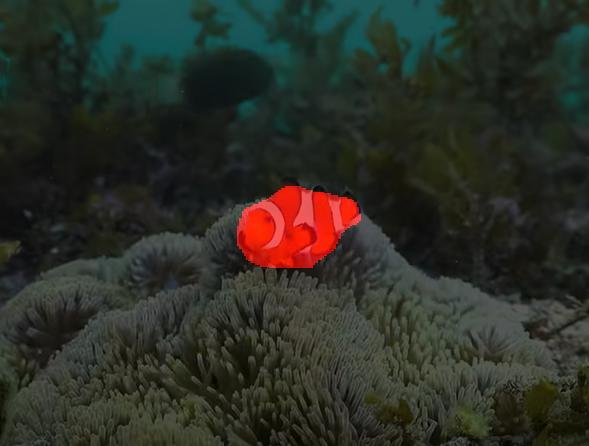} &

        \includegraphics[width=2.25cm, height=1.53cm]{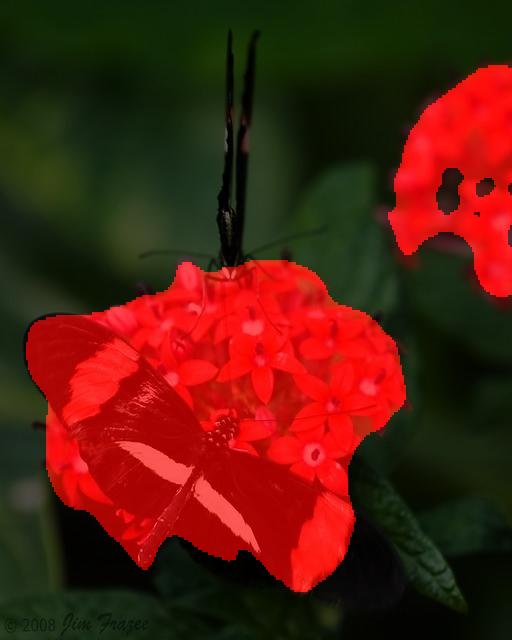} &
                \includegraphics[width=2.25cm, height=1.53cm]{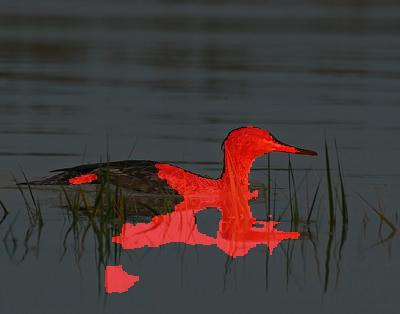} &
        \includegraphics[width=2.25cm, height=1.53cm]{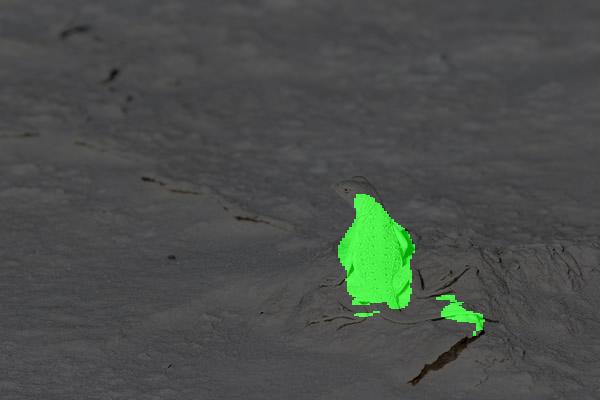} &
        \includegraphics[width=2.25cm, height=1.53cm]{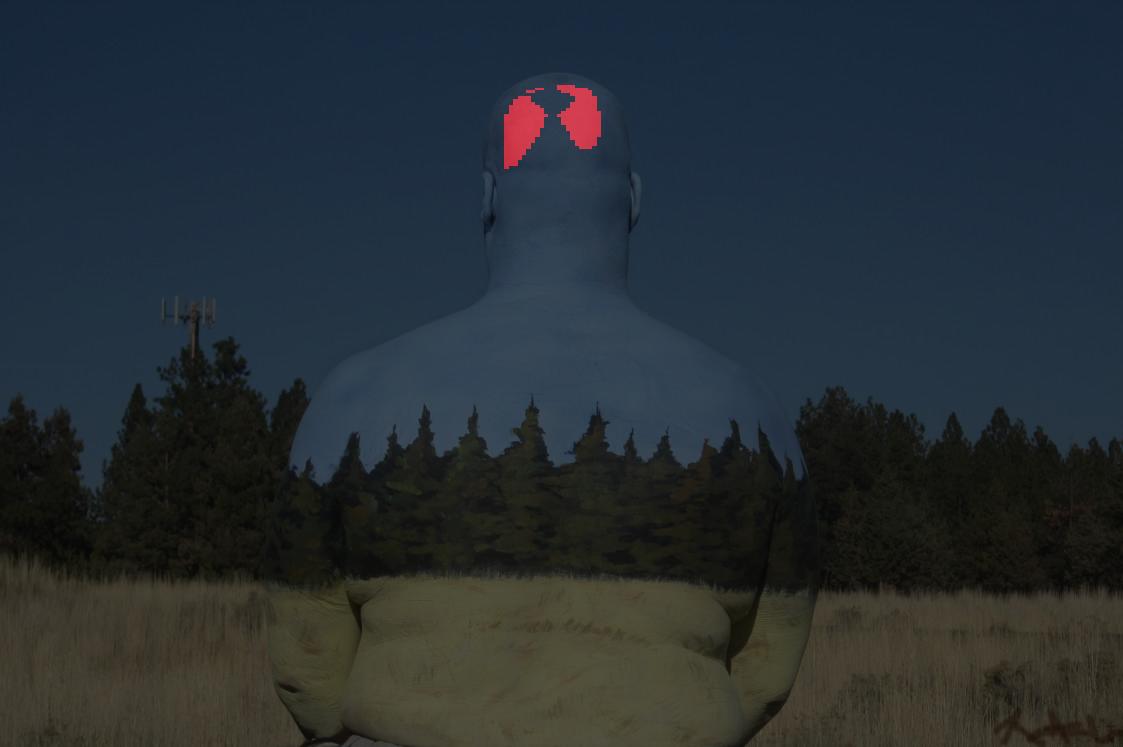} &
        \includegraphics[width=2.25cm, height=1.53cm]{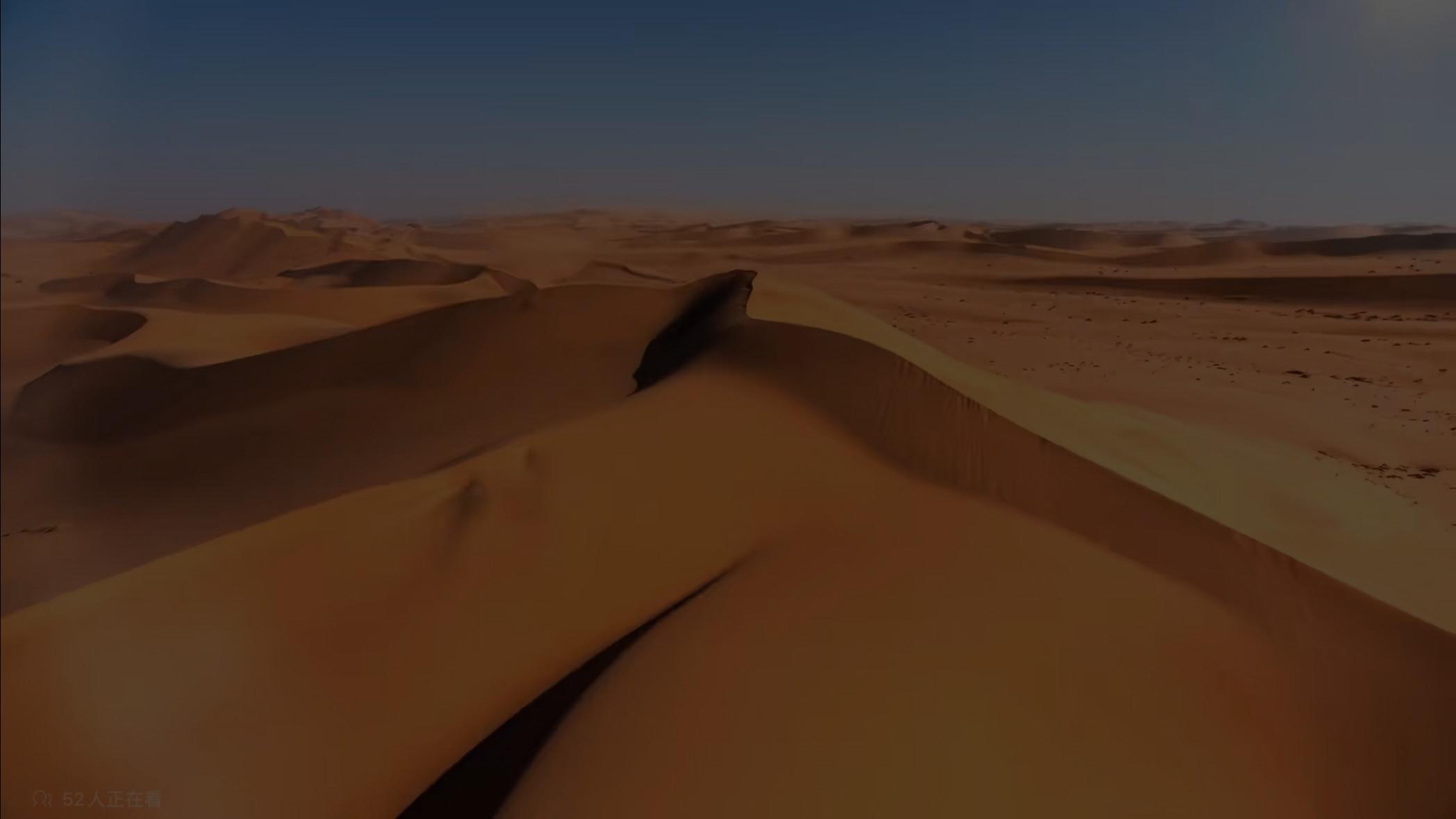} \\

        \raisebox{0.4cm}{\makebox[0pt][c]{\rotatebox{90}{\footnotesize \textit{F3Net}}}} &
        \includegraphics[width=2.25cm, height=1.53cm]{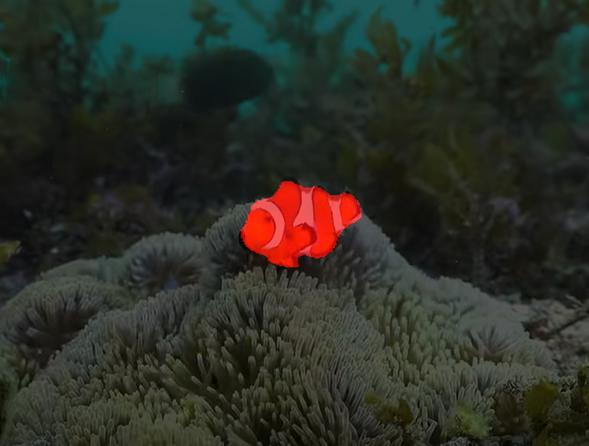} &

        \includegraphics[width=2.25cm, height=1.53cm]{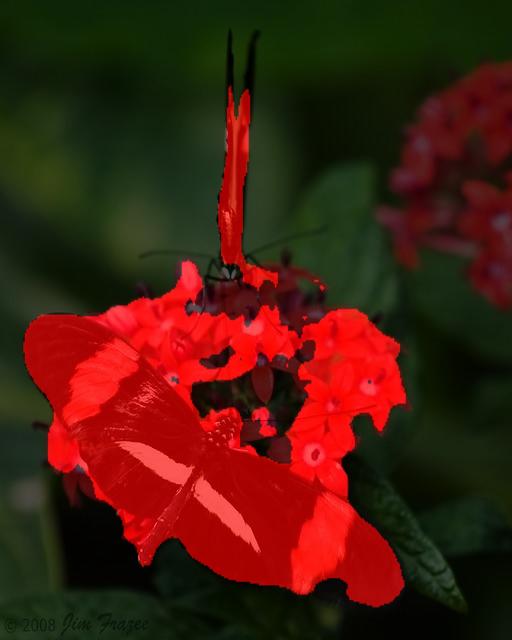} &
                \includegraphics[width=2.25cm, height=1.53cm]{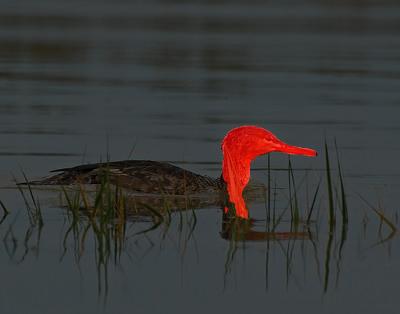} &
        \includegraphics[width=2.25cm, height=1.53cm]{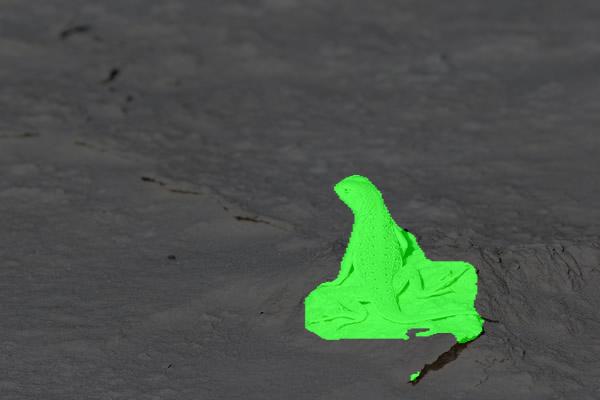} &
        \includegraphics[width=2.25cm, height=1.53cm]{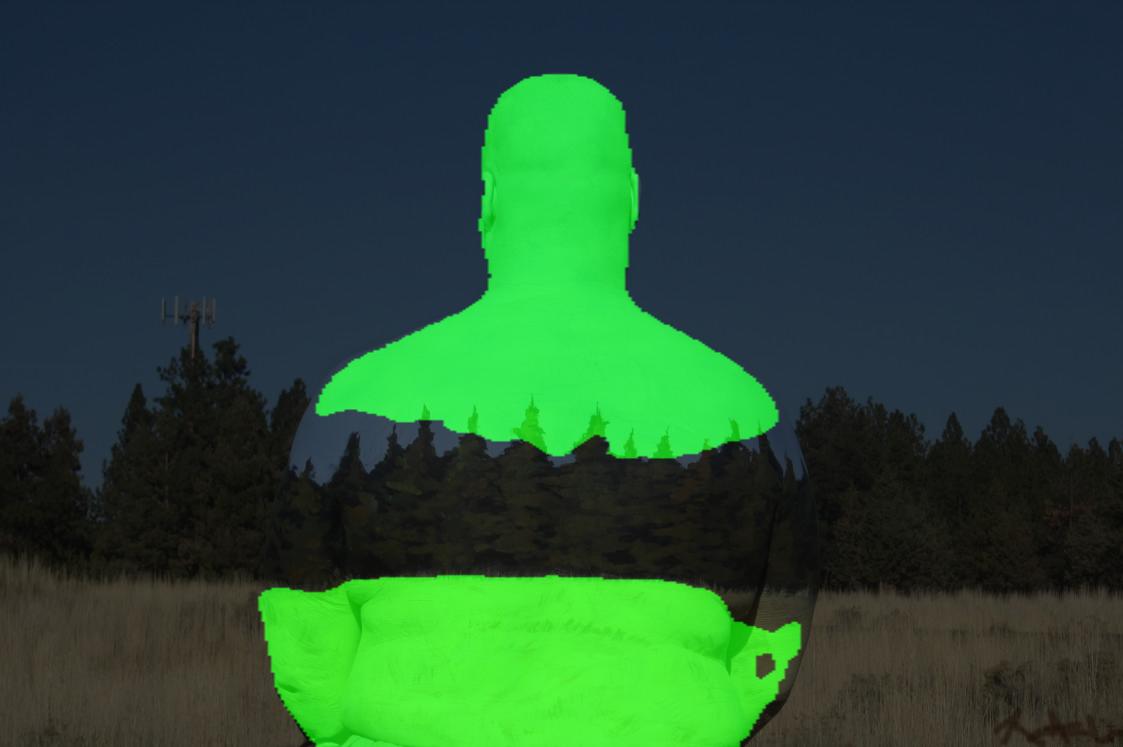} &
        \includegraphics[width=2.25cm, height=1.53cm]{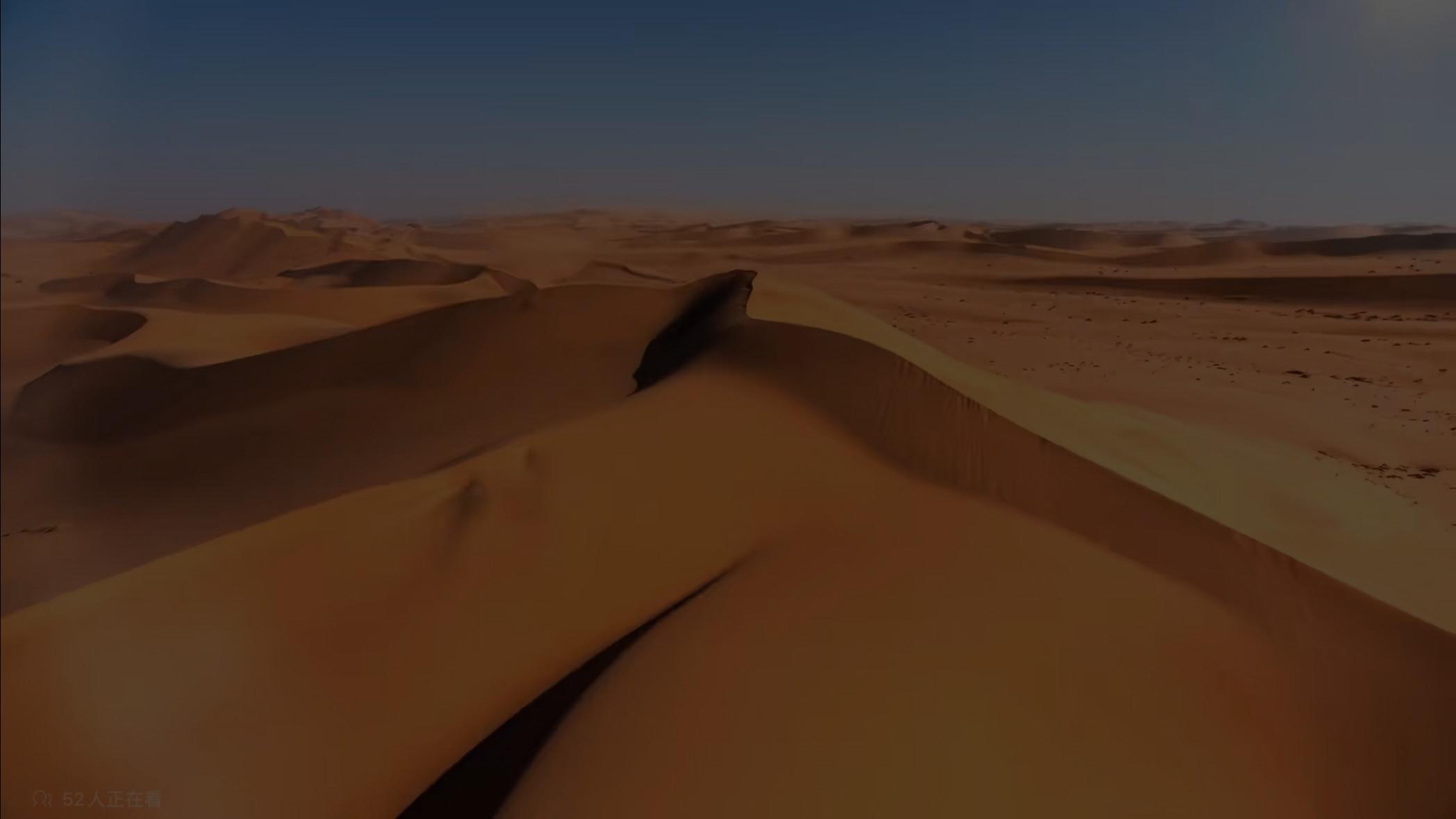} \\
        
    \end{tabular}
    \caption{Additional visualizations of the proposed~\ourmodel~and other state-of-the-art methods on the USC12K test set. \textbf{Zoom-in for better view.}}
    \label{fig:supp_add_visual_comp}
\end{figure*}


\end{document}